\documentclass[runningheads]{llncs}

\usepackage{eccv}
\RequirePackage[width=122mm,left=12mm,paperwidth=146mm,height=193mm,top=12mm,paperheight=217mm]{geometry}

\usepackage{eccvabbrv}

\usepackage{graphicx}
\usepackage{booktabs}

\usepackage[accsupp]{axessibility}  

\usepackage{hyperref}

\usepackage{orcidlink}

\usepackage{adjustbox}
\usepackage{graphicx}
\usepackage{enumitem}
\usepackage{color}
\usepackage{multirow}
\usepackage{tabularx,booktabs}
\usepackage{array, diagbox}
\usepackage{tabu}
\usepackage{arydshln}
\usepackage{amssymb}
\usepackage{float}
\usepackage{amsmath}
\usepackage{booktabs,arydshln}

\makeatletter
\def\adl@drawiv#1#2#3{%
        \hskip.5\tabcolsep
        \xleaders#3{#2.5\@tempdimb #1{1}#2.5\@tempdimb}%
                #2\z@ plus1fil minus1fil\relax
        \hskip.5\tabcolsep}
\newcommand{\cdashlinelr}[1]{%
  \noalign{\vskip\aboverulesep
           \global\let\@dashdrawstore\adl@draw
           \global\let\adl@draw\adl@drawiv}
  \cdashline{#1}
  \noalign{\global\let\adl@draw\@dashdrawstore
           \vskip\belowrulesep}}
\makeatother

\usepackage{pifont}
\newcommand{\xmark}{\textcolor{  red}{\ding{55}}}%
\newcommand{\addimgTtext}[2]{\frame{#1}\raisebox{1pt}{\makebox[0pt][r]{\textcolor{white}{#2}  \phantom{aaaa}}}}

\newcommand{\addimgTtextblack}[2]{\frame{#1}\raisebox{1pt}{\makebox[0pt][r]{\textcolor{black}{#2}  \phantom{aaaa}}}}
\usepackage{makecell}
\newcolumntype{P}[1]{>{\centering\arraybackslash}p{#1}}
\newcolumntype{M}[1]{>{\centering\arraybackslash}m{#1}}
\newcolumntype{C}{>{\centering\arraybackslash}X}

\usepackage{graphicx}
\usepackage{lipsum}
\usepackage[dvipsnames]{xcolor}
\newcommand{\boldparagraph}[1]{\vspace{0.1em}\noindent{\bf #1} }

\makeatletter
\newcommand\notsotiny{\@setfontsize\notsotiny\@vipt\@viipt}
\makeatother

\newcommand{\Fig}{Fig.~}

\newcommand{\Eq}{Eq.~}

\begin{document}

\def\MYTITLE{Retrieval Robust to Object Motion Blur}
\title{\MYTITLE}


\author{Rong Zou\inst{1}\orcidlink{0009-0002-0434-5746} \and
Marc Pollefeys\inst{1,2}\orcidlink{0000-0003-2448-2318} \and
Denys Rozumnyi\inst{1,3}\orcidlink{0000-0001-9874-1349}}

\authorrunning{R. Zou et al.}

\institute{$^1$ETH Zürich, $^2$Microsoft, $^3$Czech Technical University in Prague}

\maketitle

\begin{abstract}
\vspace{-10px}Moving objects are frequently seen in daily life and usually appear blurred in images due to their motion. While general object retrieval is a widely explored area in computer vision, it primarily focuses on sharp and static objects, and retrieval of motion-blurred objects in large image collections remains unexplored. 
We propose a method for object retrieval in images that are affected by motion blur. The proposed method learns a robust representation capable of matching blurred objects to their deblurred versions and vice versa. To evaluate our approach, we present the first large-scale datasets for blurred object retrieval, 
featuring images with objects exhibiting varying degrees of blur
in various poses and scales. We conducted extensive experiments, showing that our method outperforms state-of-the-art retrieval methods on the new blur-retrieval datasets, which validates the effectiveness of the proposed approach. 
Code, data, and model are available at \url{https://github.com/Rong-Zou/Retrieval-Robust-to-Object-Motion-Blur}.
\end{abstract}

\section{Introduction}
\label{sec:intro}

The task of image retrieval has long been a focal point in computer vision research, aiming to efficiently identify images within a large-scale database that share similarities with a given query image. In the realm of image retrieval, methods are often categorized into category-level and instance-level retrieval. 
We focus on the latter, where the objective is to identify images of a specific object featured in a query within a vast, unordered collection of database images. 
In this task, image representations play an important role since they capture the essence of the objects in images and enable the measurement of their similarities. 
Representing images based on hand-crafted features has been extensively studied~\cite{lowe2004SIFT,bay2008SURF}. 
Owing to the development in deep learning, recent strides have been made in adopting learning-based approaches to encode image content~\cite{babenko2014neural, tolias2015max_pool,  noh2017DELF, gem, tolias2020HOW, delg, dolg, lee2022cv, token}. However, the current literature focuses exclusively on image representations for sharp objects, assuming that the objects are stationary and clearly visible in images. By only considering sharp and static objects, the existing retrieval methods fall short in addressing the dynamic nature of real-world scenarios, where objects are in movement, resulting in images blurred by object motion — a phenomenon prevalent in environments with fast-moving objects~\cite{fmo,defmo}, low camera frame rates, and extended exposure times.

\global\long\def\figWidth{0.15\linewidth}
\begin{figure}[t]
	\centering
    \setlength{\tabcolsep}{2pt}
    \setlength{\fboxrule}{1pt} 
    \setlength{\fboxsep}{0pt}  
	\begin{tabular}{
	 P{\figWidth}
	P{\figWidth}
	P{\figWidth}
    P{\figWidth}
    P{\figWidth}
    P{\figWidth}
    }

 {Query}	&   \multicolumn{5}{c}{Top 10 retrieval results}
 
		\\
  		
  \raisebox{1em}{\multirow{2}{*}{\frame{\includegraphics[width=\linewidth]{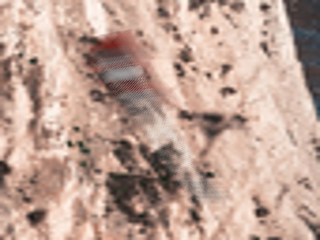}}}}
    &\frame{\includegraphics[width=\linewidth]{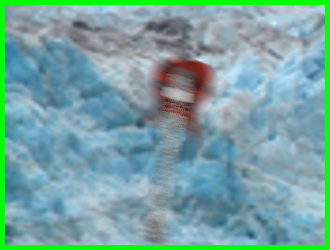}}
      &\frame{\includegraphics[width=\linewidth]{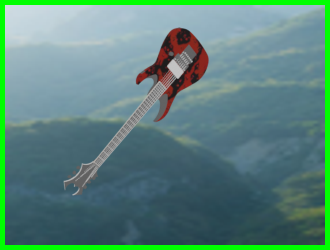}}
      &\frame{\includegraphics[width=\linewidth]{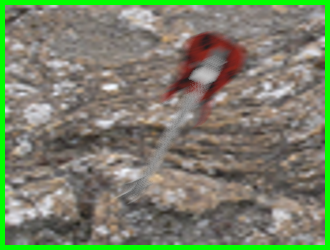}}
      &\frame{\includegraphics[width=\linewidth]{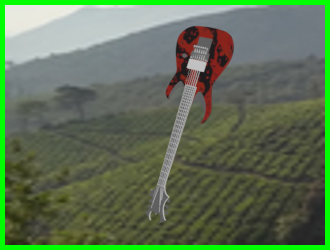}}
      &\frame{\includegraphics[width=\linewidth]{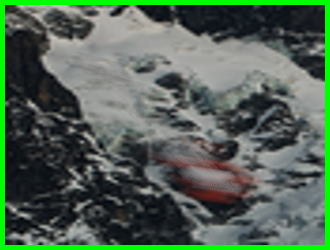}}
      \\
      &\frame{\includegraphics[width=\linewidth]{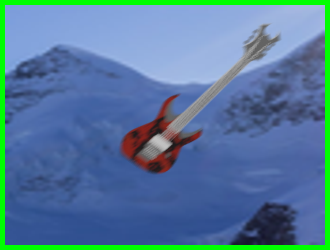}}
      &\frame{\includegraphics[width=\linewidth]{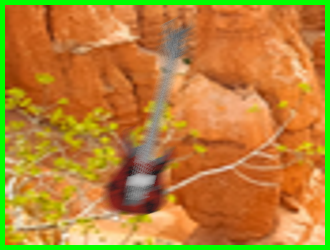}}
      &\frame{\includegraphics[width=\linewidth]{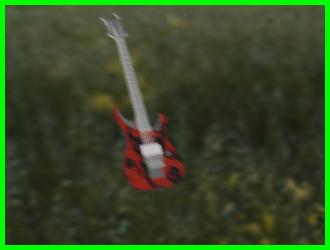}}
      &\frame{\includegraphics[width=\linewidth]{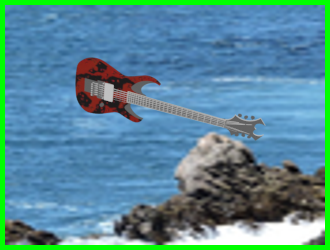}}
      &\frame{\includegraphics[width=\linewidth]{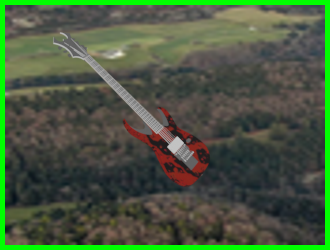}}
		\\

  \raisebox{1em}{\multirow{2}{*}{\frame{\includegraphics[width=\linewidth]{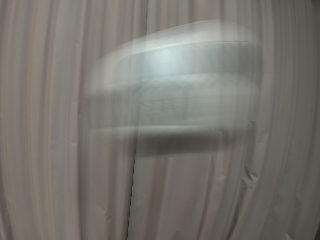}}}}
    &\frame{\includegraphics[width=\linewidth]{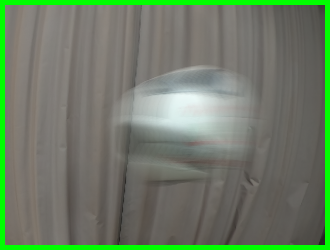}}
      &\frame{\includegraphics[width=\linewidth]{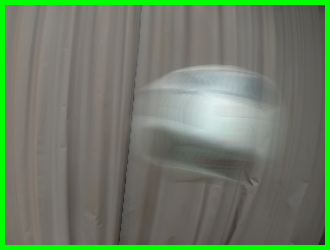}}
      &\frame{\includegraphics[width=\linewidth]{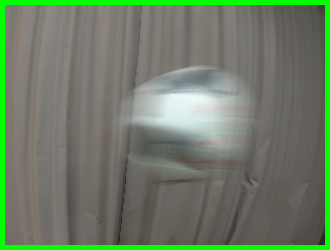}}
      &\frame{\includegraphics[width=\linewidth]{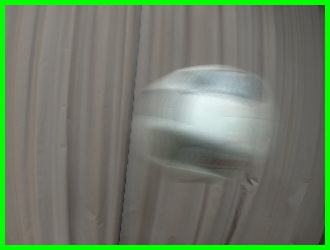}}
      &\frame{\includegraphics[width=\linewidth]{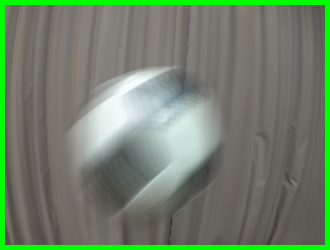}}
      \\
      &\frame{\includegraphics[width=\linewidth]{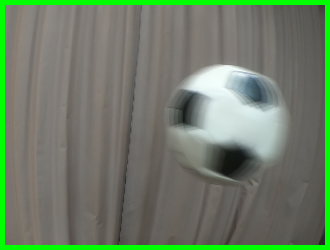}}
      &\frame{\includegraphics[width=\linewidth]{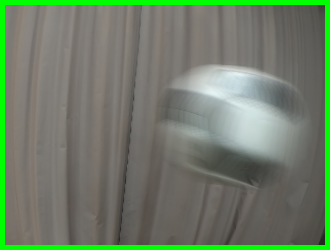}}
      &\frame{\includegraphics[width=\linewidth]{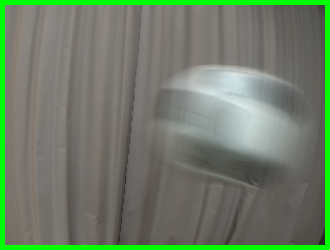}}
      &\frame{\includegraphics[width=\linewidth]{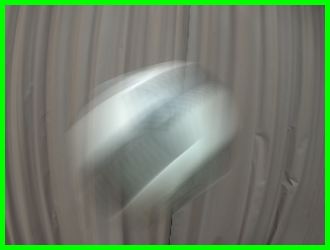}}
      &\frame{\includegraphics[width=\linewidth]{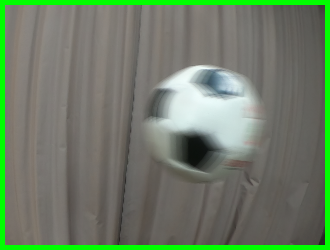}}
		\\
    
	\end{tabular}
    \vspace{-1.2em} 
\caption{\textbf{Retrieval robust to object motion blur.} We present the first approach designed for retrieval in the presence of object motion blur. Our method captures blur-invariant features and generates blur-robust representations. We show retrieval results on the newly created datasets for the task of blur-robust retrieval: synthetic (top) and real (bottom). All retrieved images match the query and are marked with green boxes.}
	\label{fig:teaser}

\vspace{-10px}\end{figure}

To bridge this gap, we introduce a novel task in image retrieval: \textit{retrieval with object motion blur}. The challenge of this task lies in finding image representations robust to object motion blur, in addition to other variations such as background 
and changes in viewpoint. We also propose the first approach targeting this task that produces representations invariant to motion blur.
In response to the unique challenge posed by object motion, we propose loss functions designed to enhance the model's comprehension of blur, thereby ensuring robust retrieval performance even under conditions of extreme blurring, as seen in \Fig \ref{fig:teaser}.

Recognizing the absence of datasets accommodating images with motion-blurred objects in existing retrieval benchmarks, we introduce the first benchmark datasets for this task, including a synthetic dataset for training and evaluation and a real-world dataset created specifically for evaluation. By re-training and evaluating state-of-the-art instance retrieval methods on this comprehensive benchmark, we reveal their shortcomings in handling motion-blurred scenarios. 

This work marks a key step in advancing practical methods for object retrieval in the challenging presence of motion blur. 
It has potential applications such as endangered wildlife monitoring, sports analysis, and security surveillance.

To summarize, our main contributions are as follows:
\begin{enumerate}[itemsep=0.1pt,topsep=3pt,leftmargin=*,label=\textbf{(\arabic*)}]
    \item We present the first method specifically designed for the novel retrieval task involving object motion blur. The method is trained with loss functions tailored to improve its understanding of motion blur.
    \item We introduce a new benchmark featuring synthetic and real-world datasets specifically constructed for this task. The datasets are carefully processed and are directly applicable for future research in blur retrieval.
    \item We conducted extensive experiments and detailed analyses to validate the effectiveness of our approach. Our method outperforms state-of-the-art retrieval methods and demonstrates superior robustness to motion blur.
\end{enumerate}

\vspace{-0.5em}
\section{Related work}
\label{sec:related}

{\boldparagraph{Standard retrieval.}
In the area of standard image retrieval, local feature-based methods have evolved from traditional hand-crafted techniques like SIFT~\cite{lowe2004SIFT} and SURF~\cite{bay2008SURF} to contemporary deep learning-based approaches such as DELF~\cite{noh2017DELF} and HOW~\cite{tolias2020HOW}. 
Considering that motion blur can corrupt local information, which may make local descriptors less robust, we adopt a global approach for our task.
Classical methods typically derive global features by aggregating hand-crafted local features using techniques such as BoW~\cite{sivic2003BoW}, Fisher vectors~\cite{perronnin2010Fisher}, VLAD~\cite{jegou2011VLAD}, AMSK~\cite{tolias2016AMSK}, \etc. Modern deep learning-based approaches, on the other hand, generate global descriptors by applying various pooling operations on feature maps encoded by convolutional neural networks (CNNs), such as sum pooling~\cite{babenko2015sum_pool1, tolias2020HOW}, weighted sum pooling~\cite{kalantidis2016wight_sum_pool}, max pooling, regional max pooling~\cite{tolias2015max_pool}, and generalized mean-pooling (GeM)~\cite{gem}, which has gained prominence in recent literature~\cite{delg,dolg,lee2022cv}. Our method integrates insights from recent advancements in global feature-based methods, adopting GeM pooling~\cite{gem} for feature extraction and incorporating contrastive loss~\cite{chopra2005contrastive} and ArcFace loss~\cite{arcface} during training.}

Among the single-pass global feature-based methods, state-of-the-art standard retrieval approaches such as DELG global branch~\cite{delg}, DOLG~\cite{dolg}, and Token~\cite{token} face challenges when handling images affected by motion blur, leading to degraded performance. 
Our work stands out in addressing this critical gap in the literature, as existing retrieval methods overlook the impact of motion blur. To bridge this research gap, we propose novel loss functions aimed at enhancing the model's comprehension of object motion blur, thereby contributing to the advancement of image retrieval in motion-blurred scenarios.

\boldparagraph{Blur handling.}
Handling blur in images is an important topic in computer vision. 
~\cite{vasiljevic2016cls_seg_blur} investigated the effect of different types of blur (defocus, uniform subject motion, camera shake) on the performance of CNNs for image classification and semantic segmentation. ~\cite{Sayed_2021_object_detect} studied online object detection under the influence of camera ego-motion-induced blur. 
Image and video deblurring are long-standing tasks in computer vision, which have been addressed by a variety of methods~\cite{Kupyn_2018_CVPR,8099888,zhou2019stfan, wang2022video_deblur1, zhang2022video_deblur2}. 
Joint deblurring and trajectory estimation for motion-blurred objects have also been extensively explored, with notable contributions from methods like~\cite{tbd,tbd3d,tbdnc,defmo}, some extending their scope to 3D reconstruction~\cite{sfb,mfb}. These existing works predominantly focus on other types of blur or address distinct tasks, emphasizing the need for specialized efforts in the context of our targeted task of retrieval involving object motion blur.

\boldparagraph{Datasets.}
Several datasets have been proposed that contain generic corruptions including blur, such as ImaneNet-C~\cite{hendrycks2019robustness} and Coco-C~\cite{michaelis2019dragon}, but they do not focus purely on blurry images, and the types of blur they encompass (Gaussian, defocus, frosted glass, camera shake, and zoom) are not induced by object motion.
Many datasets contain blurry-sharp image pairs, such as RealBlur~\cite{realblur} and GoPro~\cite{gopro}, but they are created for deblurring tasks and not for retrieval, and the blur in images is mainly caused by camera motion or camera shake, not by object motion. 
A large variety of standard retrieval datasets have been proposed, including Revisited Oxford5k and Paris6k~\cite{radenovic2018revisiting}, Google Landmarks V2~\cite{weyand2020google,yokoo2020two}, SfM120K~\cite{radenovic2017fine}, etc.
However, none of them incorporate motion-blurred images. 
To the best of our knowledge, there is currently no existing dataset tailored for the task of retrieval with object motion blur.
In view of this, we created synthetic and real-world datasets specifically designed for the novel task of blur retrieval.
\begin{figure*}[t] 
    \centering
    \includegraphics[width=\textwidth]{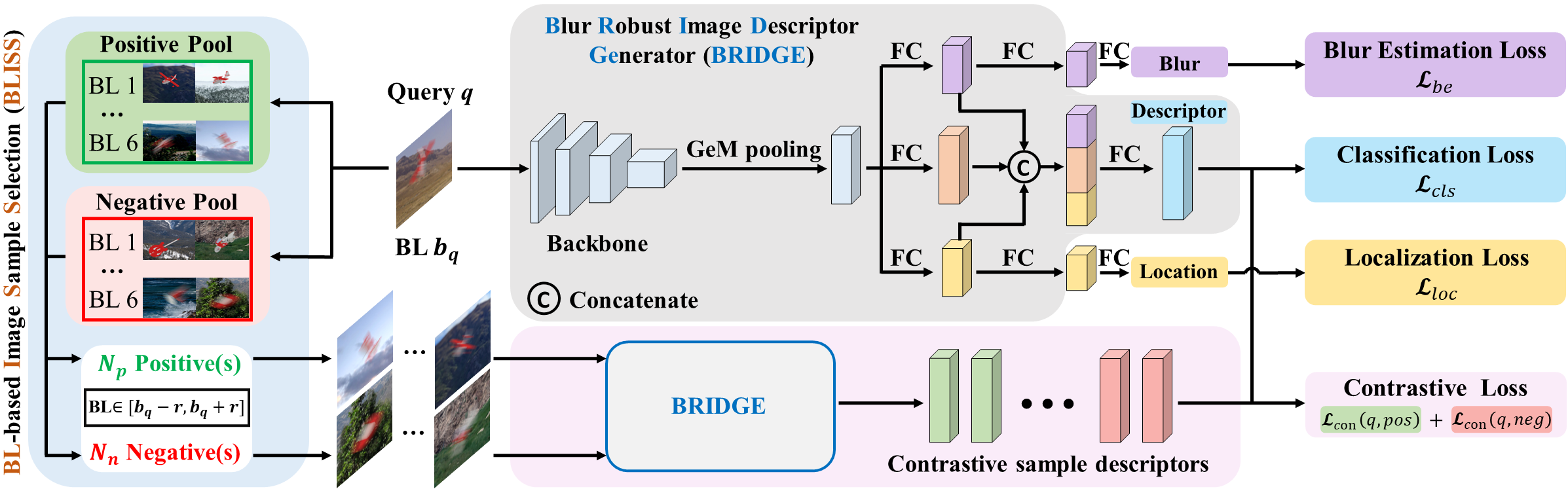} 
    
    \vspace{-0.8em} 
    \caption{
    \textbf{Overview of the proposed method for blur-robust retrieval.} Our model takes a query image as input. Then, it generates a descriptor robust to object motion blur through the \textit{Blur Robust Image Descriptor Generator} (BRIDGE) module that encodes the input image using a CNN backbone followed by generalized mean-pooling (GeM) \cite{gem}, extracting a feature vector. This vector is subsequently fed into three heads: the blur estimation head, classification head, and localization head. Each head processes the feature vector to extract relevant information for identifying motion-blurred objects. The resulting three processed feature vectors are concatenated and passed through the final fully connected (FC) layer to produce the blur-robust descriptor. During training, the \textit{Blur Level-based Image Sample Selection} (BLISS) mechanism is employed to select contrastive samples based on the query blur level (BL, Eq.~\eqref{eq:bl}) $b_q$ and a specified blur level range $r$. Then, these selected image samples are input into the BRIDGE module to extract descriptors for subsequent contrastive learning.
    \label{fig:method_overview}
}

\vspace{-3px}\end{figure*} 

\section{Method}
\label{sec:method}

We perform retrieval via nearest neighbor search in a representation space.
Therefore, we require a representation model that is robust to object motion blur. 
We train a deep learning-based model that maps images to a vector space. 
The network is trained on our proposed synthetic dataset using multiple losses, including those tailored to object motion blur.
As we show, the method generalizes well to real data despite being trained on synthetic data.

The network takes an RGB image $\mathbf{I} \in \mathbb{R}^{3 \times H \times W}$ containing a sharp or blurred object as input and generates a descriptor $\mathbf{D} \in \mathbb{R}^{d}$ robust to object motion blur through the \textit{Blur Robust Image Descriptor Generator} module (BRIDGE, Sec.~\ref{sec:bridge}). 
For training only, contrastive samples are selected by the \textit{Blur Level-based Image Sample Selection} mechanism (BLISS, Sec.~\ref{sec:bliss}) based on the query blur level $b_q$ and the contrast blur level range $r$. 
The selected image samples are sent into the BRIDGE module to extract descriptors for contrastive learning. 
The architecture of the proposed blur retrieval method is shown in \Fig \ref{fig:method_overview}. 

\subsection{BRIDGE module}
\label{sec:bridge}
The BRIDGE module encodes an image into a descriptor that is robust to object motion blur.
The input image is encoded by a ResNet-50 backbone~\cite{resnet} to produce a feature map \( \mathbf{\Omega} \in \mathbb{R}^{C \times U \times V} \). 
Then, we use GeM pooling~\cite{gem} to process the feature map into a feature vector $\mathbf{f} \in \mathbb{R}^{C}$, expressed as 
\begin{equation}
    \textbf{f} = \left[\left(\frac{1}{|\boldsymbol{\Omega}|}\sum_{(u,v)\in\boldsymbol{\Omega}}x^{p}_{(c,u,v)}\right)^{\frac{1}{p}}\right]_{c=1,\cdots,C},
\end{equation}
where $p>0$ is a parameter, and setting it to values larger than 1 forces the pooled vector to focus on the salient areas. 
This feature vector then undergoes multi-head processing to generate a motion blur-robust descriptor.
Specifically, in the blur estimation head, we obtain blur estimation features $\mathbf{f}_{be} \in \mathbb{R}^{C_b}$ by a fully connected (FC) layer. 
The same dimension reduction is performed in the localization head, which generates features for location prediction $\mathbf{f}_{loc} \in \mathbb{R}^{C_l}$. 
The features generated from the two heads are then concatenated with the whitened \cite{gordo2016whiten} features $\mathbf{f}_{cls} \in \mathbb{R}^{C_c}$ from the classification head. 
We further reduce the dimension of the concatenated features to obtain the final blur-robust descriptor of the image. 
This can be expressed as:
\begin{equation}
\mathbf{D} = \mathbf{W} \cdot Concat (\mathbf{W}_{be} \mathbf{f}, \mathbf{W}_{loc} \mathbf{f}, \mathbf{W}_{cls} \mathbf{f}),
\end{equation}
where $\mathbf{W}_{be} \in \mathbb{R}^{C_b \times C}$,  $\mathbf{W}_{loc} \in \mathbb{R}^{C_l \times C}$ and $\mathbf{W}_{cls} \in \mathbb{R}^{C_c \times C}$ are weights of the FC layers in the blur estimation, localization, and classification heads, respectively, and $\mathbf{W} \in \mathbb{R}^{d \times (C_b + C_l + C_c)}$ is the weight of the final dimension-reduction layer. 

\subsection{BLISS mechanism}
\label{sec:bliss}

While the specific representation of the degree of motion blur for objects in images is yet to be formally defined, it is intuitively perceived as a continuous value. The level of detail exhibited by objects varies with the degree of blur: sharper objects show clear structures, textures, and poses, while more blurred objects primarily reveal color characteristics (see \Fig \ref{fig:data_examples}). This continuous nature of blur allows us to indirectly control the distance between samples with significantly different blur conditions during learning image representations. In this context, 
we manage the distance between samples that exhibit visual features and patterns of the same level of detail due to similar degrees of blur, thus indirectly influencing the distance between samples with significantly different levels of blur.
Additionally, we can directly control the distance between samples with large differences in blur conditions by imposing constraints directly between such samples in the descriptor space. There are therefore two key research questions: 
What is the optimal value for the range of differences in the blur levels of image pairs used in contrastive learning? How does adjusting the range of differences in blur levels affect the model’s performance? 
To answer these questions and identify the optimal blur level range for contrastive learning, we introduce the BLISS mechanism. This mechanism selects contrastive samples based on the blur level in the query image for contrastive learning.

\global\long\def\figWidth{0.15\linewidth}
\begin{figure*}[t]
	\centering
 \begin{subfigure}{0.48\linewidth} 
    \setlength{\tabcolsep}{2pt}
	\begin{tabular}{
	M{\figWidth}
	M{\figWidth}
	M{\figWidth}
    M{\figWidth}
    M{\figWidth}
    M{\figWidth}
    }
		BL 1 & BL 2 & BL 3 & BL 4 & BL 5 & BL 6
		\\
		
  \frame{\includegraphics[width=\linewidth]{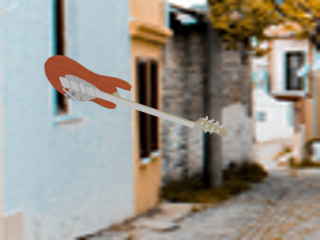}}
	   &\frame{\includegraphics[width=\linewidth]{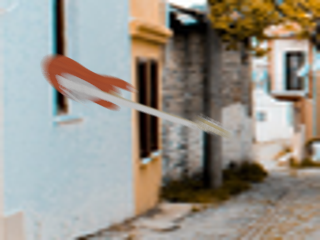}}
	   &\frame{\includegraphics[width=\linewidth]{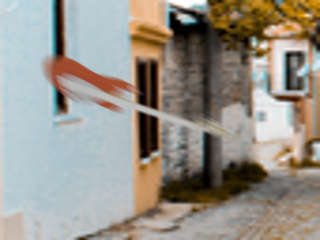}}
      &\frame{\includegraphics[width=\linewidth]{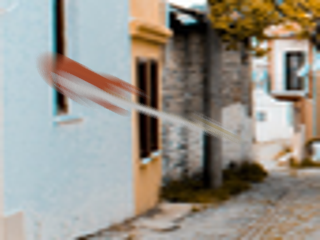}}
      &\frame{\includegraphics[width=\linewidth]{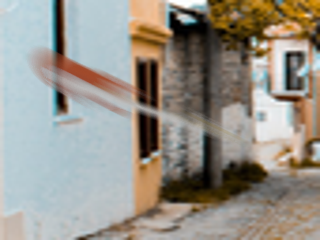}}
      &\frame{\includegraphics[width=\linewidth]{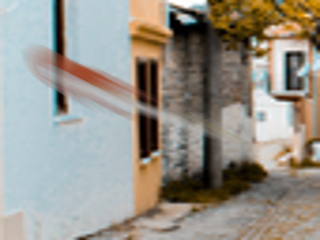}}
		\\
  \frame{\includegraphics[width=\linewidth]{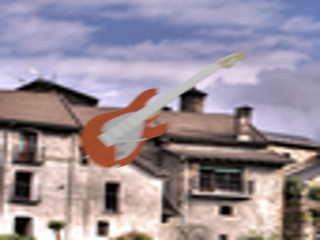}}
	   &\frame{\includegraphics[width=\linewidth]{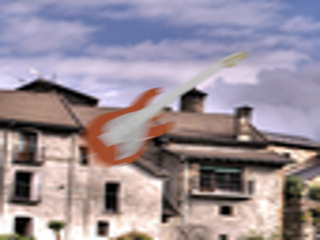}}
	   &\frame{\includegraphics[width=\linewidth]{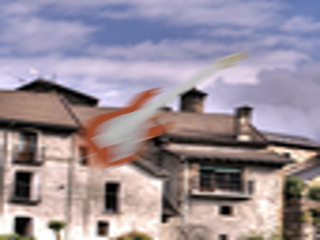}}
      &\frame{\includegraphics[width=\linewidth]{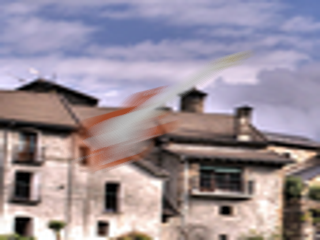}}
      &\frame{\includegraphics[width=\linewidth]{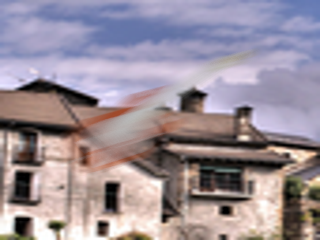}}
      &\frame{\includegraphics[width=\linewidth]{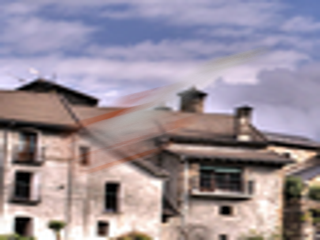}}
		\\
  \frame{\includegraphics[width=\linewidth]{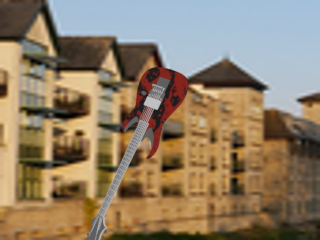}}
	   &\frame{\includegraphics[width=\linewidth]{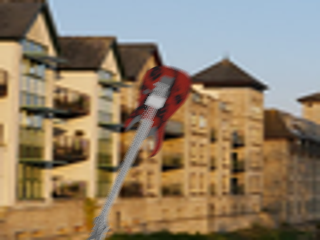}}
	   &\frame{\includegraphics[width=\linewidth]{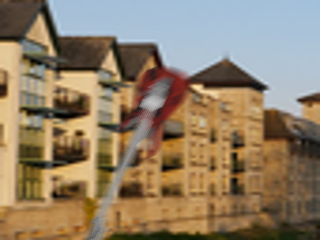}}
      &\frame{\includegraphics[width=\linewidth]{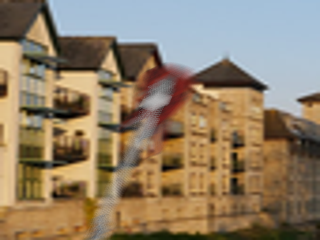}}
      &\frame{\includegraphics[width=\linewidth]{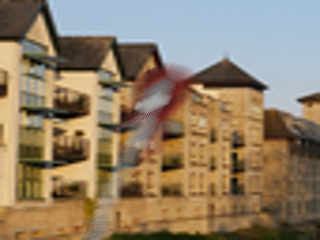}}
      &\frame{\includegraphics[width=\linewidth]{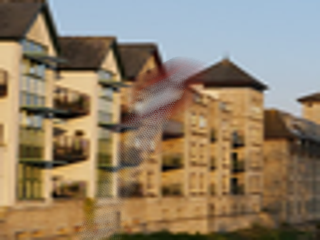}}
		\\
  \frame{\includegraphics[width=\linewidth]{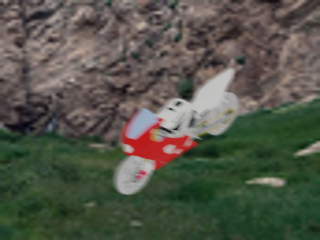}}
	   &\frame{\includegraphics[width=\linewidth]{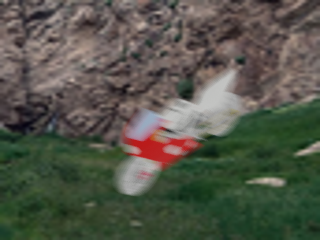}}
	   &\frame{\includegraphics[width=\linewidth]{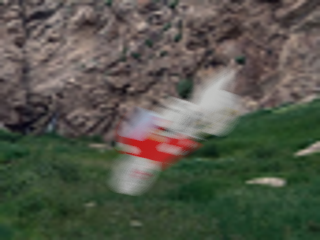}}
      &\frame{\includegraphics[width=\linewidth]{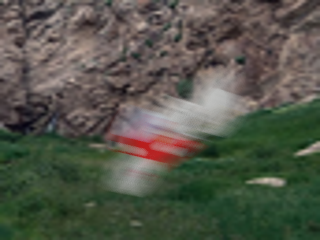}}
      &\frame{\includegraphics[width=\linewidth]{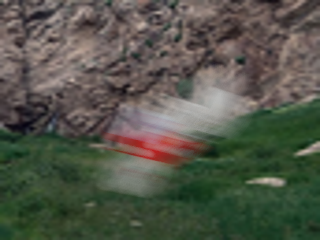}}
      &\frame{\includegraphics[width=\linewidth]{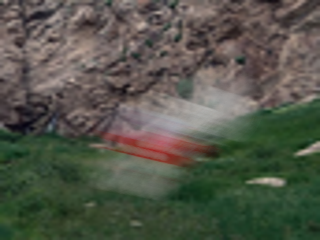}}
		\\
  \frame{\includegraphics[width=\linewidth]{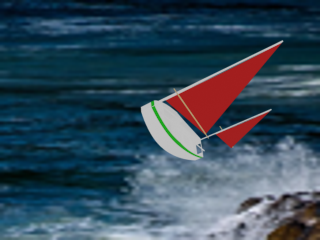}}
	   &\frame{\includegraphics[width=\linewidth]{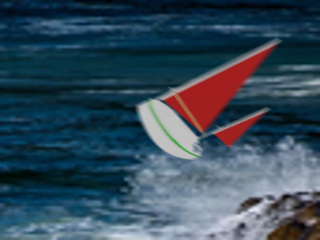}}
	   &\frame{\includegraphics[width=\linewidth]{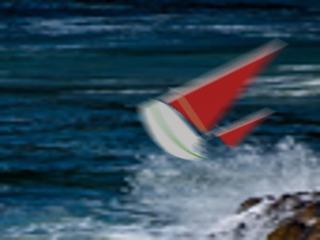}}
      &\frame{\includegraphics[width=\linewidth]{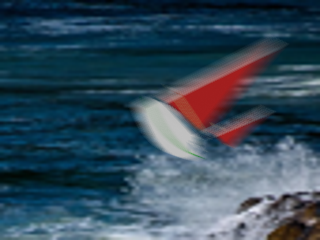}}
      &\frame{\includegraphics[width=\linewidth]{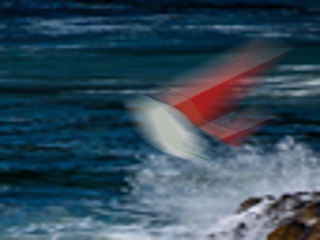}}
      &\frame{\includegraphics[width=\linewidth]{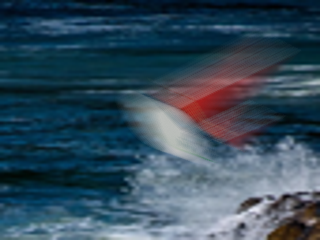}}
		\\
  \frame{\includegraphics[width=\linewidth]{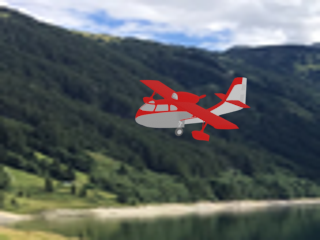}}
	   &\frame{\includegraphics[width=\linewidth]{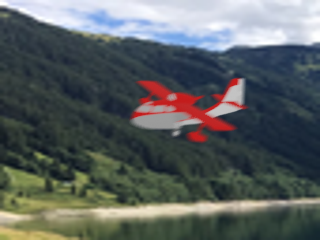}}
	   &\frame{\includegraphics[width=\linewidth]{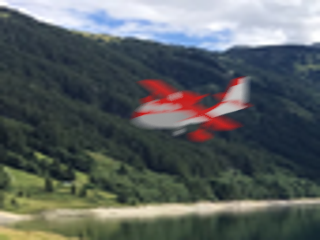}}
      &\frame{\includegraphics[width=\linewidth]{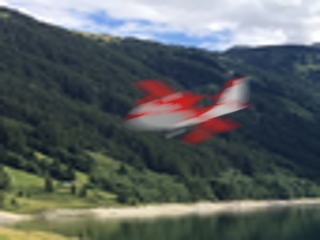}}
      &\frame{\includegraphics[width=\linewidth]{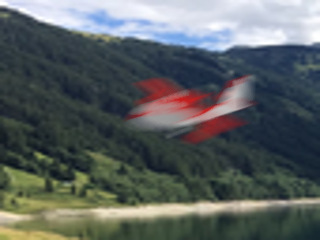}}
      &\frame{\includegraphics[width=\linewidth]{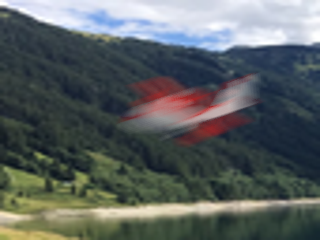}}
		\\

	\end{tabular}
    \vspace{-1ex}
\caption{\textbf{Synthetic dataset.} 
}
	\label{fig:syn_data_examples}
 \end{subfigure}
 \hfill
 \begin{subfigure}{0.48\linewidth} 

    \setlength{\tabcolsep}{2pt}
	\begin{tabular}{
	M{\figWidth}
	M{\figWidth}
	M{\figWidth}
    M{\figWidth}
    M{\figWidth}
    M{\figWidth}
    }
		BL\textsuperscript{r} 1 & BL\textsuperscript{r} 2 & BL\textsuperscript{r} 3 & BL\textsuperscript{r} 4 & BL\textsuperscript{r} 5 & BL\textsuperscript{r} 6
		\\
		
  \frame{\includegraphics[width=\linewidth]{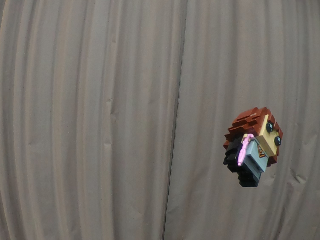}}
	   &\frame{\includegraphics[width=\linewidth]{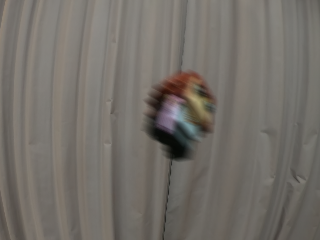}}
	   &\frame{\includegraphics[width=\linewidth]{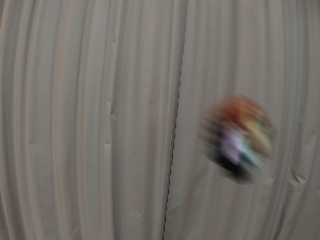}}
      &\frame{\includegraphics[width=\linewidth]{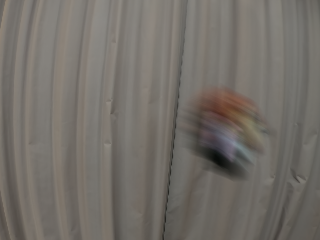}}
      &\frame{\includegraphics[width=\linewidth]{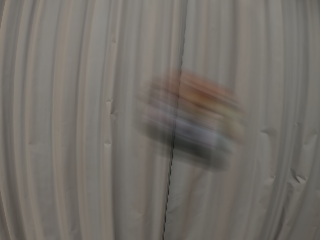}}
      &\frame{\includegraphics[width=\linewidth]{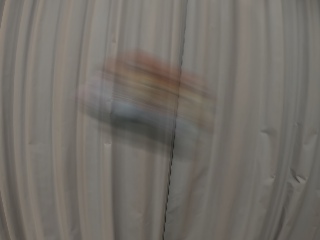}}
		\\
    \frame{\includegraphics[width=\linewidth]{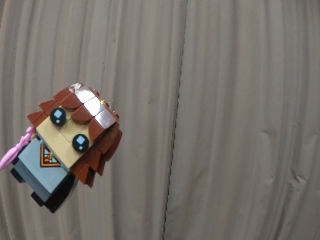}}
	   &\frame{\includegraphics[width=\linewidth]{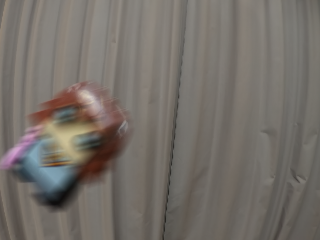}}
	   &\frame{\includegraphics[width=\linewidth]{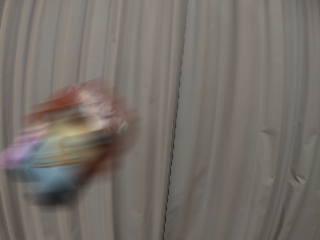}}
      &\frame{\includegraphics[width=\linewidth]{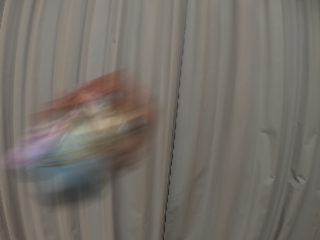}}
      &\frame{\includegraphics[width=\linewidth]{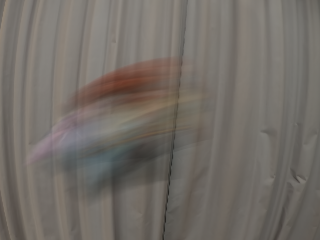}}
      &\frame{\includegraphics[width=\linewidth]{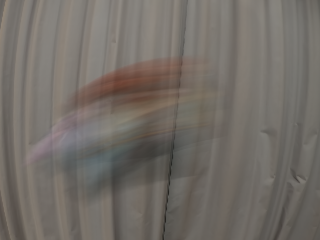}}
		\\
    \frame{\includegraphics[width=\linewidth]{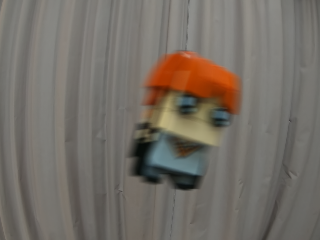}}
	   &\frame{\includegraphics[width=\linewidth]{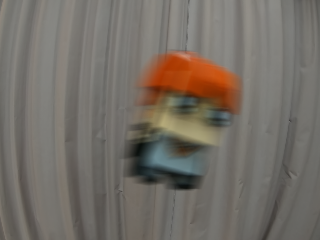}}
	   &\frame{\includegraphics[width=\linewidth]{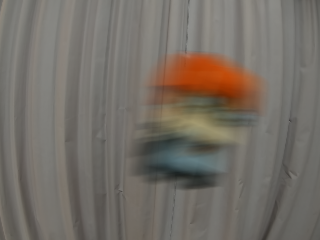}}
      &\frame{\includegraphics[width=\linewidth]{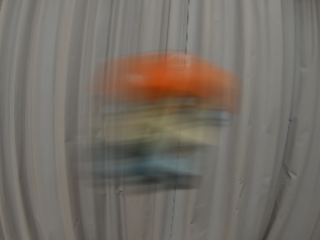}}
      &\frame{\includegraphics[width=\linewidth]{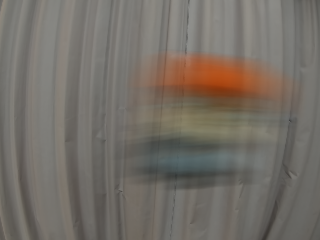}}
      &\frame{\includegraphics[width=\linewidth]{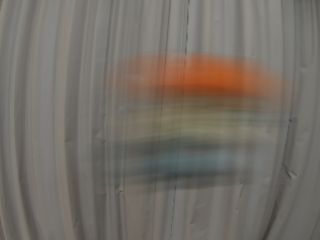}}
		\\
    \frame{\includegraphics[width=\linewidth]{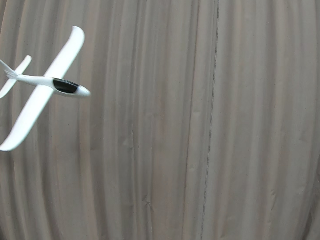}}
	   &\frame{\includegraphics[width=\linewidth]{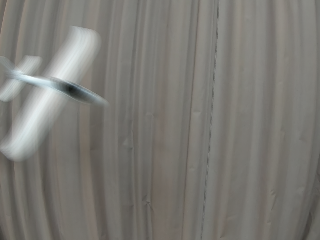}}
	   &\frame{\includegraphics[width=\linewidth]{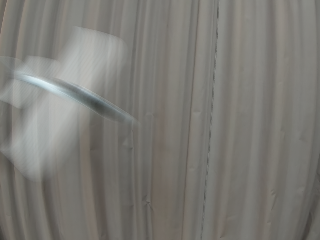}}
      &\frame{\includegraphics[width=\linewidth]{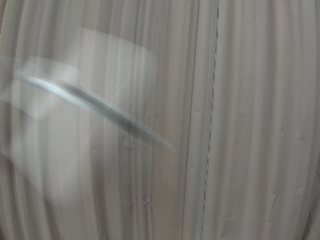}}
      &\frame{\includegraphics[width=\linewidth]{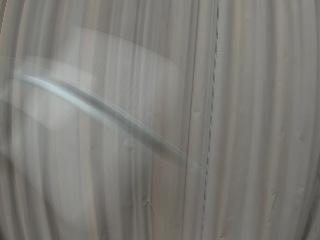}}
      &\frame{\includegraphics[width=\linewidth]{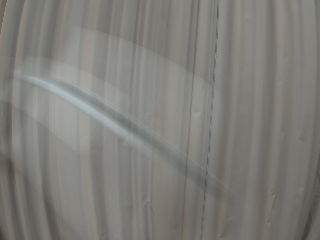}}
		\\
   \frame{\includegraphics[width=\linewidth]{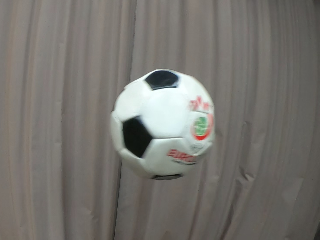}}
	   &\frame{\includegraphics[width=\linewidth]{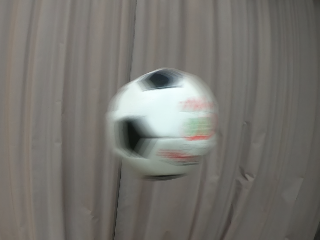}}
	   &\frame{\includegraphics[width=\linewidth]{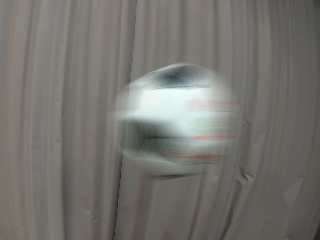}}
      &\frame{\includegraphics[width=\linewidth]{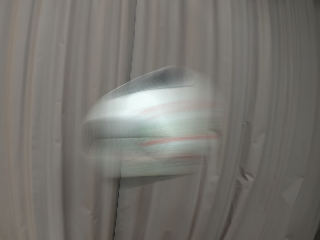}}
      &\frame{\includegraphics[width=\linewidth]{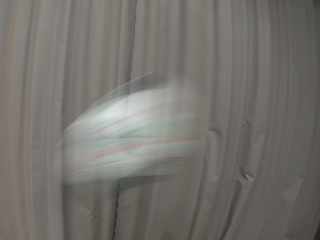}}
      &\frame{\includegraphics[width=\linewidth]{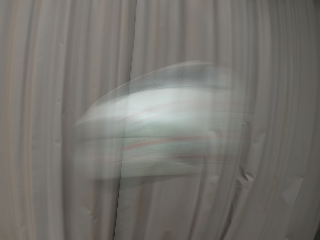}}
		\\
 \frame{\includegraphics[width=\linewidth]{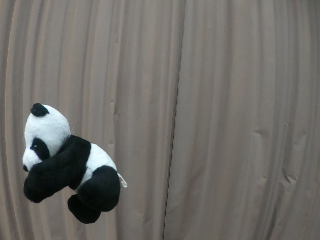}}
	   &\frame{\includegraphics[width=\linewidth]{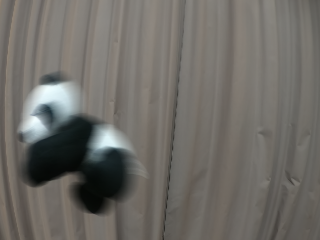}}
	   &\frame{\includegraphics[width=\linewidth]{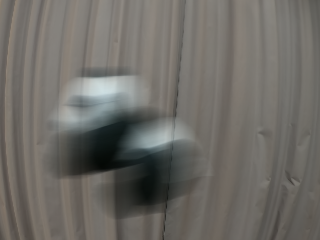}}
      &\frame{\includegraphics[width=\linewidth]{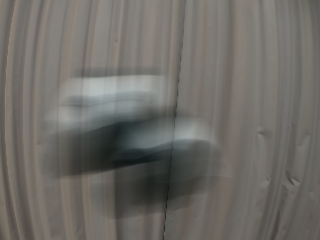}}
      &\frame{\includegraphics[width=\linewidth]{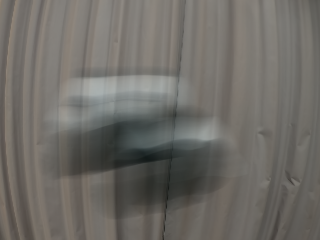}}
      &\frame{\includegraphics[width=\linewidth]{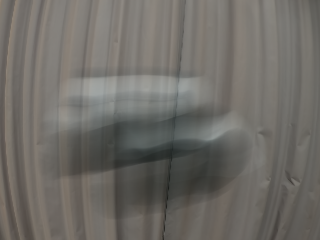}}
		\\

	\end{tabular}
    \vspace{-1ex}
\caption{\textbf{Real-world dataset.}}
	\label{fig:real_data_examples}
 \end{subfigure}
\vspace{-0.8em}
\caption{\textbf{The introduced synthetic (a) and real-world (b) datasets.}
Rows 1-2: different trajectories of the same object. 
Rows 2-3: two different objects from the same category with similar shapes. 
Rows 4-6: objects from different categories but share similar textures.
Columns correspond to different blur levels, from 1 to 6, and they are different for synthetic (BL) and real-world (BL\textsuperscript{r}) datasets.
}
 
\label{fig:data_examples}

\vspace{-10px}\end{figure*}

\boldparagraph{Blur severity definition.}
Our definition of blur severity is inspired by the image generation process. 
The formation of an image $\mathbf{I}$ containing a motion-blurred object in front of a background can be expressed~\cite{fmo} as
\begin{equation}
\label{eq:matting}
\mathbf{I} = \mathbf{P} * \mathbf{O} + (1-\mathbf{P}*\mathbf{M})\mathbf{B}
\end{equation}
where the first term denotes the convolution ($*$) of a point spread function $\mathbf{P}$ corresponding to a motion trajectory and the object's sharp appearance $\mathbf{O}$, and the second term depicts the contribution of the background $\mathbf{B}$, in which $\mathbf{M}$ is the binary mask of $\mathbf{O}$. 
Let 
\begin{equation}
\label{eq:alpha}
    \boldsymbol{\alpha} = \mathbf{P}*\mathbf{M},
\end{equation}
$\boldsymbol{\alpha} \in \mathbb{R}^{H \times W}$, $\boldsymbol\alpha(i,j) \in [0, 1], \forall (i,j) \in \mathbf{I}.$  
Considering only objects with opaque materials, $\boldsymbol{\alpha}$ represents the visibility of the object. 
In other words, it can characterize the amount of blur. 
Generally, the sharper the object in the image, the larger the $\boldsymbol{\alpha}$ values are. 
However, at the edge pixels, the object blends with the background, and the alpha values of these edge pixels gradually fade to 0. 
Combining the alpha mask from Eq.~\eqref{eq:alpha} and considering edge effects, we define the \textit{Blur Severity} (BS) of an image with a motion-blurred object as:
\begin{equation}
\label{eq:bs}
BS = 1- \frac{\sum_{i=1}^{H} \sum_{j=1}^{W} \boldsymbol{\beta}(i, j) \cdot\boldsymbol{\alpha}(i,j) }{\sum_{i=1}^{H} \sum_{j=1}^{W} \boldsymbol{\beta}(i, j)},
\end{equation}
where $\boldsymbol{\beta}$ is a thresholded (by 0) and eroded (by 3 pixels) binary mask of $\boldsymbol{\alpha}$.
Erosion effectively removes the adverse contribution of the edge pixels to the blur severity measure.
As a result, $BS \in (0, 1)$, and the larger the value, the more severe the motion blur in the image.

\begin{table}[t] 
    \centering
    \begin{adjustbox}{max width=\linewidth}
    \setlength{\tabcolsep}{7.5pt} 
    \begin{tabular}{lc c cccccc}
        \toprule
        \multirow{2}{*}{Dataset} & {\#Images}  && \multicolumn{6}{c}{\#Images by BL (Syn.) or BL\textsuperscript{r} (Real)}  \\ \cline{4-9}
         &Total   && 1 &  2&  3&  4&  5 &  6\\
        \midrule
        Syn. Q &20,995&& 4,288 & 3,932 & 4,078 & 4,089 & 2,930 & 1,678  \\
        Syn. DB & 91,621 && 18,871 & 17,508 & 17,888 & 18,029 & 12,546 & 67,79 \\
        Syn. Dist. & 1,091,939 && 214,364 & 177,869 & 222,542 & 235,662 & 149,828 & 91,674 \\
        \midrule
        Real Q & 2,753 && 612 & 620 & 561 & 396 & 315 & 249\\ 
        Real DB &10,340 &&1,923 &1,803& 2,080& 1,745& 1,375& 1,414\\
        \bottomrule
    \end{tabular}
    \end{adjustbox}
\caption{Distribution of images in our synthetic and real datasets, sorted by blur levels. 'Q' stands for 'query', 'DB' denotes database, and 'Dist.' represents distractors.}
    \label{tab:distribution}

\vspace{-15px}\end{table} 

\boldparagraph{Blur level-based sample selection.} 
Since blur severity is a continuous quantity, finding images within a certain blur severity range for each query image is expensive. We thus use a discretized version of blur severity — \textit{Blur Level} (BL):
\begin{equation}
\label{eq:bl}
    BL = \lceil10 {BS}\rceil, 
\end{equation}
where $\lceil \cdot \rceil$ denotes ceiling operation. 
Each image is assigned to a $BL$ according to its $BS$. 
During training, we select contrastive samples based on the query blur level $b_q$ and the specified contrastive $BL$ range $r$. The selected $N_p$ positive and $N_n$ negative samples thus have blur levels within the range of $b_q-r$ to $b_q+r$.

\subsection{Training objectives}

\boldparagraph{Blur estimation loss.}
Blur severity contains information about the visibility of the blurred object in an image (Sec.~\ref{sec:bliss}). 
From another perspective, it reflects a signal-to-noise ratio (SNR), since the more severe the motion blur, the greater the ratio of background (noise) to foreground (signal).
In view of this, we define blur estimation loss as
\begin{equation}
\label{eq:be_loss}
    \mathcal{L}_{be} = |p -(1-BS)| \, ,
\end{equation}
where $p \in \mathbb{R}$ is the blur value predicted by the blur estimation head. 
By constraining the estimate of the continuous $BS$ (Eq.~\eqref{eq:bs}), we implicitly force the network to identify blur and estimate the SNR, which helps it to focus on and extract signal information in the blurred region while ignoring background noises.

\boldparagraph{Localization loss.} This loss is defined as 
\begin{equation}
    \mathcal{L}_{loc} = |x - \hat{x}| + |y - \hat{y}| + |h - \hat{h}| + |w - \hat{w}| \, ,
\end{equation}
where $(x,y,w,h)$ and $(\hat{x},\hat{y},\hat{w},\hat{h})$ represent the normalized ground truth bounding box and the predicted one from the localization head, respectively. 
Localization loss requires the network to distinguish motion-blurred objects from backgrounds under various blur conditions, 
forcing it to recognize blur, identify objects at different blur levels, and further, focus more on the foreground objects.

\boldparagraph{Classification loss.} Following DELG~\cite{delg}, we apply ArcFace margin loss~\cite{arcface} to the descriptor $\mathbf{D}$, expressed as 
\begin{equation}
     \mathcal{L}_{cls} = -\log\left(\frac{\exp\left(\gamma \times AF(\mathbf{\hat{w}}^{T}_{t}\mathbf{\hat{D}}, 1)\right)} {\sum_{n} \exp\left(\gamma \times AF(\mathbf{\hat{w}}^{T}_{n}\mathbf{\hat{D}}, y_{n})\right)}\right) \, ,
\end{equation}
where $\gamma$ is a scale factor, $\mathbf{\hat{w}_{i}}$ refers to the $i$-th row of the $L_{2}$ normalized $N$-class classifier $\mathbf{\hat{W}} \in \mathbb{R}^{d \times N}$, $\mathbf{\hat{D}}$ denotes the $L_{2}$ normalized $\mathbf{D}$. $y$ is the one-hot label, and $t$ represents the ground truth class index ($y_{t}=1$). $AF$ is the ArcFace-adjusted cosine similarity and it can be expressed as 
\begin{equation}
    AF(s, g) = \begin{cases}
    \cos(\arccos(s) + m), & \text{if } g = 1 \\
    s, & \text{if } g = 0
    \end{cases},
\end{equation}
where $s$ is the cosine similarity, $m$ is the ArcFace margin, and $g \in \{0, 1\}$. $g = 1$ indicates the ground truth class. 
The classification loss requires the model to group sharp or motion-blurred images of the same object together, while distinguishing between different objects, which may appear similar, especially when blurred. This implicitly forces the model to focus on blur-invariant features.

\boldparagraph{Contrastive loss.} We employ contrastive learning \cite{chopra2005contrastive} to construct a descriptor space, where the distance between similar samples is minimized, and between dissimilar samples maximized. The contrastive loss is applied to $(query, positive)$ and $(query, negative)$ descriptor pairs:
\begin{equation}
    \mathcal{L}_{con}(i,j) = \begin{cases}
        0.5 \cdot D^{2}_{i,j}, & \text{if } y(i,j) = 1 \\
        0.5(\text{max}\{0, \tau - D_{i,j}\})^{2} ,& \text{if } y(i,j) = 0
    \end{cases},
\end{equation}
where $D_{i,j} = ||{\mathbf{\hat{D}}(i)-\mathbf{\hat{D}}(j)}||$ and $\mathbf{\hat{D}}(i)$ is the $L_{2}$ normalized descriptor of image $i$. $\tau$ is a margin parameter, and $y(i,j)$ is binary indicating whether the pair matches $(=1)$.
The contrastive loss further enhances the focus on blur-invariant features by explicitly forcing descriptors of the same object, whether sharp or blurred, to be close together, and those of different objects to be far apart.

\boldparagraph{Joint loss.} 
The joint loss is a weighted sum of all previously defined losses:
\begin{equation}
    \mathcal{L}_{joint} = \mathcal{L}_{con} + \alpha_{cls} \mathcal{L}_{cls} + \alpha_{be} \mathcal{L}_{be} + \alpha_{loc} \mathcal{L}_{loc}. 
\end{equation}

\vspace{-0.6em}
\begin{table*}[t]
    \notsotiny 
    \centering
    \begin{adjustbox}{max width=\textwidth}
    
    \setlength{\tabcolsep}{1.4pt} 
    \begin{tabularx}{\textwidth}{l ccccccc c ccccccc}
        \toprule
        \multirow{2}{*}{Method} & \multicolumn{7}{c}{mAP by query BL (synthetic)}  && \multicolumn{7}{c}{mAP by query BL (synthetic + 1M)} \\  \cline{2-8} \cline{10-16}
         & All & 1 & 2 & 3 & 4 & 5 & 6
         && All & 1 & 2 & 3 & 4 & 5 & 6 \\
        \midrule
        DELG \cite{delg}
        & 81.65 
        & 83.55 
        & 84.96 
        & 84.89 
        & 82.69 
        & 77.35 
        & 66.20 
        &
        & 68.19 
        & 73.64 
        & 75.40 
        & 73.34 
        & 68.05 
        & 58.28 
        & 42.46
        \\
        
        DOLG \cite{dolg} 
        & 83.03 
        & 85.10 
        & 86.26 
        & 85.93 
        & 84.04 
        & 78.78 
        & 68.07 
        &
        & 69.97 
        & 75.75  
        & 77.47 
        & 75.01  
        & 70.10 
        & 60.01 
        & 42.49 
        \\
        
        Token \cite{token} 
        & {84.84} 
        & {86.53} 
        & {87.99} 
        & {87.50} 
        & {85.81} 
        & {80.70} 
        & {71.57} 
        &
        & 70.65  
        & 75.32  
        & 77.66  
        & 75.51  
        & 70.24
        & 61.19 
        & 48.05
        \\
         
        Ours-sharp 
        & 27.57 
        & 43.06 
        & 38.53 
        & 29.74 
        & 19.40 
        & 10.87 
        & 6.14 
        &
        & 32.64 
        & 71.93 
        & 43.88 
        & 27.18 
        & 15.41 
        & 7.94 
        & 4.27  
        \\

        Ours 
        &\textbf{91.78} 
        & \textbf{93.05} 
        & \textbf{93.48}  
        & \textbf{93.14} 
        & \textbf{92.25} 
        & \textbf{90.20} 
        & \textbf{82.86} 
        &
        & \textbf{84.09} 
        & \textbf{88.74} 
        & \textbf{89.56} 
        & \textbf{87.68} 
        & \textbf{84.41} 
        & \textbf{76.89} 
        & \textbf{62.42} 
        \\
        \bottomrule
    \end{tabularx}
    \end{adjustbox}
    \caption{\textbf{Retrieval results} on the synthetic dataset with (+1M) and without distractors. Ours-sharp is our model trained with only sharp images. 'All' denotes the overall performance for queries of all blur levels. All methods are re-implemented and re-trained on our synthetic training set. The database contains images of all blur levels. 
    } 
    \label{tab:mAP_db_mixed}
    \vspace{-0.8em} 
\vspace{-10px}\end{table*}
\section{Dataset}

\label{sec:dataset}
We create novel synthetic and real datasets for training and evaluating retrieval methods for motion-blurred objects.
The synthetic dataset serves for both training and evaluation.
We also generate a large-scale synthetic distractor set to evaluate methods in a more challenging setup. 
The novel real-world dataset is specifically constructed to validate the methods' effectiveness in real scenarios. 
All datasets are meticulously processed for the blur retrieval task, laying a crucial foundation for future research.
Statistics of these datasets are shown in Table~\ref{tab:distribution}.

\begin{table}[t] 
    \centering
    \begin{adjustbox}{max width=\linewidth}
    \setlength{\tabcolsep}{7.5pt} 
    \begin{tabular}{lc c cccccc}
        \toprule
        \multirow{2}{*}{Method} & {mAP}  && \multicolumn{6}{c}{mAP by Query BL\textsuperscript{r}}  \\ \cline{4-9}
         &All   && 1 &  2&  3&  4&  5 &  6\\
        \midrule
        DELG \cite{delg} & 54.82 && 49.13 & 63.43 & 57.25 & 55.01 & 53.77 & 42.92  \\
        DOLG \cite{dolg}& 54.64 && 43.93 & 60.59 & 58.36 & 59.06 & 58.58 & 45.78 \\
        Token \cite{token} & 43.33 && 38.71 & 47.08 & 50.79 & 46.44 & 42.71 & 24.43\\ 
        Ours-sharp & 40.24 && 49.55 & 45.02 & 41.33 & 33.23 & 29.40 & 27.91\\
        Ours & \textbf{62.88} && \textbf{57.50} & \textbf{70.38} & \textbf{66.77} & \textbf{63.18} & \textbf{64.48} & \textbf{46.14}\\
        \bottomrule
    \end{tabular}
    \end{adjustbox}
    \caption{\textbf{Retrieval results} on the real-world dataset (trained on synthetic without fine-tuning). The database includes all blur levels. Our method outperforms all others.}
    \label{tab:mAP_real}
    \vspace{-1.2em} 

\vspace{-10px}\end{table}

\subsection{Synthetic dataset}
The images in this dataset contain 3D objects moving along different trajectories, in front of different backgrounds, and with different amounts of motion blur. 
Thus, this dataset poses many challenges to retrieval methods.

\boldparagraph{Backgrounds.}
Background images are sampled from the LHQ dataset~\cite{lhq}, cropped to a resolution of $240 \times 320$. 

\boldparagraph{Objects.}
We select 39 categories with the largest number of textured objects from the ShapeNet dataset~\cite{shapenet} and randomly sample 30 objects in each category.  

\boldparagraph{Trajectories.}
For each object, we generate 120 random linear trajectories, each divided into 23 sub-trajectories. We then select up to 10 consecutive sub-trajectories to create images featuring objects with diverse levels of motion blur. 
\boldparagraph{Rendering.}
We render blurry images with Blender Cycles~\cite{blender} by capturing moving objects during the camera exposure time.
Additionally, we render one sharp image per trajectory, placed at the central position along the trajectory.

\boldparagraph{Filtering.}
We discard renderings where the object occupies less than 1.5\% of the pixels or where the intersection-over-union between silhouette renderings at trajectory endpoints is below 20\%. The former condition suggests the object is too small, while the latter indicates the appearance changes are too drastic.

\boldparagraph{Ground truth.}
For each generated image, we estimate the bounding box coordinates by detecting the non-zero pixels in the alpha channel. 
We calculate the blur level for each image based on \Eq \eqref{eq:bl}.

\boldparagraph{Retrieval dataset.}
In total, we rendered 1.5 million images for 1138 different instances (examples in \Fig \ref{fig:syn_data_examples}).
Additionally, we create a challenging distractor set with over 1 million images (examples in the supplementary material), using the same object categories as before
to increase the difficulty of the distractors in terms of intra-class similarity. 
We randomly select 40 objects per category, excluding the 30 previously chosen objects, and generate 70 trajectories for each.
For each trajectory, we render one sharp image and 10 blurry images with varying degrees of blur.
Backgrounds are also selected from a different subset of~\cite{lhq}.

\begin{table*}[t]
    \scriptsize	
    \centering
    \begin{adjustbox}{max width=\textwidth}
    \setlength{\tabcolsep}{1.8pt} 
    \begin{tabularx}{\textwidth}{P{0.5cm} c ccccccc c ccccccc}
        \toprule
        {BL}  && \multicolumn{7}{c}{DELG~\cite{delg} (range: 57.79; std: 16.75)} && \multicolumn{7}{c}{DOLG~\cite{dolg} (range: 56.81; std: 16.93)} \\  \cline{3-9} \cline{11-17}
          \diagbox[innerwidth = 0.5cm, height=0.5cm]{D}{Q} &&All 
         &1  &2 &3  &4 &5 &6 &&All&  1 &  2&  3&  4&  5 & 6\\
         \midrule

        1 && 75.99 & 86.73 & 84.17 & 80.26 & 72.48 & 63.12 & 50.08
  && 77.78 & 87.01 & 85.70 & 81.85 & 75.11 & 66.20 & 52.39\\
2 && 77.35 & 82.74 & 86.22 & 82.87 & 76.68 & 66.46 & 50.00
&& 78.44 & 84.18 & 87.00 & 83.43 & 77.91 & 67.88 & 51.31\\
3 && 77.19 & 79.62 & 83.43 & 82.94 & 78.46 & 69.43 & 52.86
&& 78.49 & 81.84 & 84.74 & 83.85 & 79.69 & 70.58 & 53.12
\\
4 && 74.96 & 74.26 & 78.91 & 79.85 & 78.05 & 70.89 & 55.12
&& 76.40 & 77.27 & 80.55 & 80.80 & 79.10 & 71.73 & 55.28
\\
5 && 62.71 & 57.86 & 61.72 & 64.37 & 65.36 & 68.96 & 56.05
&& 64.19 & 61.31 & 63.68 & 65.38 & 66.05 & 69.68 & 55.75
\\
6 && 34.89 & 29.56 & 28.94 & 31.75 & 34.06 & 45.37 & 53.86
&& 35.51 & 31.24 & 30.20 & 32.29 & 34.18 & 45.00 & 53.41
\\

        \midrule
        {BL}  && \multicolumn{7}{c}{Token \cite{token} (range: 55.72; std: 16.11)} && \multicolumn{7}{c}{Ours (range: \textbf{53.68}; std: \textbf{15.65})} \\  \cline{3-9} \cline{11-17}
          \diagbox[innerwidth = 0.5cm, height=0.5cm]{D}{Q} &&All 
         &1  &2 &3  &4 &5 &6 &&All&  1 &  2&  3&  4&  5 & 6\\
         \midrule

        1&& 79.65 & 88.22 & 87.34 & 83.50 & 76.87 & 68.51 & 56.63
&& \textbf{88.57} & \textbf{95.09} & \textbf{94.24} & \textbf{91.54} & \textbf{87.19} & \textbf{80.91} & \textbf{68.15}\\
2&& 79.97 & 84.78 & 88.48 & 84.77 & 78.99 & 70.01 & 55.82
&& \textbf{88.75} & \textbf{93.07} & \textbf{94.65} & \textbf{92.18} & \textbf{88.77} & \textbf{82.03} & \textbf{67.18}\\
3&& 79.50 & 81.79 & 85.46 & 84.75 & 80.21 & 72.09 & 58.18
&& \textbf{88.76} & \textbf{91.70} & \textbf{93.18} & \textbf{92.32} & \textbf{89.83} & \textbf{83.41} & \textbf{68.94}\\
4&& 76.82 & 76.56 & 80.42 & 81.24 & 79.31 & 72.61 & 59.61
&& \textbf{87.16} & \textbf{88.12} & \textbf{90.11} & \textbf{90.09} & \textbf{89.31} & \textbf{84.37} & \textbf{70.35}\\
5&& 65.39 & 61.85 & 64.49 & 66.31 & 67.14 & 70.96 & 60.36
&& \textbf{76.28} & \textbf{73.87} & \textbf{74.90} & \textbf{76.04} & \textbf{77.39} & \textbf{83.06} & \textbf{71.67}\\
6&& 37.98 & 33.23 & 32.76 & 34.61 & 36.50 & 47.56 & 57.43
&& \textbf{47.31} & \textbf{43.43} & \textbf{41.41} & \textbf{42.89} & \textbf{44.55} & \textbf{58.13} & \textbf{69.59}\\

        \bottomrule
        
    \end{tabularx}
    
    \end{adjustbox}
\caption{\textbf{Retrieval results} by query (Q) and database (D) blur level (BL) on our synthetic dataset (with 1M distractors). 
'All' denotes the overall performance for queries of all blur levels. 
The standard deviation and range of mAP values are displayed alongside each method.
The best scores under the same settings are marked in bold. 
}
    \label{tab:mAP_db_pure_with_distractors}

    \vspace{-1.2em} 

\vspace{-10px}\end{table*} 

\subsection{Real-world dataset}
We record videos with a GoPro 7 Hero camera at 240 fps with full exposure.
We use 35 different objects and move each object 3 to 5 times along different trajectories in front of the camera. 
None of the real-world objects are in synthetic data and 2/3 of which are from new categories not present in synthetic data. This better tests the generalization ability of models to new objects and categories.
In total, we recorded 139 videos. 
Different amounts of blur are generated by averaging consecutive frames, ranging from just one up to 200 averaged frames.

Computing blur level based on Eq.~\eqref{eq:bl} is not feasible for the real-world dataset. 
Therefore, we manually annotate the data, assigning each image to one of 6 blur levels based on the perceived blur, and we use the ratio of the trajectory length to the average object size as a reference. 
The annotation enables a more detailed analysis and a more comprehensive evaluation of model performance.
Further annotation details can be found in the supplementary material.
Since the blur level for real data differs from that in Eq.~\eqref{eq:bl}, we denote it as BL\textsuperscript{r}.

The real-world dataset contains 13,093 images, with an average of 374 images for each object (examples in \Fig \ref{fig:real_data_examples}). 
Compared to the synthetic dataset, the object scale change is more drastic in different trajectories, \eg rows 1 \& 2. 
The trajectories are notably different from those in the synthetic dataset. They involve complex curved trajectories with acceleration and other physical forces, which poses an additional challenge for methods trained on our synthetic dataset.

\section{Experiments}
\label{sec:experiments}
\boldparagraph{Setup.} 
For each category of objects in the synthetic dataset, we randomly take 70\% (rounded down, same for validation) for training (792 objects), 15\% for validation (153 objects), and 15\% for testing (193 objects).
For the training objects, we take one image for each trajectory while ensuring a balanced distribution of images across different blur levels. 
In total, there are  95k images in the training set.
For each of the test objects, we sample 20 trajectories to form a query set, and the rest is used for the database. 
The database images are sampled from all blur levels unless specified otherwise (for per blur level database experiments).
We additionally show performance on a subset of queries for a given blur level.
In all experiments (except Ours-sharp), only one model is trained for all blur levels, and the evaluation is performed for varying blur levels.
\global\long\def\figWidth{0.077\linewidth}
\begin{figure*}[t]
	\centering
 \begin{subfigure}{\linewidth}
    \setlength{\tabcolsep}{2pt}
    \setlength{\fboxrule}{1pt} 
    \setlength{\fboxsep}{0pt} 
	\begin{tabular}{
	 P{\figWidth}
	P{0.2cm}
	P{\figWidth}
	P{\figWidth}
    P{\figWidth}
    P{\figWidth}
    P{\figWidth}
    P{\figWidth}
	P{\figWidth}
    P{\figWidth}
    P{\figWidth}
    P{\figWidth}
    }
	\multicolumn{2}{c}{Query}	&   \multicolumn{10}{c}{Top 20 retrieval results}
		\\
\raisebox{0em}{\multirow{2}{*}{
\addimgTtext{\includegraphics[width=\linewidth]{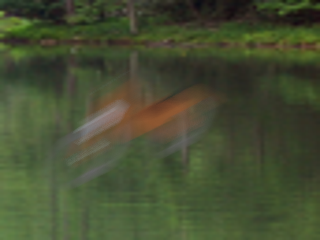}}{6}}}
&\raisebox{1.1em}{\multirow{2}{*}{\rotatebox[origin=c]{90}{Token~\cite{token}}}}
    &\addimgTtext{\includegraphics[width=\linewidth]{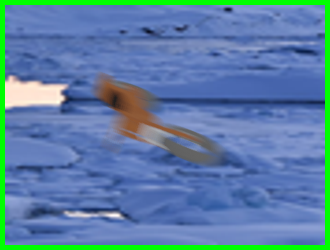}}{3}
      &\addimgTtext{\includegraphics[width=\linewidth]{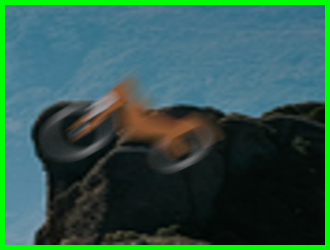}}{5}
      &\addimgTtext{\includegraphics[width=\linewidth]{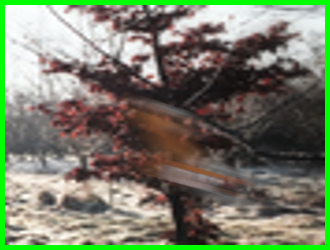}}{6}
      &\addimgTtext{\includegraphics[width=\linewidth]{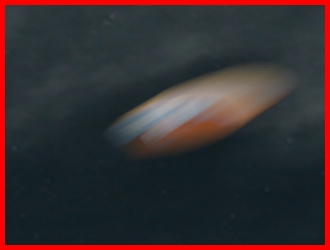}}{4}
      &\addimgTtext{\includegraphics[width=\linewidth]{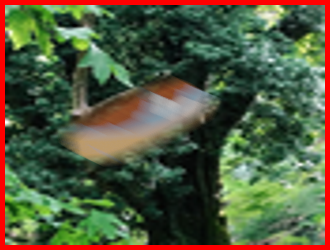}}{3}
      &\addimgTtext{\includegraphics[width=\linewidth]{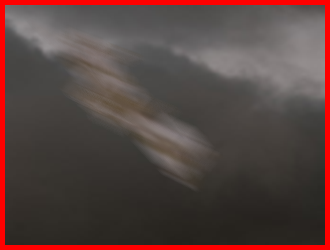}}{4}
      &\addimgTtext{\includegraphics[width=\linewidth]{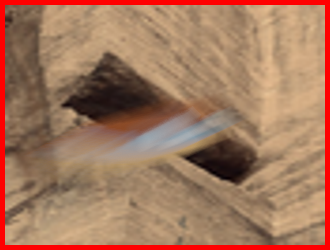}}{4}
      &\addimgTtext{\includegraphics[width=\linewidth]{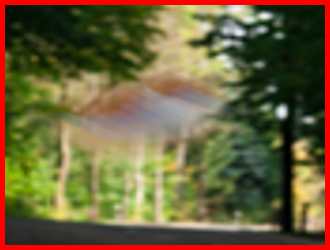}}{5}
      &\addimgTtext{\includegraphics[width=\linewidth]{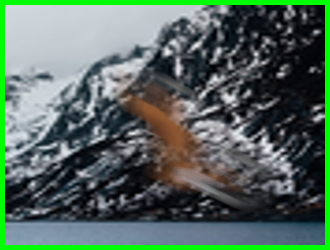}}{6}
      &\addimgTtext{\includegraphics[width=\linewidth]{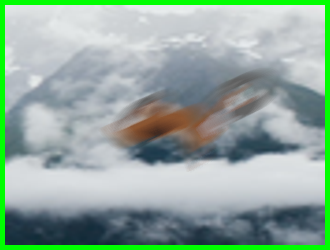}}{5}
		\\
    &&\addimgTtextblack{\includegraphics[width=\linewidth]{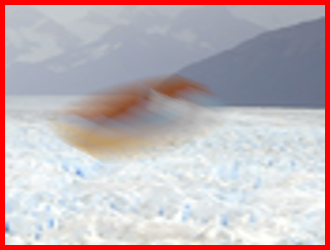}}{4}
      &\addimgTtext{\includegraphics[width=\linewidth]{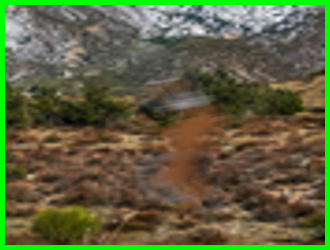}}{6}
      &\addimgTtext{\includegraphics[width=\linewidth]{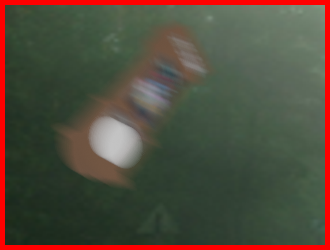}}{3}
      &\addimgTtext{\includegraphics[width=\linewidth]{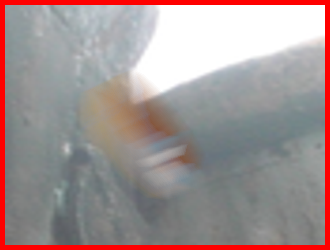}}{5}
      &\addimgTtext{\includegraphics[width=\linewidth]{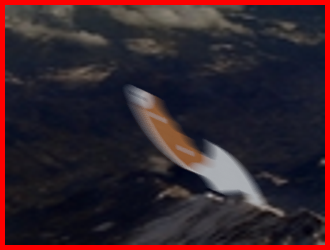}}{3}
      &\addimgTtext{\includegraphics[width=\linewidth]{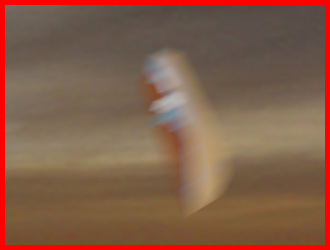}}{4}
      &\addimgTtextblack{\includegraphics[width=\linewidth]{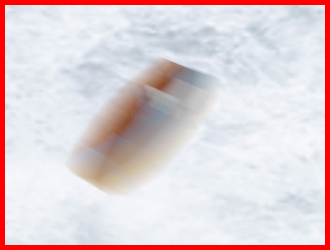}}{4}
      &\addimgTtext{\includegraphics[width=\linewidth]{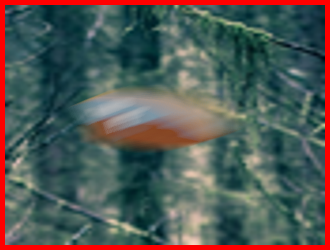}}{3}
    &\addimgTtext{\includegraphics[width=\linewidth]{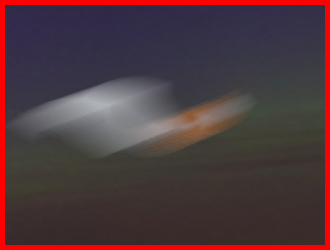}}{6}
     &\addimgTtext{\includegraphics[width=\linewidth]{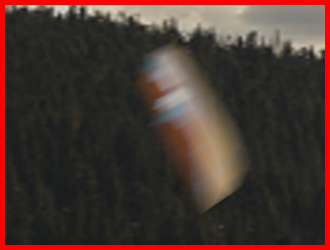}}{5} \\
\raisebox{0em}{\multirow{2}{*}{
\addimgTtext{\includegraphics[width=\linewidth]{fig/synthetic/qualitativeBL6_motor/query_image_16828.png}}{6}}}
&\raisebox{0.10em}{\multirow{2}{*}{\rotatebox[origin=c]{90}{Ours}}}
    &\addimgTtext{\includegraphics[width=\linewidth]{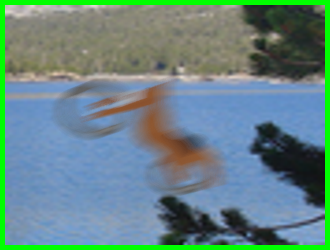}}{5}
      &\addimgTtext{\includegraphics[width=\linewidth]{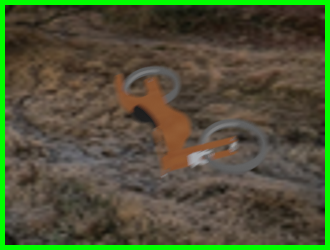}}{3}
      &\addimgTtext{\includegraphics[width=\linewidth]{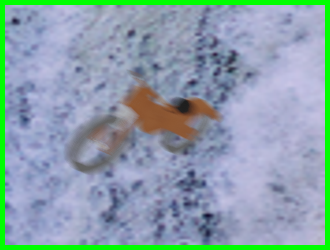}}{4}
      &\addimgTtext{\includegraphics[width=\linewidth]{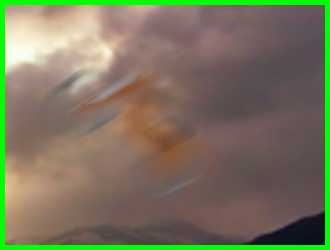}}{6}
      &\addimgTtext{\includegraphics[width=\linewidth]{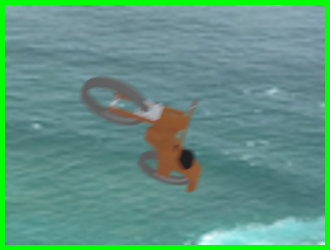}}{2}
      &\addimgTtext{\includegraphics[width=\linewidth]{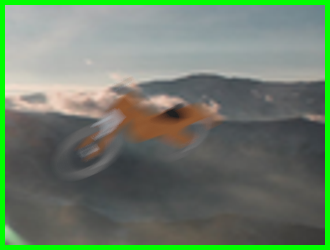}}{4}
      &\addimgTtext{\includegraphics[width=\linewidth]{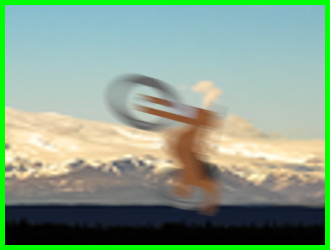}}{5}
      &\addimgTtext{\includegraphics[width=\linewidth]{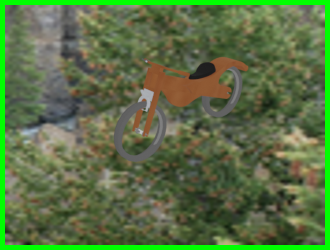}}{1}
      &\addimgTtext{\includegraphics[width=\linewidth]{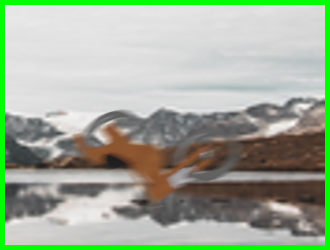}}{4}
      &\addimgTtext{\includegraphics[width=\linewidth]{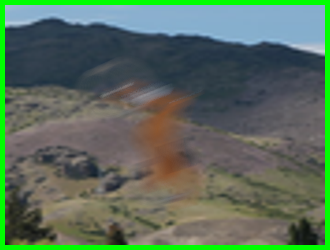}}{6}
		\\
    &&\addimgTtext{\includegraphics[width=\linewidth]{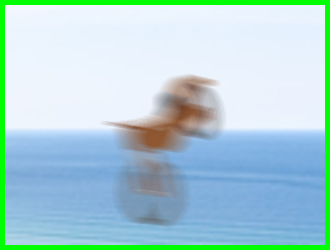}}{6}
	   &\addimgTtext{\includegraphics[width=\linewidth]{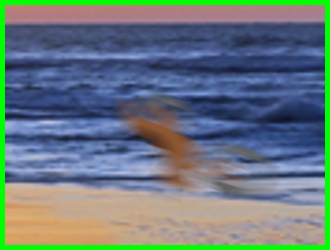}}{5}
      &\addimgTtext{\includegraphics[width=\linewidth]{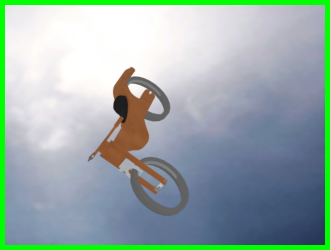}}{1}
      &\addimgTtext{\includegraphics[width=\linewidth]{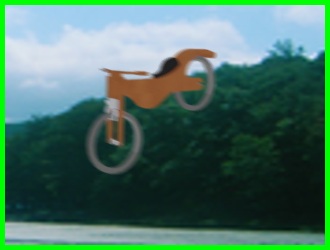}}{3}
      &\addimgTtext{\includegraphics[width=\linewidth]{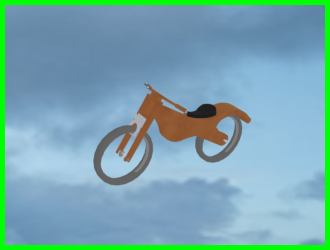}}{1}
      &\addimgTtext{\includegraphics[width=\linewidth]{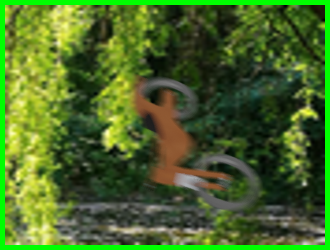}}{4}
      &\addimgTtext{\includegraphics[width=\linewidth]{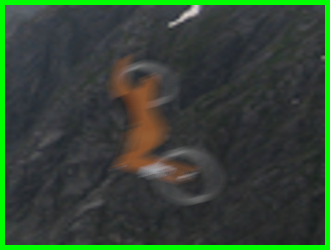}}{5}
      &\addimgTtext{\includegraphics[width=\linewidth]{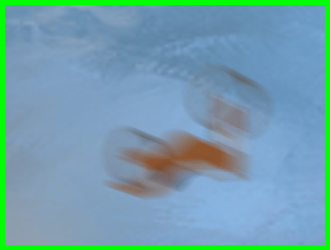}}{6}
      &\addimgTtext{\includegraphics[width=\linewidth]{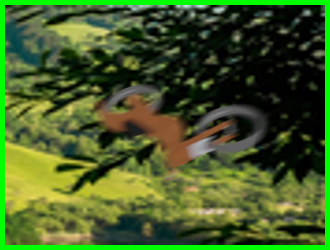}}{3}
        &\addimgTtext{\includegraphics[width=\linewidth]{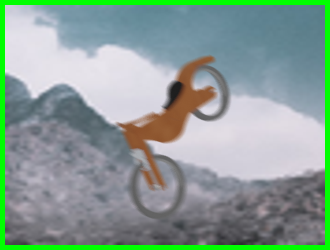}}{3}\\
	\end{tabular}
    \vspace{-1ex}
\caption{\textbf{Example of retrieval results on our synthetic dataset (with distractors).}}
	\label{fig:syn_qualitative_1}
 \end{subfigure}
 \hfill
\vspace{-0.8em}
\begin{subfigure}{\linewidth}
    
 \setlength{\tabcolsep}{2pt}
    \setlength{\fboxrule}{1pt} 
    \setlength{\fboxsep}{0pt}  
	\begin{tabular}{
	 P{\figWidth}
	P{0.2cm}
	P{\figWidth}
	P{\figWidth}
    P{\figWidth}
    P{\figWidth}
    P{\figWidth}
    P{\figWidth}
	P{\figWidth}
    P{\figWidth}
    P{\figWidth}
    P{\figWidth}
    }
 
\raisebox{0em}{\multirow{2}{*}{
\addimgTtext{\includegraphics[width=\linewidth]{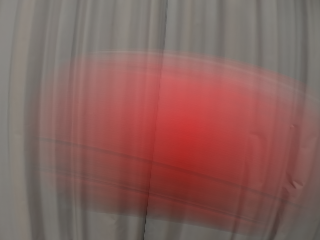}}{6}}}
&\raisebox{1.1em}{\multirow{2}{*}{\rotatebox[origin=c]{90}{DELG~\cite{delg}}}}
    &\addimgTtext{\includegraphics[width=\linewidth]{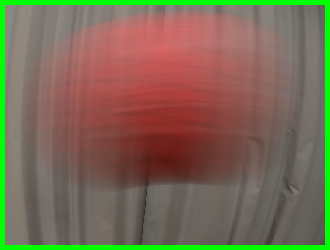}}{6}
      &\addimgTtext{\includegraphics[width=\linewidth]{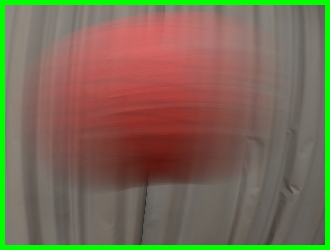}}{5}
      &\addimgTtext{\includegraphics[width=\linewidth]{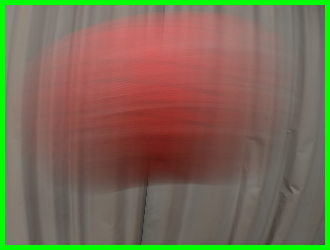}}{6}
      &\addimgTtext{\includegraphics[width=\linewidth]{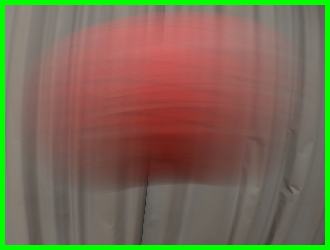}}{6}
      &\addimgTtext{\includegraphics[width=\linewidth]{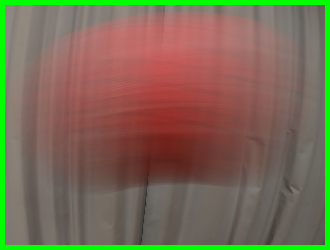}}{6}
      &\addimgTtext{\includegraphics[width=\linewidth]{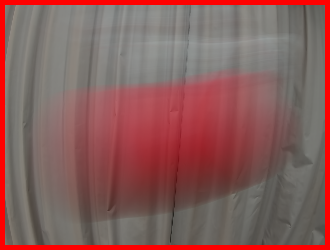}}{6}
      &\addimgTtext{\includegraphics[width=\linewidth]{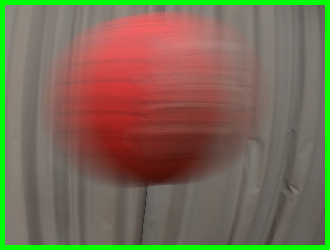}}{4}
      &\addimgTtext{\includegraphics[width=\linewidth]{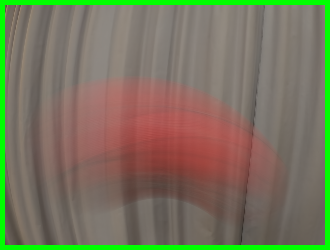}}{6}
      &\addimgTtext{\includegraphics[width=\linewidth]{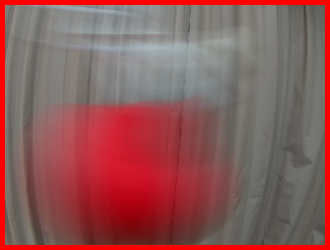}}{5}
      &\addimgTtext{\includegraphics[width=\linewidth]{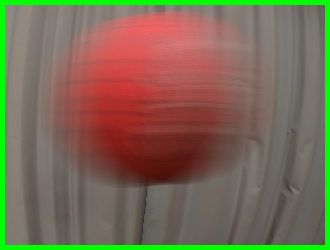}}{4}
		\\
    &&\addimgTtext{\includegraphics[width=\linewidth]{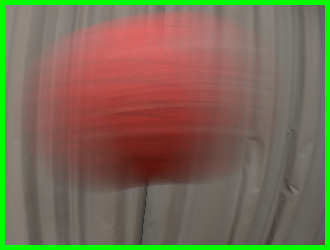}}{5}
      &\addimgTtext{\includegraphics[width=\linewidth]{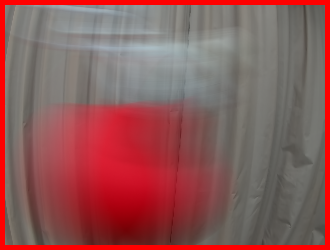}}{5}
      &\addimgTtext{\includegraphics[width=\linewidth]{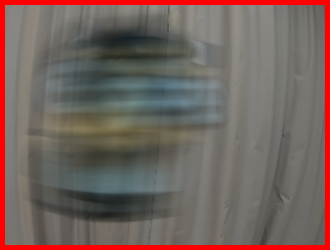}}{5}
      &\addimgTtext{\includegraphics[width=\linewidth]{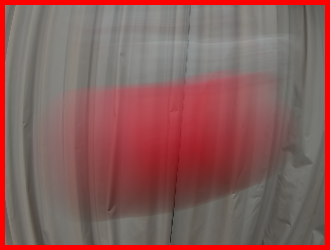}}{6}
      &\addimgTtext{\includegraphics[width=\linewidth]{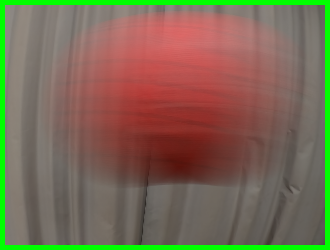}}{5}
      &\addimgTtext{\includegraphics[width=\linewidth]{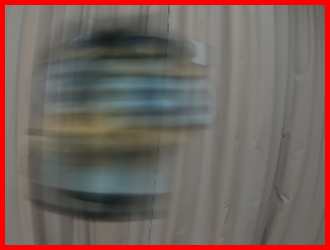}}{5}
      &\addimgTtext{\includegraphics[width=\linewidth]{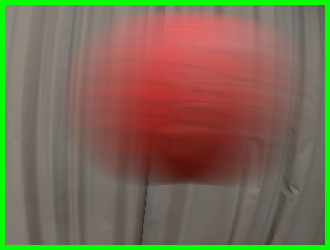}}{4}
      &\addimgTtext{\includegraphics[width=\linewidth]{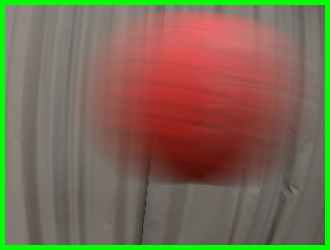}}{4}
      &\addimgTtext{\includegraphics[width=\linewidth]{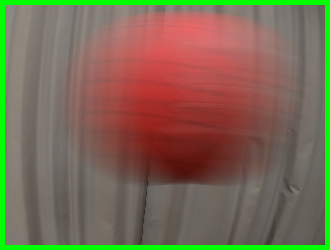}}{4}
      &\addimgTtext{\includegraphics[width=\linewidth]{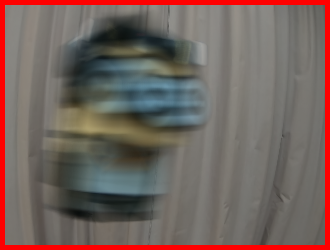}}{4} \\ 

\raisebox{0em}{\multirow{2}{*}{
\addimgTtext{\includegraphics[width=\linewidth]{fig/real/qualitativeBL6_bask/query_image_275.png}}{6}}}
&\raisebox{0.1em}{\multirow{2}{*}{\rotatebox[origin=c]{90}{Ours}}}
    &\addimgTtext{\includegraphics[width=\linewidth]{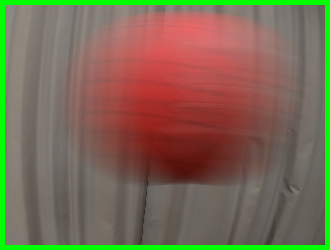}}{4}
      &\addimgTtext{\includegraphics[width=\linewidth]{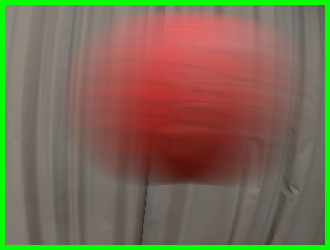}}{4}
      &\addimgTtext{\includegraphics[width=\linewidth]{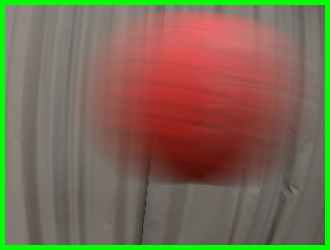}}{4}
      &\addimgTtext{\includegraphics[width=\linewidth]{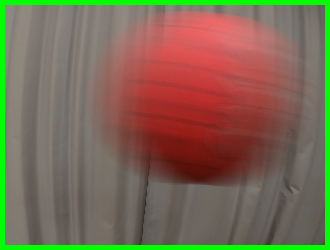}}{4}
      &\addimgTtext{\includegraphics[width=\linewidth]{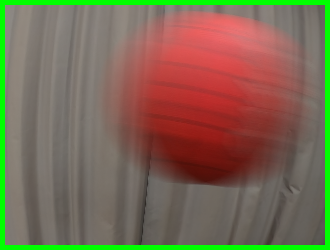}}{4}
      &\addimgTtext{\includegraphics[width=\linewidth]{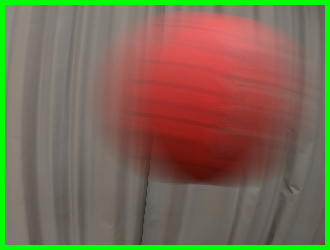}}{4}
      &\addimgTtext{\includegraphics[width=\linewidth]{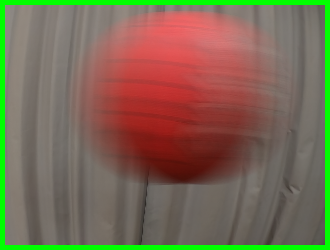}}{4}
      &\addimgTtext{\includegraphics[width=\linewidth]{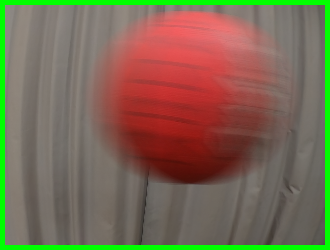}}{3}
      &\addimgTtext{\includegraphics[width=\linewidth]{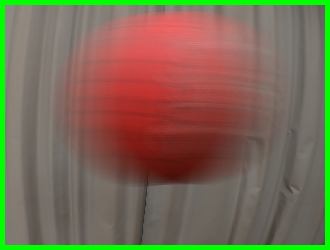}}{4}
      &\addimgTtext{\includegraphics[width=\linewidth]{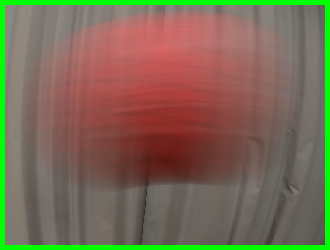}}{6}
		\\
    &&\addimgTtext{\includegraphics[width=\linewidth]{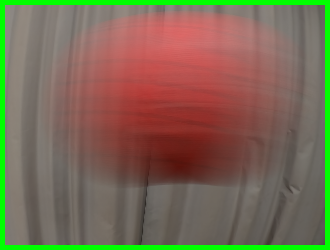}}{5}
      &\addimgTtext{\includegraphics[width=\linewidth]{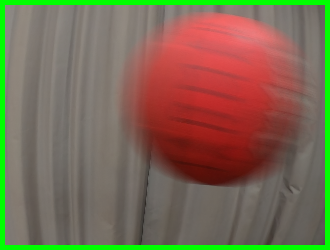}}{3}
      &\addimgTtext{\includegraphics[width=\linewidth]{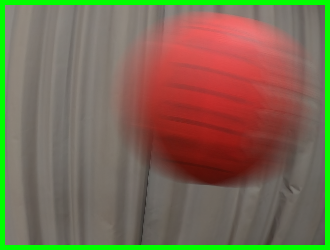}}{3}
      &\addimgTtext{\includegraphics[width=\linewidth]{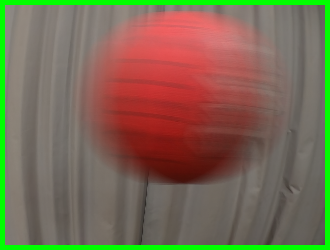}}{3}
      &\addimgTtext{\includegraphics[width=\linewidth]{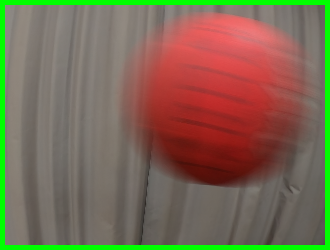}}{3}
      &\addimgTtext{\includegraphics[width=\linewidth]{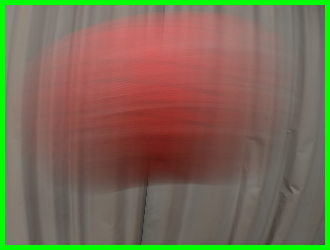}}{6}
      &\addimgTtext{\includegraphics[width=\linewidth]{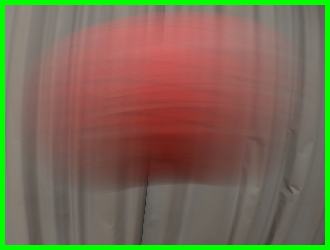}}{6}
      &\addimgTtext{\includegraphics[width=\linewidth]{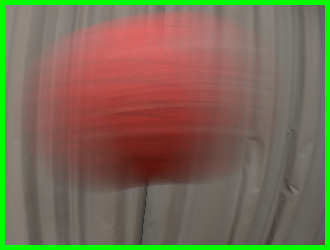}}{5}
      &\addimgTtext{\includegraphics[width=\linewidth]{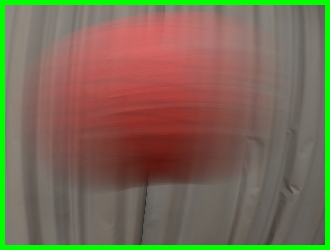}}{5}
      &\addimgTtext{\includegraphics[width=\linewidth]{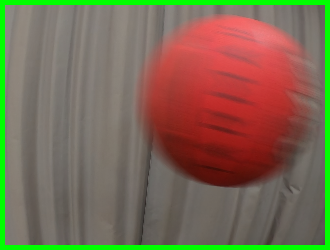}}{2} \\

	\end{tabular}
    \vspace{-1ex}
\caption{\textbf{Example of retrieval results on our real-world dataset.}}
	\label{fig:real_qualitative_1}
\end{subfigure}
\vspace{-1.8em}
\caption{\textbf{Comparison of retrieval results.} 
We compare to the best-performing state-of-the-art method for the corresponding query.
The retrieved images are sorted from left to right and top to bottom, with the ranking from 1\textsuperscript{st} to 20\textsuperscript{th}. The blur level of each image is shown in the bottom left corner. 
Correct results are in green boxes, while incorrect ones are in red boxes. 
We provide more results in the supplementary material.
}
\label{fig:qualitative}

\vspace{-10px}\end{figure*}

\boldparagraph{Implementation details.}
The proposed method is implemented in PyTorch~\cite{PyTorch}.
We use ResNet-50~\cite{resnet} up to the last downsampling block as the backbone, deactivate the normalization layers, and initialize the model with ImageNet~\cite{imagenet} pre-trained weights. 
ADAM~\cite{adam} optimizer is employed to minimize the joint loss with a fixed learning rate $10^{-4}$, batch size 32, and 30 epochs. 
We initialize the power parameter to 3 for GeM pooling~\cite{gem}. 
The contrastive margin $\tau$ is set to 0.7. 
Following \cite{gem}, we set the number of contrastive samples as $N_p=1, N_n=5$.
The contrastive blur range is experimentally set to 5.
For classification, we empirically set the margin $m$ to 0.15 and the scale $\gamma$ to 30. 
Loss weights are chosen so that the values of different losses are roughly of the same order of magnitude, \ie $\alpha_{cls}=0.1, \alpha_{be}=1.0, \alpha_{loc}=10$. 
The descriptor dimension $d$ is set to 128.

\boldparagraph{Evaluation protocols.}
We report mean average precision (mAP@all) as commonly used in retrieval. 
In case 1M distractors are used, we report mAP@100.

\boldparagraph{Compared methods.} 
Since retrieval with object motion blur has not been addressed before, there are no methods specifically designed for this task. 
Therefore, we compare to state-of-the-art standard retrieval methods: DELG-global~\cite{delg} (referred to as DELG), DOLG~\cite{dolg}, and Token \cite{token}. 
All these methods are re-implemented as in the original papers, except for changing the size of the output descriptors to 128 for fair comparison.
We re-train the compared methods using the same ArcFace parameters and on the same synthetic data as our approach.
\begin{table}[t] 
    \centering
    \begin{adjustbox}{max width=\linewidth}
    \setlength{\tabcolsep}{4pt}
    \begin{tabular}{c c:ccc c c cccccc}
        \toprule
         \multirow{2}{*}{$\mathcal{L}_{con}$} 
         & \multirow{2}{*}{$\mathcal{L}_{cls}$} 
         & \multirow{2}{*}{$\mathcal{L}_{be}$} 
         & \multirow{2}{*}{$\mathcal{L}_{loc}$} 
         
         & {mAP} 
         &
         &  \multicolumn{6}{c}{mAP by Query BL}\\
         \cline{7-12}
          
          &&&&All   &&  1 &  2&  3&  4 &  5  & 6 \\
          \midrule
          \checkmark & \xmark & \xmark & \xmark & 78.13 && 80.51 & 81.70 & 81.16 & 79.20 & 73.93 & 61.24 \\
          \checkmark & \xmark & \checkmark &\xmark & 81.66 && 83.49 & 85.01 & 84.43 & 82.69 & 77.64 & 67.08 \\
          \checkmark & \xmark & \xmark &\checkmark  & 85.94 && 87.54 & 88.25 & 87.83 & 86.52 & 83.08 & 75.58 \\
          \checkmark & \xmark &\checkmark  &\checkmark  & \textbf{87.48} && \textbf{88.69} & \textbf{89.89} & \textbf{89.40} & \textbf{88.24} & \textbf{84.97} & \textbf{76.74} \\

          \hdashline

          \xmark & \checkmark & \xmark & \xmark & 78.73 && 81.53 & 82.97 & 82.83 & 79.86 & 72.93 & 59.00 \\
          \xmark & \checkmark & \checkmark & \xmark & 83.67 && 85.19 & 87.56 & 87.40 & 85.10 & 79.22 & 65.93 \\
          \xmark & \checkmark & \xmark & \checkmark & 88.74 && 89.89 & 90.91 & 90.88 & 89.75 & 86.07 & 77.67 \\
          \xmark & \checkmark & \checkmark & \checkmark & \textbf{91.23} && \textbf{92.02} & \textbf{93.16} & \textbf{93.09} & \textbf{91.97} & \textbf{89.00} & \textbf{82.27} \\

            \hdashline
            
          \checkmark & \checkmark & \xmark & \xmark & 85.06 && 87.42 & 88.29 & 87.66 & 85.85 & 81.20 & 69.96 \\
          \checkmark & \checkmark & \checkmark & \xmark & 87.17 && 89.03 & 90.03 & 89.55 & 88.07 & 83.91 & 73.48\\
          \checkmark & \checkmark & \xmark & \checkmark & 90.39  && 91.85 & 92.45 & 92.14 & 91.20 & 88.20 & 79.36\\
          \checkmark & \checkmark & \checkmark & \checkmark & \textbf{91.78} && \textbf{93.05} & \textbf{93.48} & \textbf{93.14} & \textbf{92.25} & \textbf{90.20} & \textbf{82.86} \\

        \bottomrule
    \end{tabular}
    \end{adjustbox}
\caption{\textbf{Ablation study} of different loss components on the synthetic dataset. Results are grouped according to the loss function(s) directly applied to the descriptor. The best scores within each split are bolded.
The database contains all blur levels. 
}
\vspace{-1.2em} 
\label{tab:ablate_losses}

\vspace{-10px}\end{table} 

\subsection{Results}

\boldparagraph{All blur levels in the database.} 
Our method outperforms all state-of-the-art retrieval methods for all settings, as shown in Table~\ref{tab:mAP_db_mixed} for synthetic data and Table~\ref{tab:mAP_real} for real data.
Our model demonstrates significant superiority over the state-of-the-art retrieval approaches without introducing complex computational modules, highlighting the efficacy of our proposed loss functions.
Results on the real-world dataset show that our method generalizes well to real data even when trained solely on synthetic data.
Additionally, we train a version of our method where only sharp images are used (Ours-sharp).
When evaluating with mAP@all (no distractors), it fails completely, especially for higher blur levels.
When evaluating with mAP@100 (with distractors), it achieves reasonable results on query BL 1, but still worse than all other methods.
This underscores the necessity for tailored methods and training data with highly motion-blurred objects.

\boldparagraph{Per blur level database.} 
In this setup, we utilize datasets specific to each blur level, considering them as individual subsets of images with the designated blur level, to form the database. We then assess the retrieval performance by using query images with distinct blur levels, resulting in a comprehensive evaluation comprising 36 results (6 database BLs by 6 query BLs) for each method.
We also compute the standard deviation (std) and range of mAPs for different settings per method, reflecting the robustness of each method to different levels of motion blur. 
Our method outperforms all other state-of-the-art methods for all settings and also achieves the smallest standard deviation and range, showing the highest robustness to motion blur (Table \ref{tab:mAP_db_pure_with_distractors}). See supplementary material for more results.

\boldparagraph{Qualitative results} on our proposed synthetic (\Fig \ref{fig:syn_qualitative_1}) and real-world (\Fig \ref{fig:real_qualitative_1}) datasets. 
Our method is compared to the top-performing state-of-the-art retrieval methods on a per-query basis, which yield numerous false positives resembling the texture of the motion-blurred object in the query image. In contrast, our approach accurately retrieves positive images even under severe motion blur in the query and amidst challenging distractors. Note that standard retrieval methods retrieve positive images primarily with blur levels similar to the query, whereas our method retrieves correct images exhibiting a wide range of blur levels (from 1-6 for synthetic and 2-6 for real). This illustrates that our method effectively learns invariant representations of objects across various blur levels.

\begin{table}[t] 
    \centering
    \begin{adjustbox}{max width=\linewidth}
    \setlength{\tabcolsep}{4pt}
    \begin{tabular}{P{1.5cm} c c cccccc}
        \toprule
        
             Cont. BL & {mAP} &&  \multicolumn{6}{c}{mAP by query BL}\\ 
            \cline{4-9}
             Range &All   && 1 &  2 &  3&  4&  5&  6 \\
        \midrule
          
            S &82.26 && 87.37 & 88.46 & 86.29 & 82.61 & 74.13 & 58.26\\
            M &83.33 && 87.97 & 89.12 & 86.93 & 83.87 & 76.04 & 60.61\\
            L &\textbf{84.09} && \textbf{88.74} & \textbf{89.56} & \textbf{87.68} & \textbf{84.41} & \textbf{76.89} & \textbf{62.42}\\
            \midrule
            L (Alpha) & 82.83 && 87.62 & 88.68 & 86.43 & 82.96 & 75.68 & 60.27\\
        \bottomrule
    \end{tabular}
    \end{adjustbox}

    \caption{\textbf{Ablation study} of different blur level ranges in contrastive learning and design of blur estimation loss on the synthetic dataset (with 1M distractors).
    \textbf{S}mall: range=1. \textbf{M}edium: range=3. \textbf{L}arge: range=5. 'Alpha' corresponds to the alternative design choice for blur estimation loss. The database contains images of all blur levels. 
    }
    
    \label{tab:ablate_ranges}
        \vspace{-0.8em} 

\vspace{-10px}\end{table}

\subsection{Ablation study}
\boldparagraph{Loss components.} 
Table \ref{tab:ablate_losses} shows an ablation study of different loss functions (for more details, please see the supplementary material). 
The results show that all loss functions improve model performance.
We highlight that our model outperforms the compared methods across all query blur levels (Table~\ref{tab:mAP_db_mixed}), even when utilizing only three out of the four loss terms \textit{at any combination}. 
For instance, when using  $\mathcal{L}_{con}$, $\mathcal{L}_{cls}$ and $\mathcal{L}_{be}$, 
our method surpasses Token~\cite{token} by 2.50 in mAP at blur level 1 and 1.91 at blur level 6, with an overall improvement of 2.33.

\boldparagraph{Contrastive range.} 
We investigate the research questions posed in Sec.~\ref{sec:bliss} using Table \ref{tab:ablate_ranges}, which displays results of using different blur level ranges for contrastive learning. A wider range of contrastive blur levels consistently enhances model performance, with the best results achieved at the maximum range of 5.

\boldparagraph{Design of blur estimation loss.} 
We validate the effectiveness of our design of blur estimation loss $\mathcal{L}_{be}$ (Eq.~\eqref{eq:be_loss}) in Table~\ref{tab:ablate_ranges}. We compare with the case where edge effects are ignored in the loss function. This corresponds to directly predicting the average of the entire alpha mask (Eq.~\eqref{eq:alpha}) inside the object boundary. 
The results show that our design of $\mathcal{L}_{be}$, which effectively mitigates the detrimental impact of edge blur, improves the model's retrieval performance.

\subsection{
Application to real-world video data
}
We conduct an experiment to test our method on real-world video images. Due to the challenge of finding a sufficient number of real images of the same object with a well-distributed range of blur levels for quantitative analysis, we construct a small dataset for qualitative assessment. Specifically, we extract 190 images of the same ball from a \href{https://www.youtube.com/watch?v=U8WCRz0Yh4Q}{YouTube soccer video} to serve as both the query and database. We incorporate 4,431 images of sports balls from the MSCOCO dataset~\cite{coco} as distractors, and additionally, we extract 169 images of a different ball from the same video to act as hard distractors. 
Both the objects and the backgrounds in these images come from domains entirely different from the synthetic training data.
Fig. \ref{fig:test_youtube_video} illustrates examples of retrieval results, demonstrating our method's effectiveness in handling various blur conditions and complex and diverse backgrounds in the real world, despite being trained exclusively on synthetic data.
\global\long\def\figWidth{0.077\linewidth}
\begin{figure*}[t]
	\centering
 \begin{subfigure}{\linewidth}
    \setlength{\tabcolsep}{2pt}
    \setlength{\fboxrule}{1pt} 
    \setlength{\fboxsep}{0pt} 
	\begin{tabular}{
	 P{\figWidth}
	P{0.05cm}
	P{\figWidth}
	P{\figWidth}
    P{\figWidth}
    P{\figWidth}
    P{\figWidth}
    P{\figWidth}
	P{\figWidth}
    P{\figWidth}
    P{\figWidth}
    P{\figWidth}
    }
    \multicolumn{2}{c}{Query}	&   \multicolumn{10}{c}{Top 20 retrieval results}
		\\
    \raisebox{0em}{\multirow{2}{*}{
    {\includegraphics[width=\linewidth]{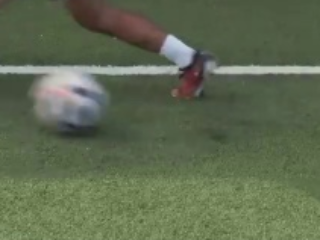}}}}
    &&{\includegraphics[width=\linewidth]{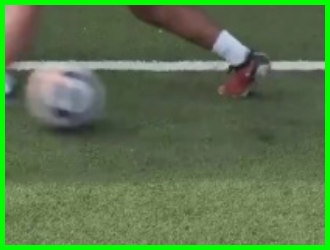}}
      &{\includegraphics[width=\linewidth]{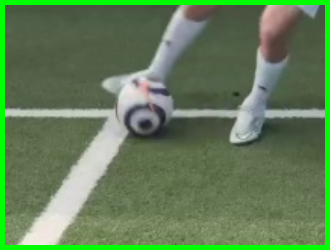}}
      &{\includegraphics[width=\linewidth]{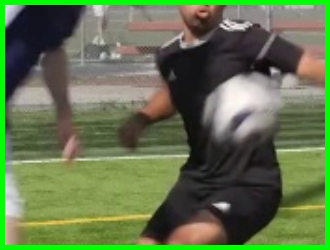}}
      &{\includegraphics[width=\linewidth]{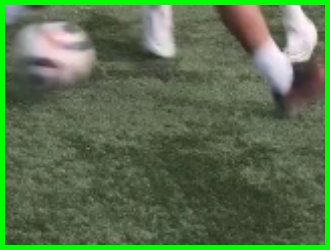}}
      &{\includegraphics[width=\linewidth]{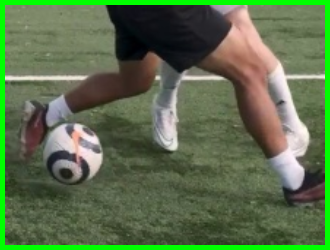}}
      &{\includegraphics[width=\linewidth]{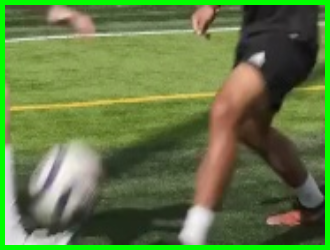}}
      &{\includegraphics[width=\linewidth]{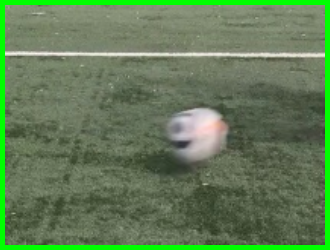}}
      &{\includegraphics[width=\linewidth]{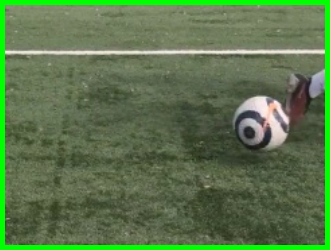}}
      &{\includegraphics[width=\linewidth]{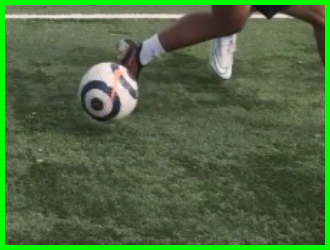}}
      &{\includegraphics[width=\linewidth]{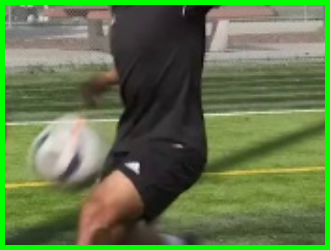}}
		\\
    &&{\includegraphics[width=\linewidth]{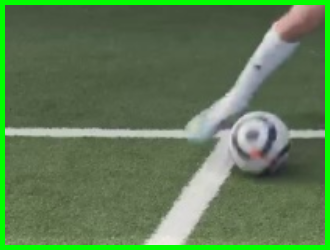}}
      &{\includegraphics[width=\linewidth]{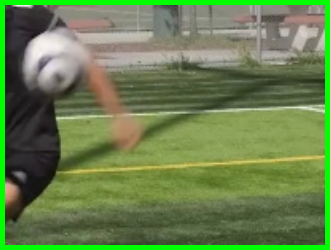}}
      &{\includegraphics[width=\linewidth]{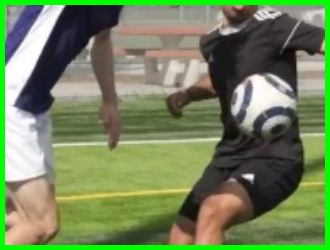}}
      &{\includegraphics[width=\linewidth]{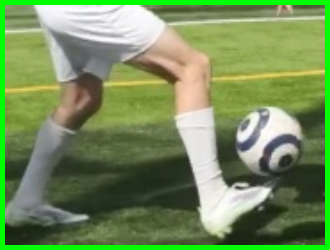}}
      &{\includegraphics[width=\linewidth]{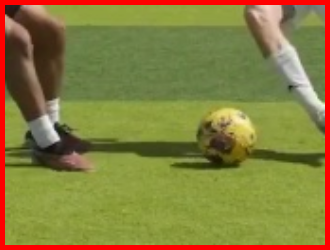}}
      &{\includegraphics[width=\linewidth]{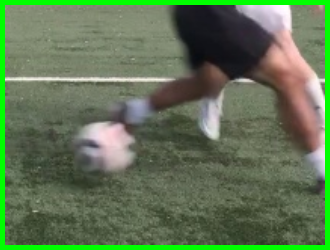}}
      &{\includegraphics[width=\linewidth]{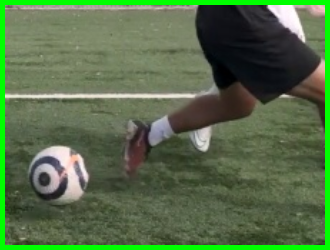}}
      &{\includegraphics[width=\linewidth]{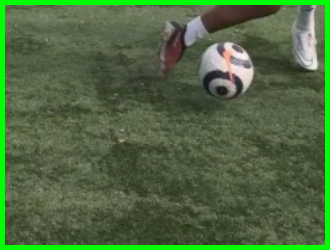}}
    &{\includegraphics[width=\linewidth]{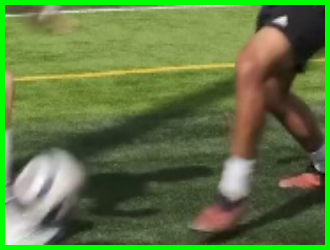}}
     &{\includegraphics[width=\linewidth]{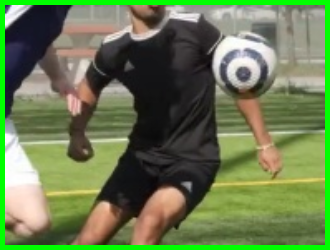}} \\
     \addlinespace[5pt]
    \raisebox{0em}{\multirow{2}{*}{
    {\includegraphics[width=\linewidth]{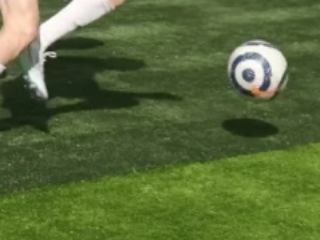}}}}
    &&{\includegraphics[width=\linewidth]{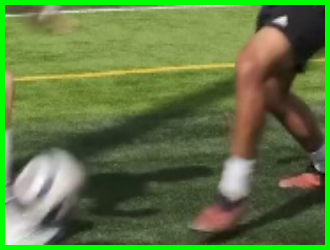}}
      &{\includegraphics[width=\linewidth]{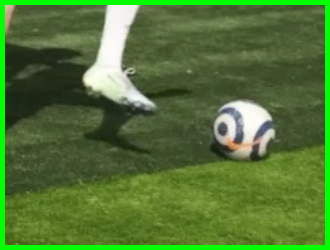}}
      &{\includegraphics[width=\linewidth]{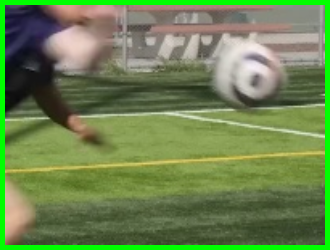}}
      &{\includegraphics[width=\linewidth]{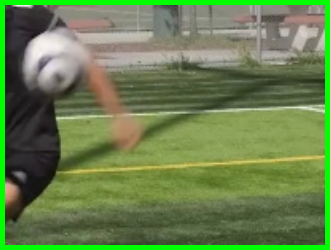}}
      &{\includegraphics[width=\linewidth]{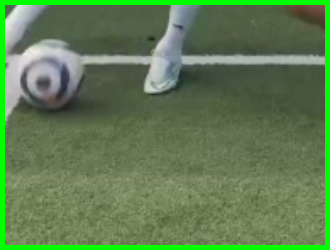}}
      &{\includegraphics[width=\linewidth]{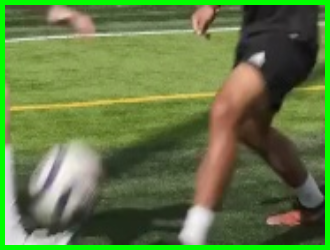}}
      &{\includegraphics[width=\linewidth]{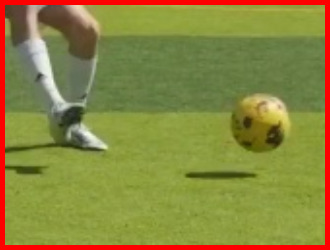}}
      &{\includegraphics[width=\linewidth]{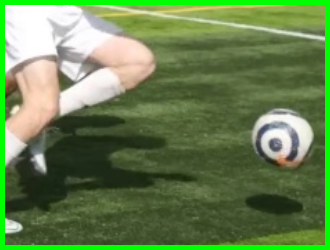}}
      &{\includegraphics[width=\linewidth]{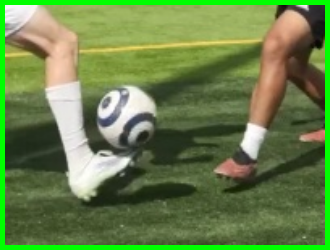}}
      &{\includegraphics[width=\linewidth]{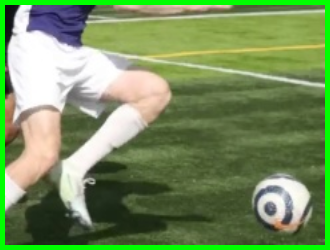}}
		\\
    &&{\includegraphics[width=\linewidth]{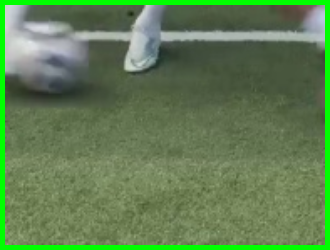}}
      &{\includegraphics[width=\linewidth]{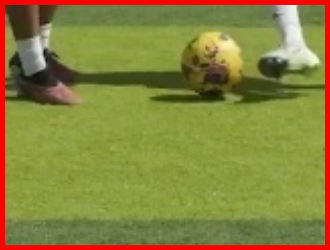}}
      &{\includegraphics[width=\linewidth]{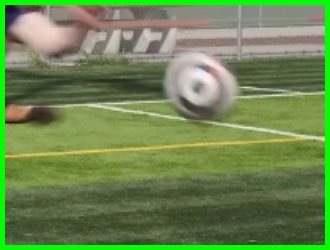}}
      &{\includegraphics[width=\linewidth]{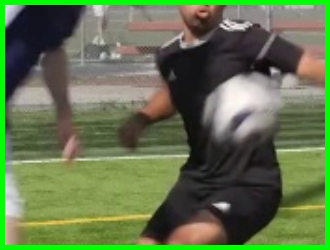}}
      &{\includegraphics[width=\linewidth]{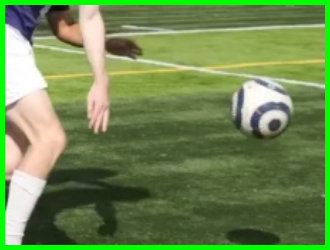}}
      &{\includegraphics[width=\linewidth]{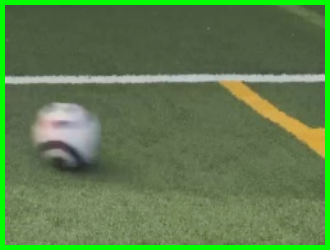}}
      &{\includegraphics[width=\linewidth]{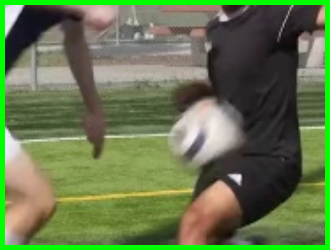}}
      &{\includegraphics[width=\linewidth]{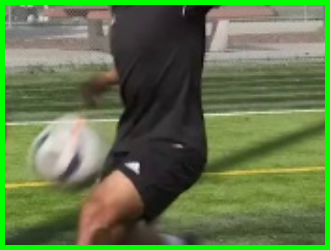}}
    &{\includegraphics[width=\linewidth]{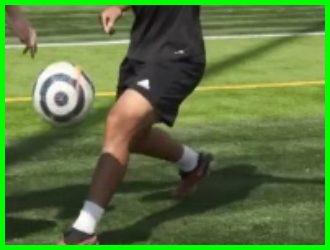}}
     &{\includegraphics[width=\linewidth]{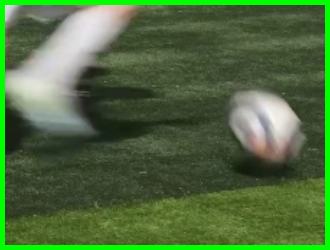}} \\
	\end{tabular}
    \vspace{-1ex}
\end{subfigure}
\vspace{-1.8em}
\caption{\textbf{Retrieval results on a real-world video} (available on \href{https://www.youtube.com/watch?v=U8WCRz0Yh4Q}{YouTube}). 
The retrieved images are sorted from left to right and top to bottom, with the ranking from 1\textsuperscript{st} to 20\textsuperscript{th}. 
Correct results are in green boxes, while incorrect ones are in red boxes. }
\label{fig:test_youtube_video}

\end{figure*}

\section{Conclusion}

In this paper, we introduce an innovative task: image retrieval in the presence of object motion blur. To tackle this challenge, we propose a global retrieval approach that integrates specialized loss functions tailored for the network to better comprehend motion blur. Our method stands as the first specifically designed solution for this new task.
Moreover, we introduce the first benchmark datasets for blur retrieval, comprising both synthetic and real-world data. These datasets are meticulously processed and annotated to ensure their immediate applicability for future research in this emerging area.
Through extensive experiments conducted on these benchmark datasets, we demonstrate the limitations of state-of-the-art standard retrieval methods in handling motion-blurred images and showcase the superior performance and robustness of our proposed approach in accurately retrieving images even in challenging motion-blurred scenarios. 

\boldparagraph{Acknowledgement.} We would like to thank Giorgos Tolias for the highly valuable discussions with him and the support he provided for this work. 
\title{\MYTITLE\\(Supplementary Material)}
\author{Rong Zou\inst{1}\orcidlink{0009-0002-0434-5746} \and
Marc Pollefeys\inst{1,2}\orcidlink{0000-0003-2448-2318} \and
Denys Rozumnyi\inst{1,3}\orcidlink{0000-0001-9874-1349}}

\authorrunning{R. Zou et al.}

\institute{$^1$ETH Zürich, $^2$Microsoft, $^3$Czech Technical University in Prague}

\maketitle
\thispagestyle{empty}

\setcounter{page}{1}  

\section{Privacy and ethical considerations of data}
Both our synthetic and real-world datasets comprise only objects, guaranteeing the absence of any personal or identifiable information. All images, whether real or synthetic, are devoid of human presence, thus eliminating privacy concerns. We have adhered to ethical guidelines in data collection and utilization. The real-world object data has been gathered with permission, ensuring compliance with ethical standards.

\section{Distractor set examples} 
Examples from our distractor set are presented in \Fig \ref{fig:distractor_examples}. For every query example, we display 8 negative images from the distractor set. In addition, we include 2 positive images from the database: one blurred and another in a sharp state. These examples vividly illustrate the obfuscation caused by object motion blur, highlight the challenging nature of the distractors, and underscore the inherent difficulty of the task of retrieval in the presence of object motion blur introduced in this study. 

\section{Qualitative criteria for real data annotation}
Establishing a uniform quantitative criterion to measure object motion blur in the real world is challenging, particularly considering the diversity of objects and motion types. Therefore, we manually annotate real data to better analyze and evaluate the model performance at different blur levels. To facilitate this process, we define a set of qualitative criteria for assigning each image to a specific blur level, as follows:
\begin{itemize}[label=\textbullet]
    \item \textbf{Blur Level 1}: The object is highly recognizable, with no or very subtle loss of details.
    \item \textbf{Blur Level 2}: The object is still easily recognizable, with a slight reduction in clarity and mild loss of defining features.
    \item \textbf{Blur Level 3}: The object is recognizable but with a moderate reduction in clarity; some features may be challenging to discern.
    \item \textbf{Blur Level 4}: The object is recognizable but with considerable difficulty due to a substantial loss of details.
    \item \textbf{Blur Level 5}: The object may be barely recognizable, and object details are almost completely obscured; the overall form may be distorted.
    \item \textbf{Blur Level 6}: The object becomes abstract, making recognition extremely challenging.
\end{itemize}
In addition, we use the ratio of the trajectory length to the average object size as a quantitative reference during the annotation process.

 \global\long\def\figWidth{0.0775\linewidth} 
\begin{figure*}[t]
	\centering

    \setlength{\tabcolsep}{2pt}
    \setlength{\fboxrule}{1pt}
    \setlength{\fboxsep}{0pt}
	\begin{tabular}{
	 P{\figWidth}
  c
	P{\figWidth}
	P{\figWidth}
    P{\figWidth}
    P{\figWidth}
    P{\figWidth}
    P{\figWidth}
	P{\figWidth}
    P{\figWidth}
    c
    P{\figWidth}
    P{\figWidth}
    }
	{Query}   &&  \multicolumn{8}{c}{Hard Distractors} && \multicolumn{2}{c}{Positives}
		\\

\includegraphics[width=\linewidth]{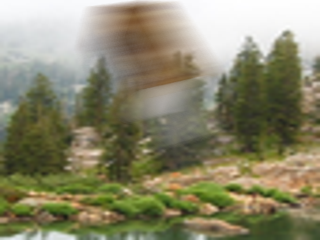}
    &&  \includegraphics[width=\linewidth]{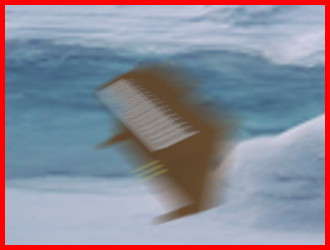}
    &  \includegraphics[width=\linewidth]{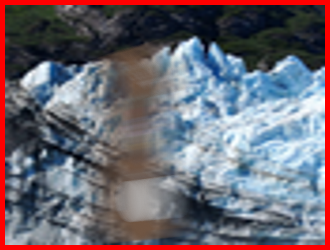}
   &   \includegraphics[width=\linewidth]{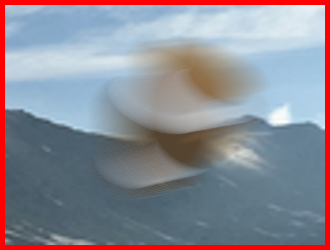}
   &   \includegraphics[width=\linewidth]{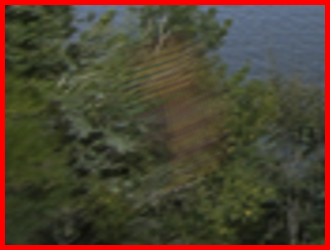}
   &   \includegraphics[width=\linewidth]{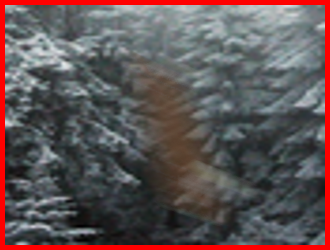}
   &   \includegraphics[width=\linewidth]{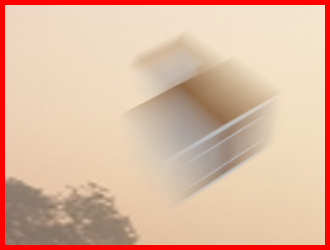}
   &   \includegraphics[width=\linewidth]{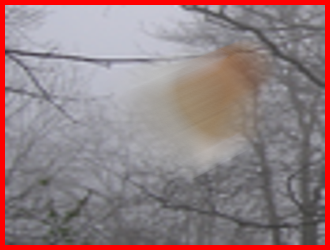}
   &   \includegraphics[width=\linewidth]{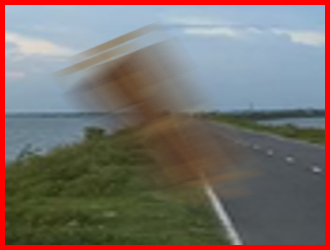}
   &&\includegraphics[width=\linewidth]{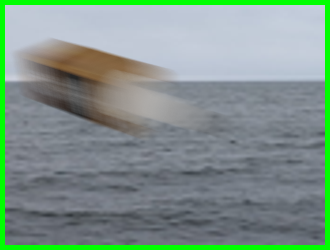}
   &\includegraphics[width=\linewidth]{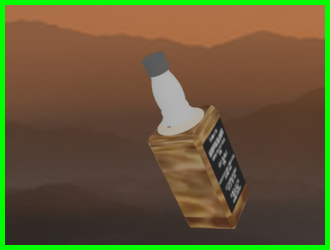}\\
    \includegraphics[width=\linewidth]{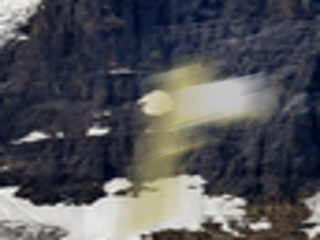}
  &&  \includegraphics[width=\linewidth]{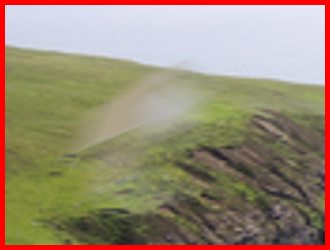}
    &  \includegraphics[width=\linewidth]{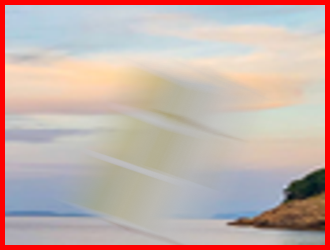}
   &   \includegraphics[width=\linewidth]{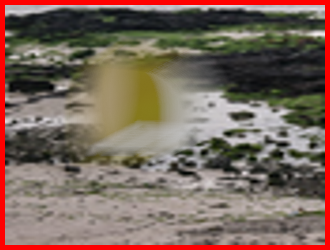}
   &   \includegraphics[width=\linewidth]{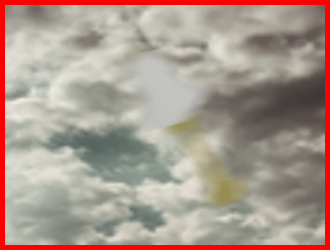}
   &   \includegraphics[width=\linewidth]{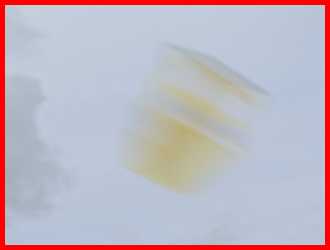}
   &   \includegraphics[width=\linewidth]{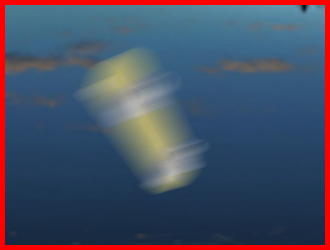}
   &   \includegraphics[width=\linewidth]{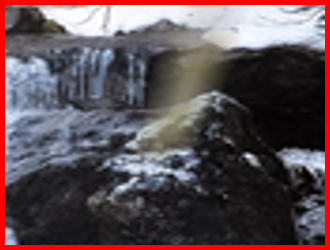}
   &   \includegraphics[width=\linewidth]{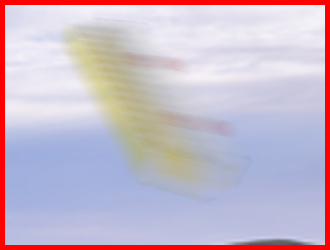}
   &&\includegraphics[width=\linewidth]{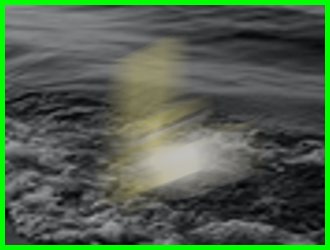}
   &\includegraphics[width=\linewidth]{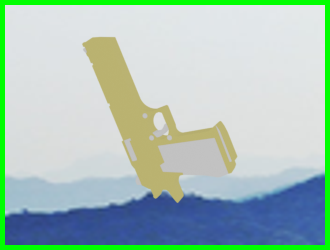}
		\\
  \includegraphics[width=\linewidth]{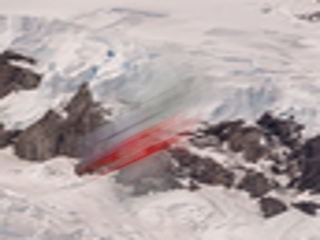}
    &&  \includegraphics[width=\linewidth]{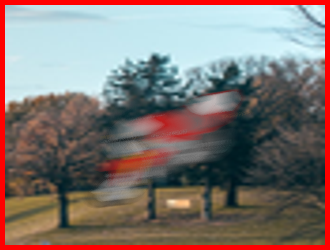}
    &  \includegraphics[width=\linewidth]{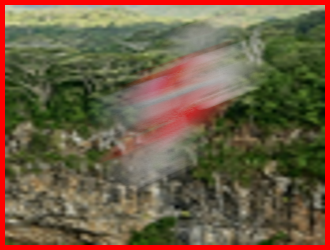}
   &   \includegraphics[width=\linewidth]{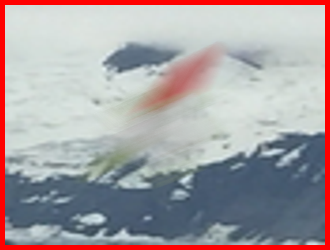}
   &   \includegraphics[width=\linewidth]{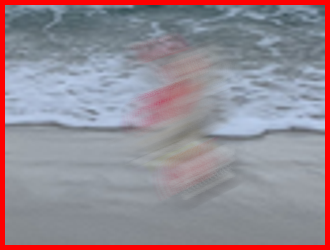}
   &   \includegraphics[width=\linewidth]{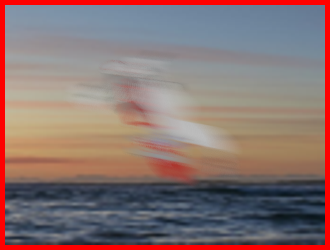}
   &   \includegraphics[width=\linewidth]{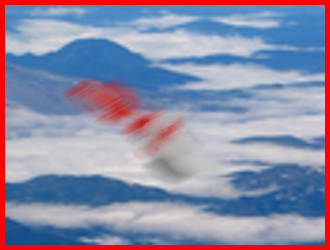}
   &   \includegraphics[width=\linewidth]{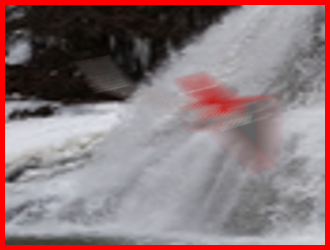}
   &   \includegraphics[width=\linewidth]{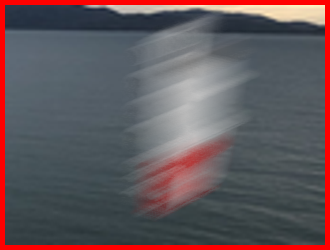}
   &&\includegraphics[width=\linewidth]{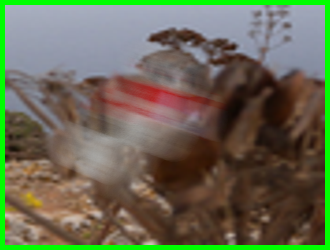}
   &\includegraphics[width=\linewidth]{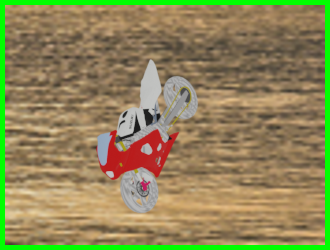}
		\\
\includegraphics[width=\linewidth]{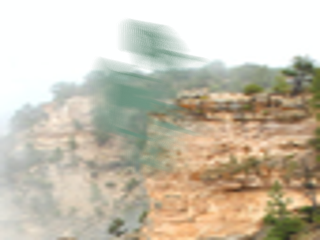}
    &&  \includegraphics[width=\linewidth]{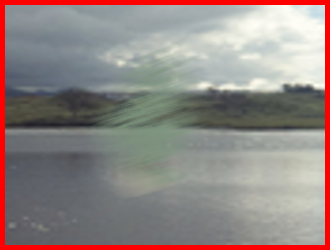}
    &  \includegraphics[width=\linewidth]{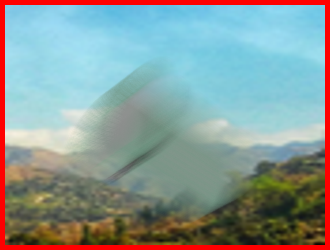}
   &   \includegraphics[width=\linewidth]{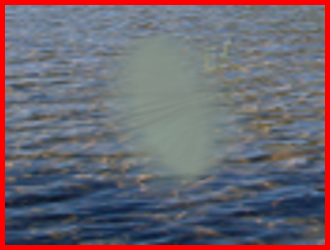}
   &   \includegraphics[width=\linewidth]{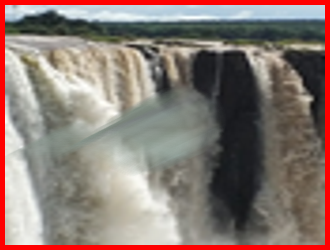}
   &   \includegraphics[width=\linewidth]{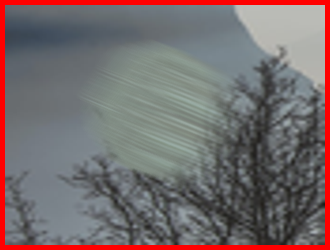}
   &   \includegraphics[width=\linewidth]{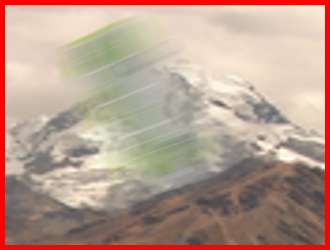}
   &   \includegraphics[width=\linewidth]{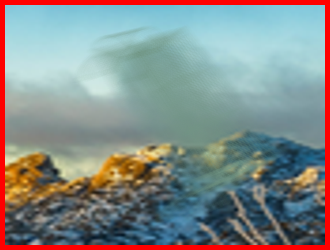}
   &   \includegraphics[width=\linewidth]{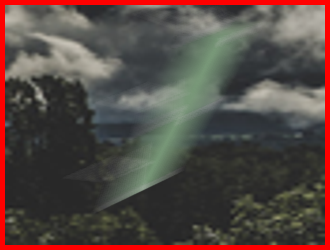}
   &&\includegraphics[width=\linewidth]{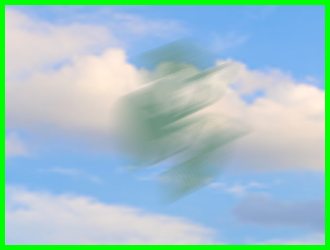}
   &\includegraphics[width=\linewidth]{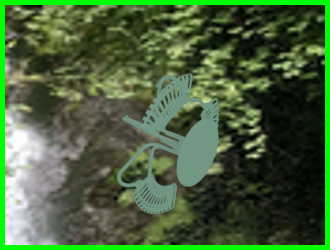}
		\\

  \includegraphics[width=\linewidth]{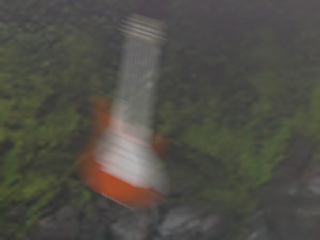}
  &&  \includegraphics[width=\linewidth]{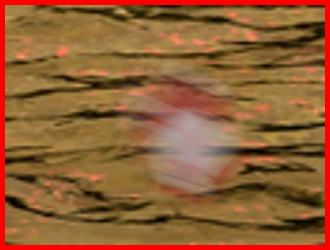}
    &  \includegraphics[width=\linewidth]{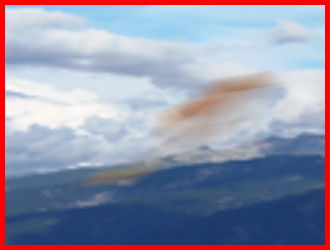}
   &   \includegraphics[width=\linewidth]{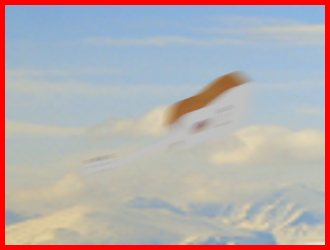}
   &   \includegraphics[width=\linewidth]{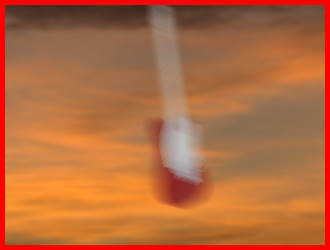}
   &   \includegraphics[width=\linewidth]{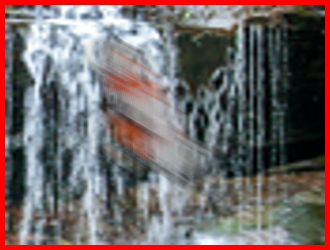}
   &   \includegraphics[width=\linewidth]{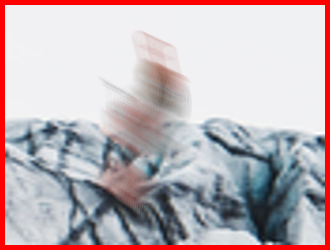}
   &   \includegraphics[width=\linewidth]{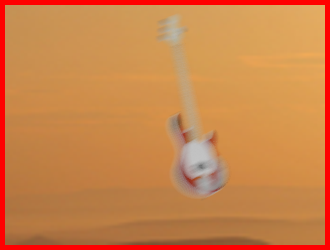}
   &   \includegraphics[width=\linewidth]{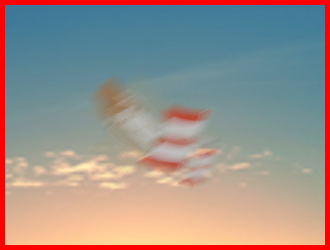}
   &&\includegraphics[width=\linewidth]{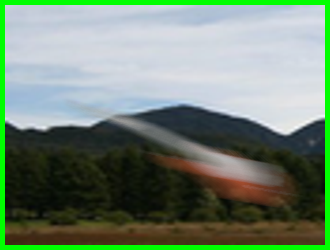}
   &\includegraphics[width=\linewidth]{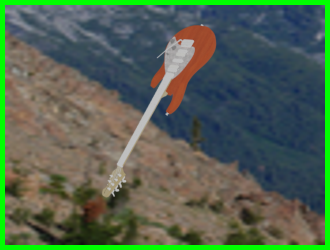}
		\\

  \includegraphics[width=\linewidth]{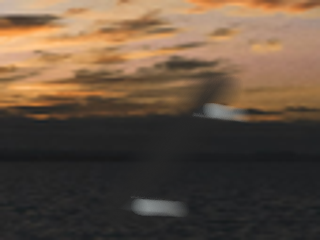}
  &&  \includegraphics[width=\linewidth]{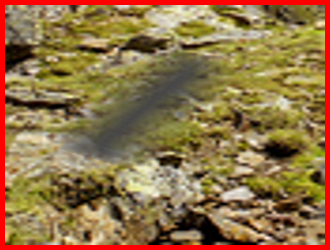}
    &  \includegraphics[width=\linewidth]{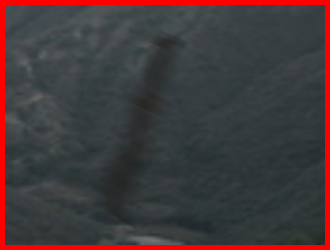}
   &   \includegraphics[width=\linewidth]{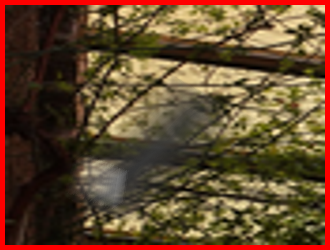}
   &   \includegraphics[width=\linewidth]{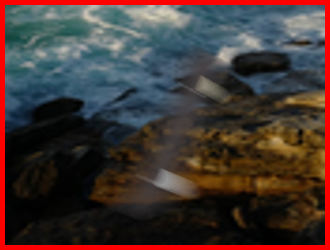}
   &   \includegraphics[width=\linewidth]{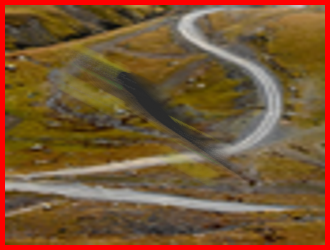}
   &   \includegraphics[width=\linewidth]{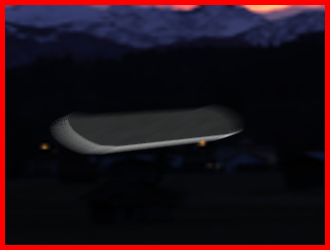}
   &   \includegraphics[width=\linewidth]{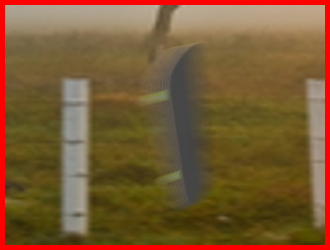}
   &   \includegraphics[width=\linewidth]{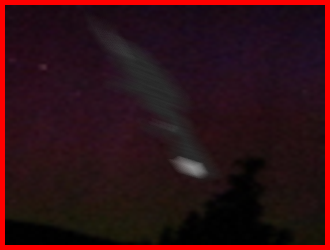}
   &&\includegraphics[width=\linewidth]{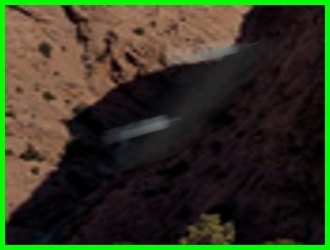}
   &\includegraphics[width=\linewidth]{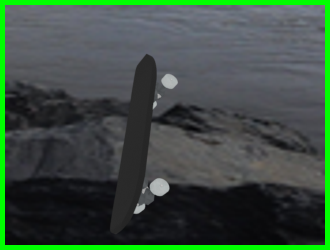}
		\\
\includegraphics[width=\linewidth]{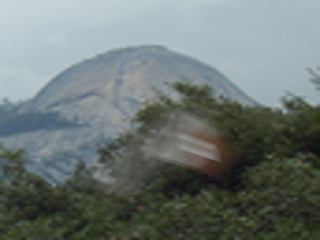}
    &&  \includegraphics[width=\linewidth]{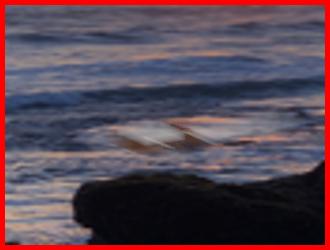}
   &   \includegraphics[width=\linewidth]{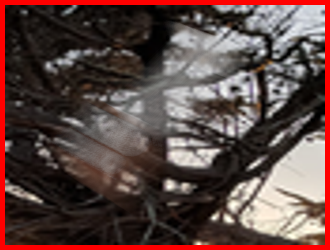}
   &   \includegraphics[width=\linewidth]{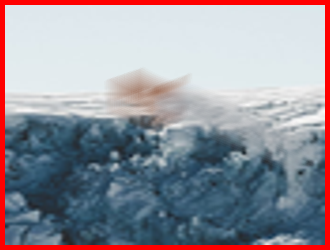}
   &   \includegraphics[width=\linewidth]{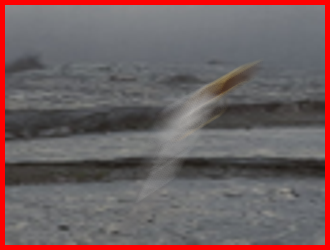}
   &   \includegraphics[width=\linewidth]{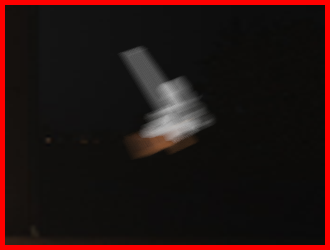}
   &   \includegraphics[width=\linewidth]{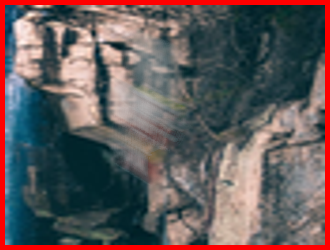}
   &   \includegraphics[width=\linewidth]{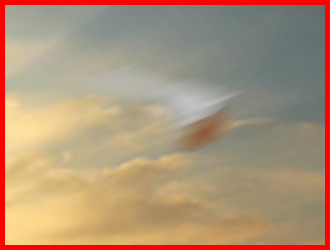}
   &   \includegraphics[width=\linewidth]{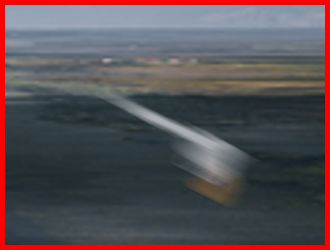}
   &&\includegraphics[width=\linewidth]{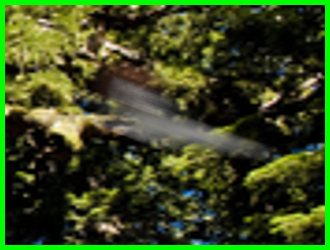}
   &\includegraphics[width=\linewidth]{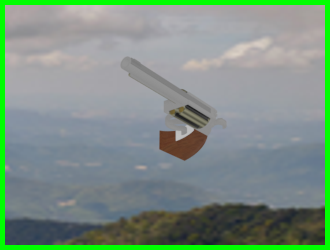}
		\\

\includegraphics[width=\linewidth]{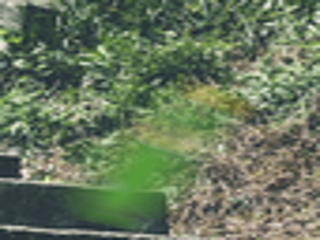}
    &&  \includegraphics[width=\linewidth]{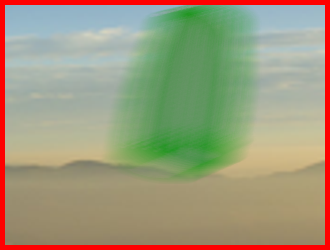}
   &   \includegraphics[width=\linewidth]{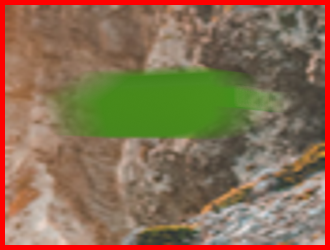}
   &   \includegraphics[width=\linewidth]{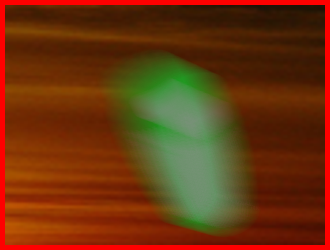}
   &   \includegraphics[width=\linewidth]{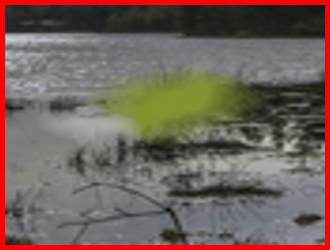}
   &   \includegraphics[width=\linewidth]{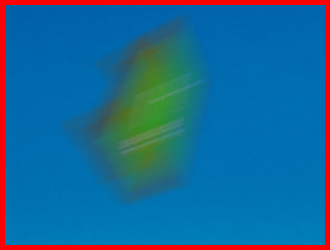}
   &   \includegraphics[width=\linewidth]{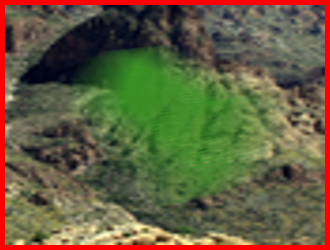}
   &   \includegraphics[width=\linewidth]{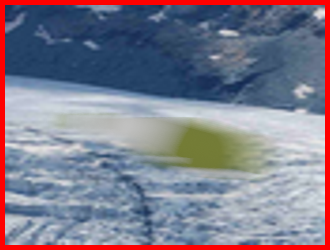}
   &   \includegraphics[width=\linewidth]{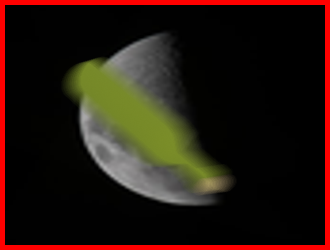}
   &&\includegraphics[width=\linewidth]{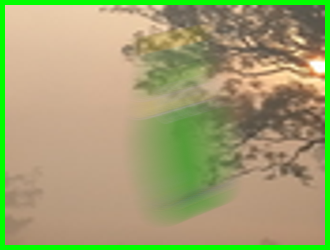}
   &\includegraphics[width=\linewidth]{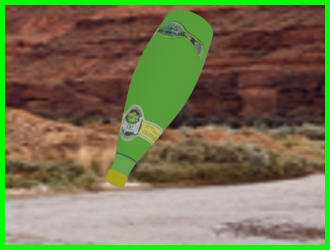}
		\\
  \includegraphics[width=\linewidth]{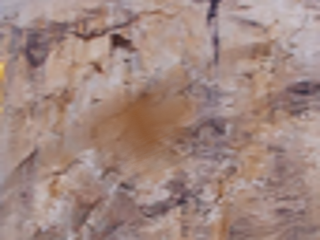}
  &&  \includegraphics[width=\linewidth]{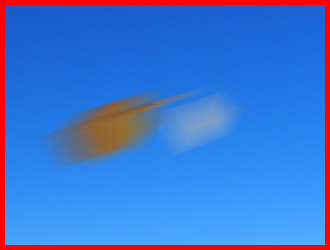}
    &  \includegraphics[width=\linewidth]{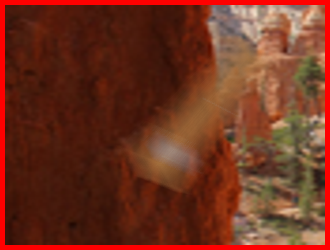}
   &   \includegraphics[width=\linewidth]{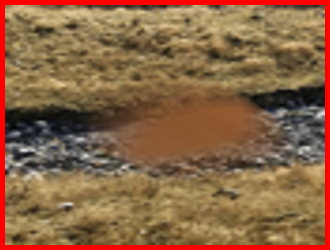}
   &   \includegraphics[width=\linewidth]{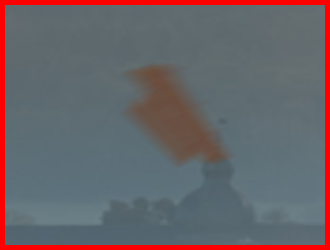}
   &   \includegraphics[width=\linewidth]{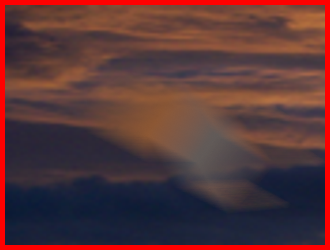}
   &   \includegraphics[width=\linewidth]{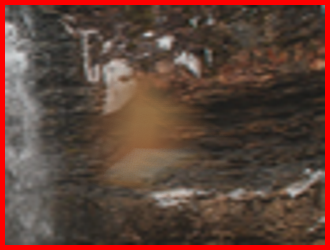}
   &   \includegraphics[width=\linewidth]{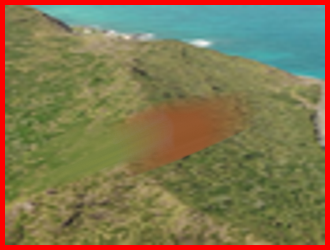}
   &   \includegraphics[width=\linewidth]{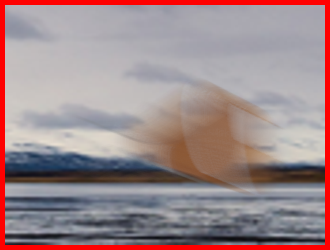}
   &&\includegraphics[width=\linewidth]{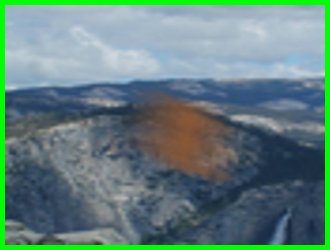}
   &\includegraphics[width=\linewidth]{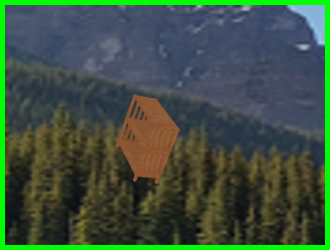}
		\\

  \includegraphics[width=\linewidth]{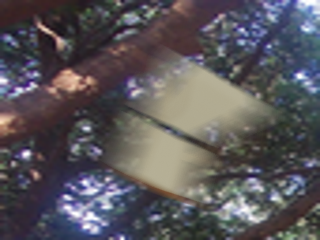}
  &&  \includegraphics[width=\linewidth]{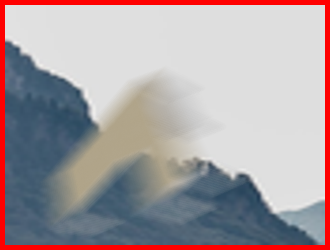}
    &  \includegraphics[width=\linewidth]{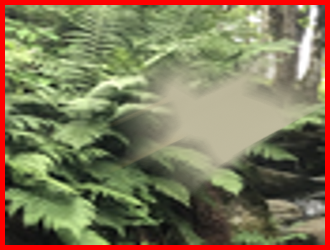}
   &   \includegraphics[width=\linewidth]{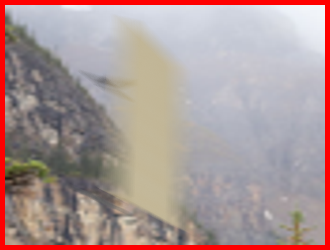}
   &   \includegraphics[width=\linewidth]{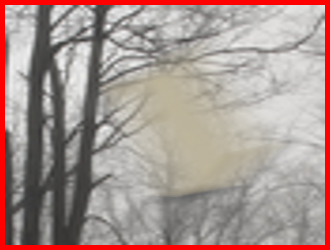}
   &   \includegraphics[width=\linewidth]{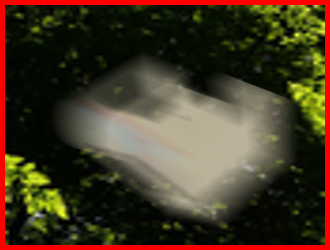}
   &   \includegraphics[width=\linewidth]{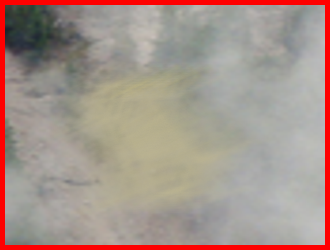}
   &   \includegraphics[width=\linewidth]{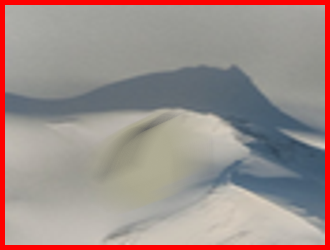}
   &   \includegraphics[width=\linewidth]{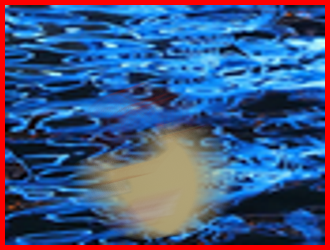}
   &&\includegraphics[width=\linewidth]{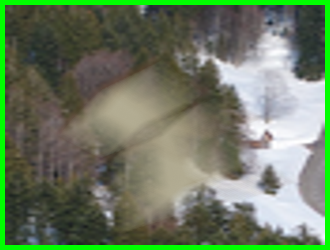}
   &\includegraphics[width=\linewidth]{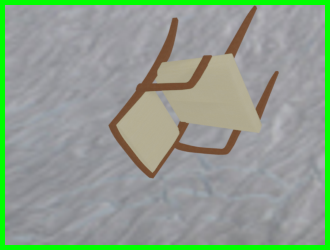}
		\\
	\end{tabular}
    \vspace{-1ex}

 \hfill
\vspace{-0.5em}

\caption{\textbf{Examples of hard distractors} in our distractor set. Each row displays a query image on the left, 8 hard distractors in the middle (marked with red boxes), and 2 matching images from database on the right (1 blurred and 1 sharp, marked with green boxes).}
\label{fig:distractor_examples}
    \vspace{0.5em}

\end{figure*}

\section{More per blur level database results}
\subsection{Per BL database (without distractors)}
\begin{figure*}
    \centering
    \begin{subfigure}{0.48\textwidth}
        \centering
        \includegraphics[width=\linewidth]{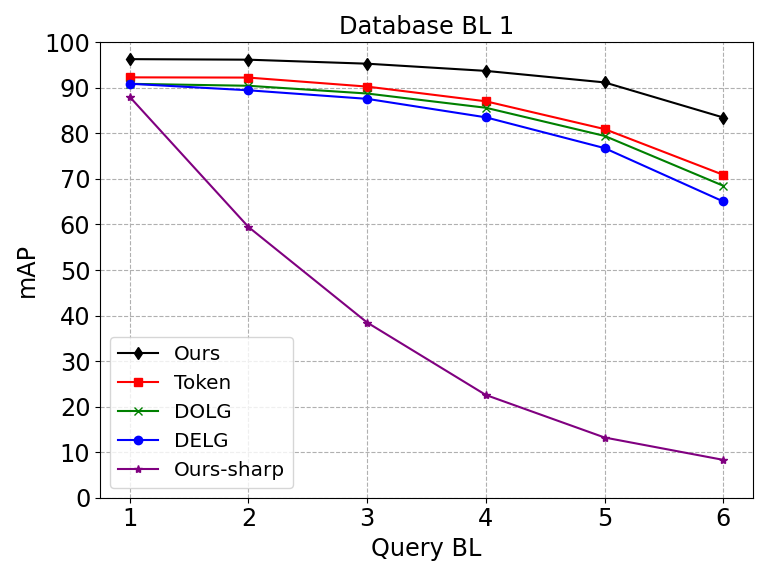}
        \label{fig:a}
    \end{subfigure}\hspace*{\fill}
    \begin{subfigure}{0.48\textwidth}
        \centering
        \includegraphics[width=\linewidth]{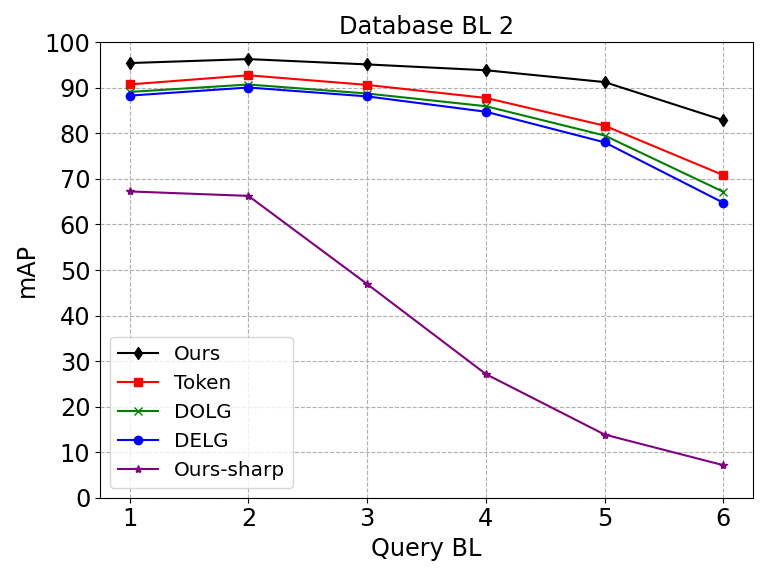}
        \label{fig:b}
    \end{subfigure}
    
    \medskip
    \begin{subfigure}{0.48\textwidth}
        \centering
        \includegraphics[width=\linewidth]{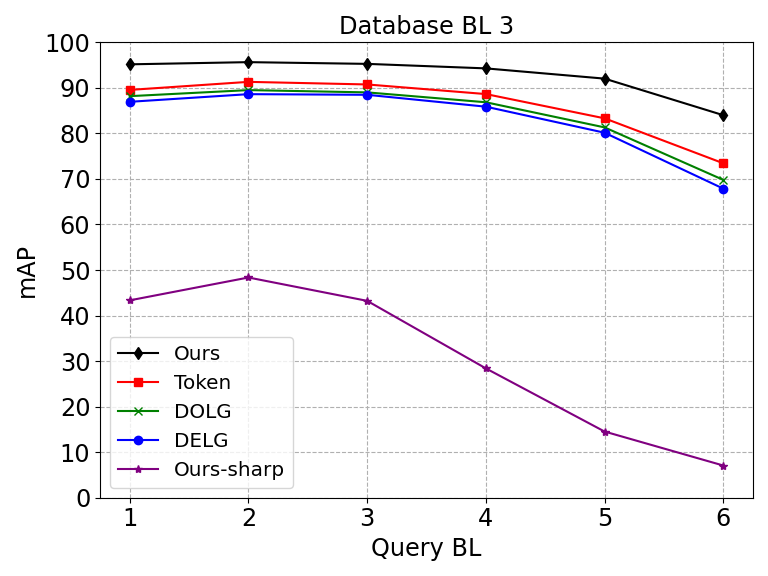}
        \label{fig:c}
    \end{subfigure}\hspace*{\fill}
    \begin{subfigure}{0.48\textwidth}
        \centering
        \includegraphics[width=\linewidth]{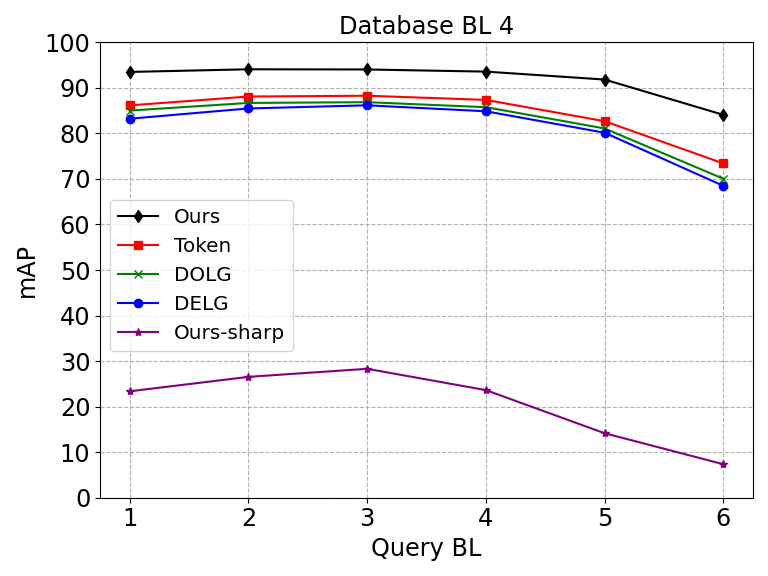}
        \label{fig:d}
    \end{subfigure}
    
    \medskip
    \begin{subfigure}{0.48\textwidth}
        \centering
        \includegraphics[width=\linewidth]{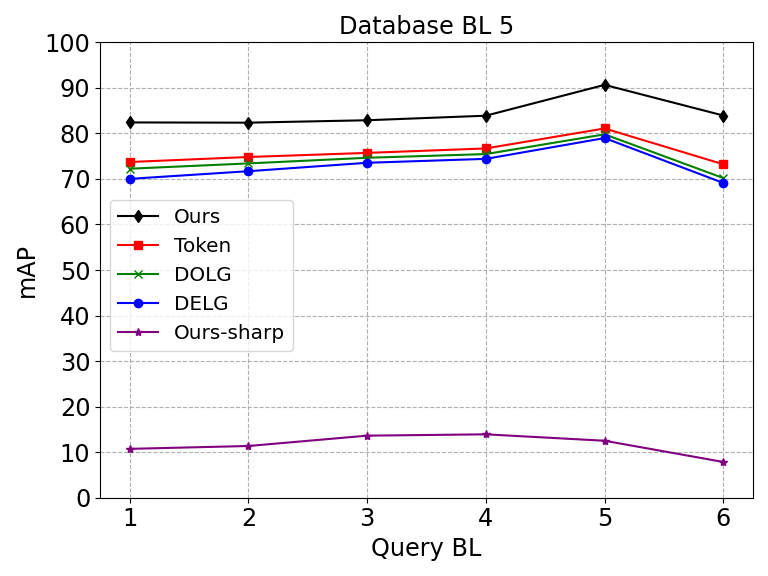}
        \label{fig:e}
    \end{subfigure}\hspace*{\fill}
    \begin{subfigure}{0.48\textwidth}
        \centering
        \includegraphics[width=\linewidth]{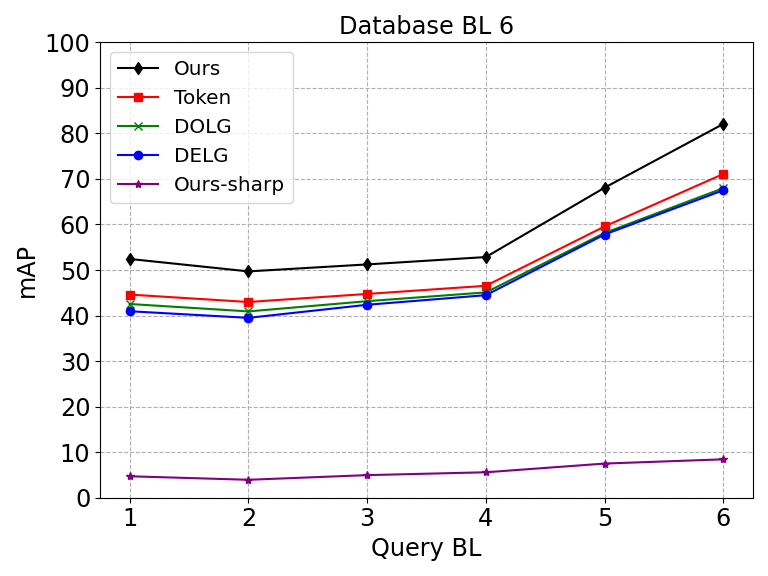}
        \label{fig:f}
    \end{subfigure}

\caption{\textbf{Retrieval results} of different methods by query and database blur level (BL) on our synthetic dataset (without 1M distractors). }
\label{fig:vis_perBL_no1M}
\end{figure*}

We show detailed per blur level database retrieval results of our method and state-of-the-art approaches on our synthetic dataset (without distractors) in Table \ref{tab:mAP_db_pure_diff_methods}. 
For a clearer representation of the trend in retrieval performance with varying database and query blur levels, as well as a more intuitive comparison among different methods, we cluster the retrieval results in Table \ref{tab:mAP_db_pure_diff_methods} according to the database blur level. The visualization of these results for each specific database blur level is presented in \Fig \ref{fig:vis_perBL_no1M}, where we additionally provide results for Ours-sharp (the version of our model trained with only sharp images). 

Figure \ref{fig:vis_perBL_no1M} clearly shows that our method outperforms others across all settings.
Notably, at lower database blur levels, the performance gap between our method and others becomes more noticeable as the query blur level increases, while at higher database blur levels, this gap remains pronounced for both low and high query blur levels. 
This highlights the significant advantage of our method in handling images affected by severe object motion blur. 
Furthermore, it can be observed that our method exhibits overall minimal performance fluctuations with changing query blur levels across various database blur levels, as evident from the range and standard deviation values in Table \ref{tab:mAP_db_pure_diff_methods}, indicating that our method has the highest robustness for retrieving images in the presence of object motion blur compared to state-of-the-art retrieval methods.

Regarding the model trained solely with sharp images, it produces reasonable results only when both the database and query blur levels are low. Its performance rapidly declines as the database blur level increases, which emphasizes the significance of training with images containing motion-blurred objects for effective retrieval in the context of object motion blur.

\begin{table*}[t]
     \scriptsize	 
    \centering
    \begin{adjustbox}{max width=\textwidth}
    \setlength{\tabcolsep}{1.8pt} 
    \begin{tabularx}{\textwidth}{P{0.5cm} c ccccccc c ccccccc}
        \toprule
        {BL}  && \multicolumn{7}{c}{DELG~\cite{delg} (Range: 51.41; Std: 14.54)}
        && \multicolumn{7}{c}{DOLG~\cite{dolg} (Range: 49.99; Std: 14.42)} \\  \cline{3-9} \cline{11-17}
          \diagbox[innerwidth = 0.5cm, height=0.5cm]{D}{Q} &&All 
         &1  &2 &3  &4 &5 &6 &&All&  1 &  2&  3&  4&  5 & 6\\
         \midrule

        1 && 84.52 & 90.91 & 89.47 & 87.59 & 83.53 & 76.77 & 65.05
          && 85.99 & 90.89 & 90.46 & 88.76 & 85.63 & 79.45 & 68.50
        \\
        2 && 84.60 & 88.30 & 90.08 & 88.13 & 84.76 & 78.04 & 64.78
          && 85.65 & 89.11 & 90.74 & 88.76 & 85.97 & 79.54 & 67.18
        \\
        3 && 84.87 & 86.94 & 88.62 & 88.48 & 85.87 & 80.15 & 67.87
          && 85.90 & 88.18 & 89.51 & 89.01 & 86.84 & 81.32 & 69.79
        \\
        4 && 82.93 & 83.23 & 85.48 & 86.17 & 84.86 & 80.15 & 68.49
          && 84.09 & 85.02 & 86.69 & 86.86 & 85.76 & 81.09 & 70.04
        \\
        5 && 73.05 & 70.02 & 71.70 & 73.55 & 74.41 & 78.98 & 69.12
          && 74.45 & 72.25 & 73.41 & 74.65 & 75.46 & 79.80 & 70.20
        \\
        6 && 46.12 & 40.95 & 39.50 & 42.37 & 44.48 & 57.80 & 67.52
          && 47.06 & 42.54 & 40.91 & 43.16 & 45.09 & 58.14 & 67.98
        \\

        \midrule
        {BL}  && \multicolumn{7}{c}{Token \cite{token} (Range: 49.77; Std: 14.27)} && \multicolumn{7}{c}{Ours (Range: \textbf{46.62}; Std: \textbf{13.51})} \\  \cline{3-9} \cline{11-17}
          \diagbox[innerwidth = 0.5cm, height=0.5cm]{D}{Q} &&All 
         &1  &2 &3  &4 &5 &6 &&All&  1 &  2&  3&  4&  5 & 6\\
         \midrule
        1 && 87.59 & 92.31 & 92.26 & 90.28 & 87.05 & 80.95 & 70.90

          &&\textbf{93.85} & \textbf{96.31} & \textbf{96.19} & \textbf{95.31} & \textbf{93.73} & \textbf{91.20} & \textbf{83.49}
        \\
        2 &&  87.68 & 90.76 & 92.76 & 90.65 & 87.79 & 81.70 & 70.84

          && \textbf{93.66} & \textbf{95.45} & \textbf{96.32} & \textbf{95.15} & \textbf{93.87} & \textbf{91.27} & \textbf{82.90}
        \\
        3 && 87.78 & 89.55 & 91.32 & 90.75 & 88.64 & 83.31 & 73.44

          && \textbf{93.78} & \textbf{95.17} & \textbf{95.65} & \textbf{95.27} & \textbf{94.28} & \textbf{92.03} & \textbf{84.00}
        \\
        4 &&  85.66 & 86.15 & 88.10 & 88.27 & 87.36 & 82.68 & 73.41

          &&\textbf{92.74} & \textbf{93.50} & \textbf{94.08} & \textbf{94.04} & \textbf{93.57} & \textbf{91.83} & \textbf{84.08}

        \\
        5 &&  75.88 & 73.72 & 74.82 & 75.72 & 76.70 & 81.09 & 73.22
          &&\textbf{84.05} & \textbf{82.41} & \textbf{82.36} & \textbf{82.89} & \textbf{83.89} & \textbf{90.66} & \textbf{83.93}

        \\
        6 &&  48.91 & 44.61 & 42.98 & 44.75 & 46.54 & 59.57 & 71.09

          &&\textbf{56.32} & \textbf{52.43} & \textbf{49.70} & \textbf{51.23} & \textbf{52.85} & \textbf{68.08} & \textbf{82.08}

        \\
        
        \bottomrule
        
    \end{tabularx}
    
    \end{adjustbox}
    \vspace{0.8em} 
    \caption{\textbf{Retrieval results} by query (Q) and database (D) blur level (BL) on our synthetic dataset (without 1M distractors). 
'All' denotes the overall performance for queries of all blur levels. 
The standard deviation and range of mAP values are displayed alongside each method.
The best scores under the same settings are marked in bold. 
}
\vspace{-0.8em} 
    \label{tab:mAP_db_pure_diff_methods}
\end{table*}

\subsection{Per BL database (with distractors)}
\begin{figure*}
    \centering
    \begin{subfigure}{0.48\textwidth}
        \centering
        \includegraphics[width=\linewidth]{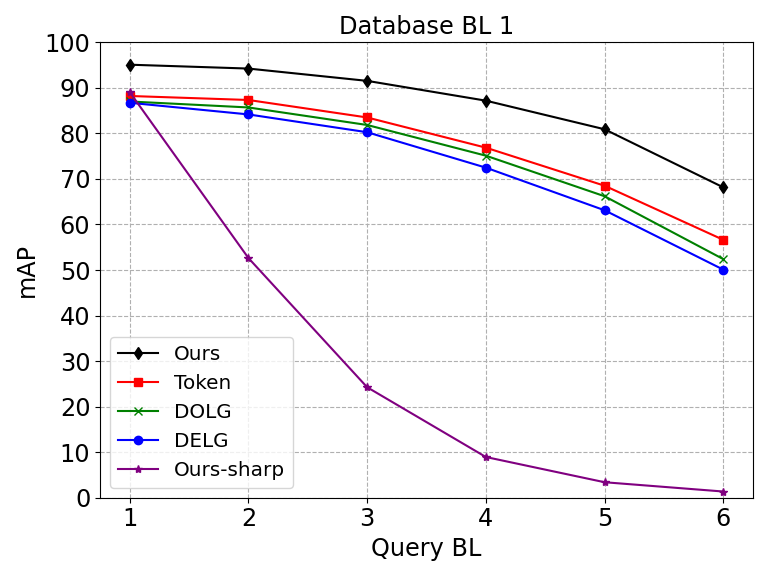}
        \label{fig:a}
    \end{subfigure}\hspace*{\fill}
    \begin{subfigure}{0.48\textwidth}
        \centering
        \includegraphics[width=\linewidth]{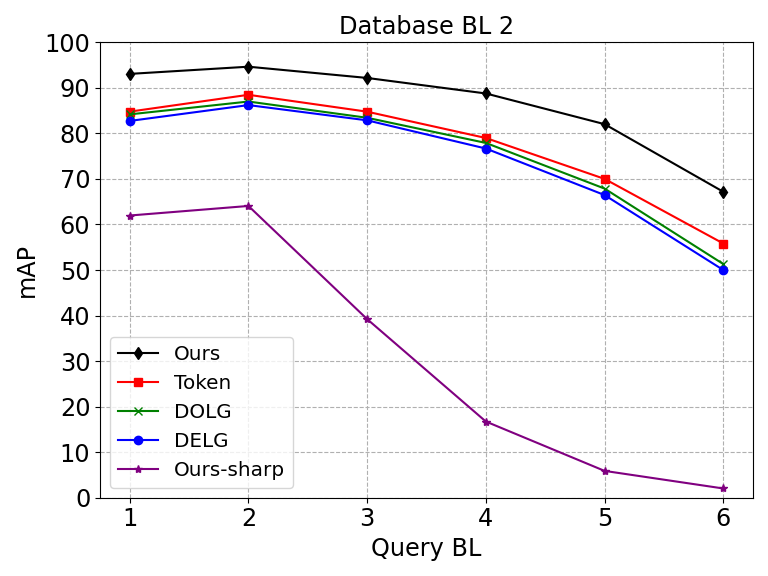}
        \label{fig:b}
    \end{subfigure}
    
    \medskip
    \begin{subfigure}{0.48\textwidth}
        \centering
        \includegraphics[width=\linewidth]{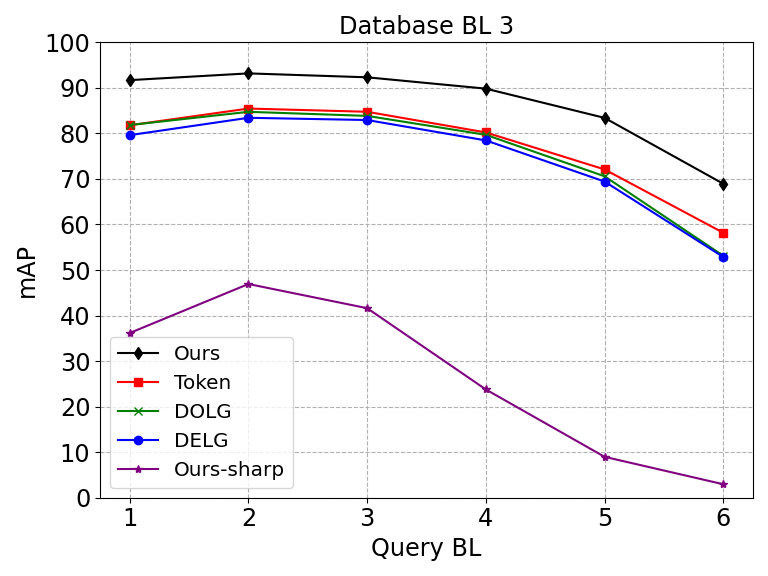}
        \label{fig:c}
    \end{subfigure}\hspace*{\fill}
    \begin{subfigure}{0.48\textwidth}
        \centering
        \includegraphics[width=\linewidth]{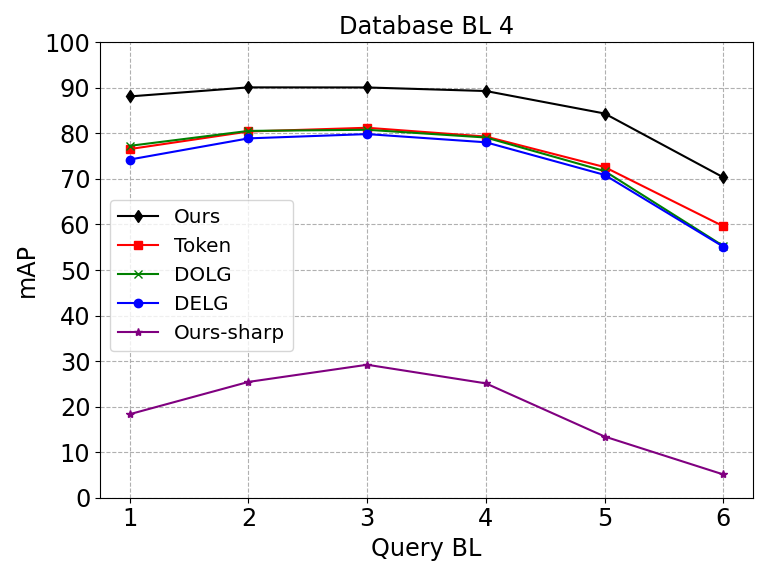}
        \label{fig:d}
    \end{subfigure}
    
    \medskip
    \begin{subfigure}{0.48\textwidth}
        \centering
        \includegraphics[width=\linewidth]{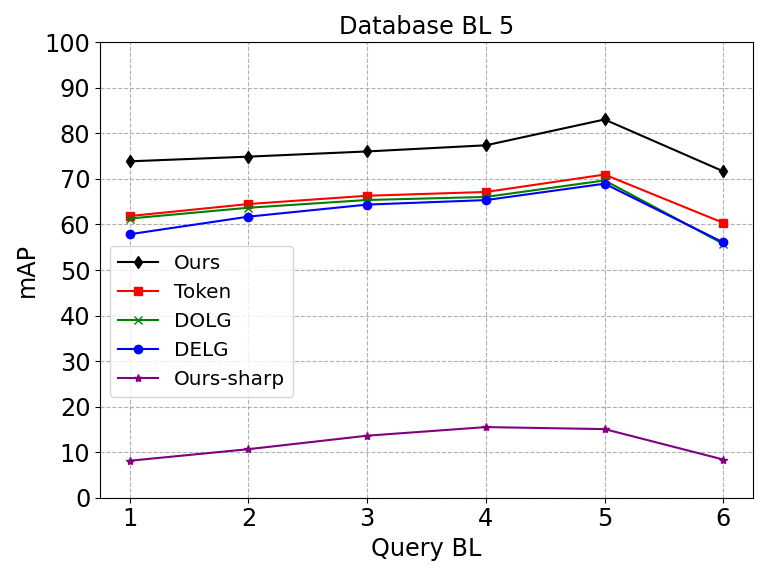}
        \label{fig:e}
    \end{subfigure}\hspace*{\fill}
    \begin{subfigure}{0.48\textwidth}
        \centering
        \includegraphics[width=\linewidth]{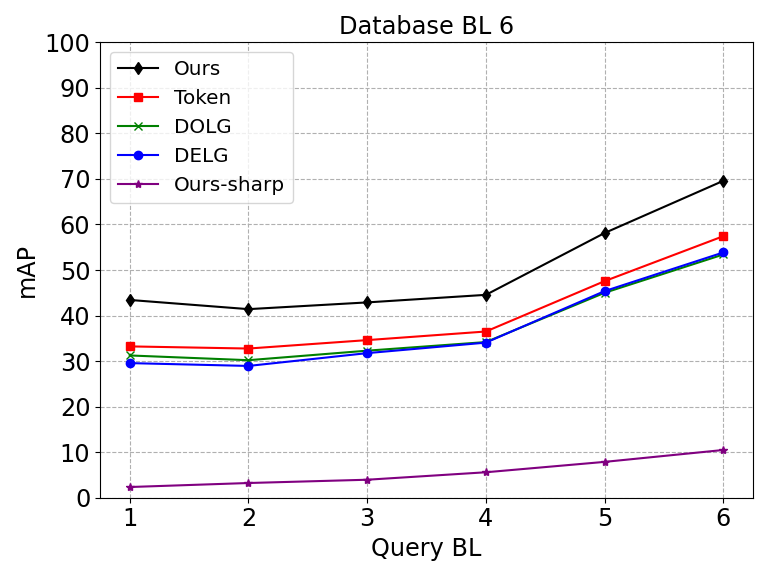}
        \label{fig:f}
    \end{subfigure}

\caption{\textbf{Retrieval results} of different methods by query and database blur level (BL) on our synthetic dataset (with 1M distractors). }
\label{fig:vis_perBL}
\end{figure*}

We further present per blur level database retrieval results with the addition of 1M distractors in \Fig \ref{fig:vis_perBL}. This illustration showcases the performance of our model in large-scale retrieval. It should be noted that when both the database blur level and query blur level are 1, the model trained exclusively with sharp images attains a satisfactory score of 89.07, which surpasses the results of state-of-the-art methods (DELG \cite{delg}: 86.73, DOLG \cite{dolg}: 87.01, Token \cite{token}: 88.22), and our model is the sole solution that outperforms it (95.09). This observation suggests that even in large-scale settings and evaluated with mAP@100, training with blurred images does not compromise our model's retrieval performance for sharp or near-sharp images.

\section{More ablation studies}
We conducted extensive experiments on our synthetic dataset with 1M distractors to investigate the impact of different loss components on the retrieval performance of images with different blur levels. 
The models are categorized into three groups based on the loss function(s) directly applied to the descriptor: those employing solely the contrastive loss $\mathcal{L}_{con}$, those utilizing only the classification loss $\mathcal{L}_{cls}$, and those incorporating both.
The detailed results are shown in Table \ref{tab:mAP_db_pure_contra} and visualizations of results for each database blur level can be found in \Fig \ref{fig:loss_ablation_perBL}.

The results of our experiments reveal that within each group of models, utilizing either blur estimation loss $\mathcal{L}_{be}$ or localization loss $\mathcal{L}_{loc}$ alone enhances the model's performance across all blur level settings. 
Moreover, combining both $\mathcal{L}_{be}$ and $\mathcal{L}_{loc}$ leads to superior overall performance. 
Additionally, models applying only classification loss directly to the descriptor outperform their counterparts with only contrastive loss directly applied to the descriptor. When compared to models using only $\mathcal{L}_{con}$ or $\mathcal{L}_{cls}$ directly on the descriptor, the combination of both losses applied to the descriptor results in better overall performance. 
Ultimately, the best overall retrieval performance is achieved when incorporating all four loss components.

\section{More qualitative results}
In this section, we show more qualitative results for our approach and all compared state-of-the-art retrieval methods in \Fig \ref{fig:qualitative_suppmat_syn} for synthetic data (with 1M distractors) and in \Fig \ref{fig:qualitative_suppmat_real} for real-world data.

These examples highlight the efficacy of our method in object retrieval under the influence of motion blur.
State-of-the-art methods tend to retrieve a substantial number of negative images for motion-blurred queries, whereas our approach excels in identifying more positive matches. 
Furthermore, when comparing the blur levels in the retrieved positive images between our method and others, it can be observed that the images retrieved by our method cover a wider range of blur levels, indicating the superior performance of our method in learning blur-invariant image representations compared to state-of-the-art retrieval approaches.

Additionally, the first examples in \Fig \ref{fig:qualitative_suppmat_syn} and \Fig \ref{fig:qualitative_suppmat_real} demonstrate our model's ability to distinguish intra-class similar objects, while the subsequent examples illustrate its capability for inter-class distinctions.
In scenarios where the query object is severely blurred by motion, the retrieval by other methods is notably affected by objects of the same class or, in cases of different classes, by objects with similar textures or shapes. In contrast, our model distinguishes between similar intra-class objects or objects with comparable textures/shapes with higher accuracy, demonstrating its effectiveness in challenging retrieval scenarios.

\section{Comparison with additional methods}
We compare our method with more state-of-the-art global standard retrieval methods, including CVNet~\cite{lee2022cv}, SuperGlobal~\cite{shao2023global} and SENet~\cite{lee2023senet}, on both our synthetic dataset (without distractors) and our real-world dataset. All methods are re-trained using our training dataset. A single scale and a descriptor size of 128 are used for all methods. For methods using momentum contrast~\cite{he2020moco}, we scale the momentum contrastive queue length to 4096 based on the size of the training datasets. All other settings are maintained at their default values. Our approach demonstrates superior performance over the compared methods, as illustrated by the results in Table~\ref{tab:compare_more_sota_syn} for synthetic data and Table~\ref{tab:compare_more_sota_real} for real data.

\begin{table}[t]
    \centering
    \begin{adjustbox}{max width=\linewidth}
    \setlength{\tabcolsep}{7.5pt}

    \begin{tabular}{lc c cccccc}
        \toprule
        \multirow{2}{*}{Method} & {mAP}  && \multicolumn{6}{c}{mAP by Query BL}  \\ \cline{4-9}
         &All   && 1 &  2&  3&  4&  5 &  6\\
        \midrule
        SENet \cite{lee2023senet} & 76.41 && 79.25 
        & 80.39 
        & 79.75 
        & 77.16 
        & 70.86 
        & 59.55  \\
        CVNet \cite{lee2022cv} & 89.42 && 91.27 
        & 91.95 
        & 91.52 
        & 89.84 
        & 86.08 
        & 78.50 \\
        SuperGlobal \cite{shao2023global} & 89.29 && 91.18 
        & 91.85 
        & 91.41 
        & 89.69 
        & 85.93 
        & 78.27 \\ 
        Ours & \textbf{91.78} && \textbf{93.05} 
        & \textbf{93.48}  
        & \textbf{93.14} 
        & \textbf{92.25} 
        & \textbf{90.20} 
        & \textbf{82.86}\\
        \bottomrule
    \end{tabular}
    \end{adjustbox}

    \caption{\textbf{Comparison with additional methods} on the synthetic dataset (without distractors). 'All' denotes the overall performance for queries of all blur levels. All methods are re-implemented and re-trained on our synthetic training set. The database contains images of all blur levels. Our method outperforms other methods.}

    \label{tab:compare_more_sota_syn}
    \vspace{-1.2em}

\end{table}

\section{Discussions and future directions}
\boldparagraph{Diverse real backgrounds.} We propose the first large-scale datasets for blur retrieval. For the real-world dataset, in order to make it practical to achieve such a large scale ($>$13k) that allows for sufficient variations in motion, pose, scale, and blur conditions, we opt for similar backgrounds. While having similar backgrounds for positive samples might seem to simplify retrieval, the significantly larger number of negative samples with also similar backgrounds (effectively serving as hard negatives) increases retrieval difficulty. Moreover, although our backgrounds may appear simple, they represent a completely different domain from the training images' backgrounds, allowing us to assess the model's generalization ability. However, ideally, each image would have a dissimilar real-world background. Incorporating more varied real-world backgrounds is a potential future direction for our research.

\boldparagraph{Multiple objects in images.} Given that our work is the initial exploration of the problem of retrieval in the presence of object motion blur, the focus on single-object scenarios provides a foundational understanding of the challenges associated with this specific task. However, real-world scenarios can involve multiple objects with diverse appearances and motion characteristics. Future work could extend our proposed method to handle cases where multiple objects in an image exhibit different levels of motion blur. This extension would necessitate addressing complexities introduced by the interactions between multiple objects and their varying degrees of blur. Moreover, considering scenarios with multiple objects, which may overlap, could open avenues for investigating how the motion blur of one object influences the retrieval of another object within the scene. 

\boldparagraph{Multiple forms of blur.} Our study focuses on retrieval in the presence of object motion blur, which is a common occurrence in various visual scenes. However, other types of blur, such as camera motion-induced blur and out-of-focus blur, are also prevalent phenomena that degrade image quality and likely affect retrieval accuracy. To tackle the broader challenges posed by various types of blur and to enhance the adaptability of retrieval methods across a wider range of scenarios, future research could explore a unified and robust retrieval framework that can effectively handle the integration of multiple forms of blur. Investigating the interplay between different types of blur and their impact on retrieval performance could provide valuable insights into the development of more adaptive and resilient retrieval methodologies. 

\begin{table}[t] 
    \centering
    \begin{adjustbox}{max width=\linewidth}
    \setlength{\tabcolsep}{7.5pt} 
    \begin{tabular}{lc c cccccc}
        \toprule
        \multirow{2}{*}{Method} & {mAP}  && \multicolumn{6}{c}{mAP by Query BL\textsuperscript{r}}  \\ \cline{4-9}
         &All   && 1 &  2&  3&  4&  5 &  6\\
        \midrule
        SENet \cite{lee2023senet} & 48.81 && 52.29 
        & 51.45 
        & 48.95 
        & 46.77 
        & 48.90 
        & 37.83  \\
        CVNet \cite{lee2022cv} & 56.99 &&58.10
        & 64.26 
        & 60.88  
        & 57.52 
        & 51.41 
        & 34.94  \\
        SuperGlobal \cite{shao2023global} & 58.68 && \textbf{59.67}  
        & 66.20  
        & 62.93  
        & 59.54
        & 52.77 
        & 35.36\\ 
        Ours & \textbf{62.88}  && {57.50 }
        & \textbf{70.38} 
        & \textbf{66.77} 
        & \textbf{63.18} 
        & \textbf{64.48} 
        & \textbf{46.14}\\
        \bottomrule
    \end{tabular}
    \end{adjustbox}

    \caption{\textbf{Comparison with additional methods} on the real-world dataset (trained on synthetic without fine-tuning). 'All' denotes the overall performance for queries of all blur levels. The database contains images of all blur levels.}
    \label{tab:compare_more_sota_real}
    \vspace{-1.2em} 

\end{table}

\definecolor{mydarkgreen}{RGB}{0, 100, 0}
\definecolor{mydarkred}{RGB}{139, 0, 0}
\definecolor{mydarkblue}{RGB}{0, 0, 139}
\begin{table*}
    \scriptsize	 
    \centering
    \begin{adjustbox}{max width=\textwidth} 
    \setlength{\tabcolsep}{1.8pt} 
    \begin{tabularx}{\textwidth}{P{0.5cm} c ccccccc c ccccccc}

        \toprule[0.5mm]

        {BL}  && \multicolumn{7}{c}{$\mathcal{L}_{con}$}
        && \multicolumn{7}{c}{$\mathcal{L}_{con} + \mathcal{L}_{be}$}
        \\  \cline{3-9} \cline{11-17}
         \diagbox[innerwidth = 0.5cm, height=0.5cm]{D}{Q} &&All 
         &1  &2 &3  &4 &5 &6 &&All&  1 &  2&  3&  4&  5 & 6\\
         \midrule

        1 && 68.84 & 80.24 & 78.56 & 72.94 & 64.59 & 55.39 & 40.82
          && 71.59 & 81.70 & 80.17 & 75.61 & 68.55 & 59.37 & 44.64
        \\
        2 && 70.12 & 76.63 & 80.72 & 75.71 & 68.28 & 58.05 & 40.68
          && 74.01 & 78.63 & 83.48 & 79.41 & 73.41 & 63.43 & 46.83
        \\
        3 && 69.67 & 72.93 & 77.37 & 76.21 & 70.69 & 60.09 & 41.70
          && 73.11 & 75.06 & 80.33 & 79.56 & 74.68 & 64.54 & 46.63
        \\
        4 && 66.61 & 66.85 & 71.90 & 72.36 & 69.72 & 60.96 & 41.95
          && 70.50 & 69.37 & 75.55 & 75.94 & 74.08 & 65.90 & 47.62
        \\
        5 && 54.51 & 51.82 & 55.10 & 56.35 & 56.73 & 59.09 & 42.14
          && 58.89 & 55.06 & 59.81 & 60.79 & 61.42 & 63.66 & 47.42
        \\
        6 && 29.05 & 26.54 & 25.52 & 26.86 & 27.98 & 35.90 & 39.70
          && 31.57 & 28.21 & 28.10 & 29.23 & 30.02 & 39.16 & 44.55
        \\

        \addlinespace[2pt]
        \hdashline
        \addlinespace[2pt]
        {BL}  && \multicolumn{7}{c}{$\mathcal{L}_{con} + \mathcal{L}_{loc}$}
        && \multicolumn{7}{c}{$\mathcal{L}_{con} + \mathcal{L}_{be} + \mathcal{L}_{loc}$}
        \\  \cline{3-9} \cline{11-17}
         \diagbox[innerwidth = 0.5cm, height=0.5cm]{D}{Q} &&All 
         &1  &2 &3  &4 &5 &6 &&All&  1 &  2&  3&  4&  5 & 6\\
         \midrule
        
          1 && 76.22 & 83.90 & 83.05 & 79.58 & 74.06 & 66.53 & \textcolor{mydarkblue}{54.59} 
          && \textcolor{mydarkblue}{77.20} & \textcolor{mydarkblue}{85.30} & \textcolor{mydarkblue}{84.10} & \textcolor{mydarkblue}{80.48} & \textcolor{mydarkblue}{75.09} & \textcolor{mydarkblue}{68.30} & 53.08 \\
        2 && 76.93 & 81.59 & 84.54 & 81.33 & 76.06 & 68.10 & 54.03 
          && \textcolor{mydarkblue}{78.93} & \textcolor{mydarkblue}{83.30} & \textcolor{mydarkblue}{86.62} & \textcolor{mydarkblue}{83.16} & \textcolor{mydarkblue}{78.44} & \textcolor{mydarkblue}{71.16} & \textcolor{mydarkblue}{54.21} \\
        3 && 76.16 & 78.59 & 82.02 & 81.22 & 76.88 & 69.16 & 54.40 
          && \textcolor{mydarkblue}{78.66} & \textcolor{mydarkblue}{81.22} & \textcolor{mydarkblue}{84.55} & \textcolor{mydarkblue}{83.51} & \textcolor{mydarkblue}{79.54} & \textcolor{mydarkblue}{72.28} & \textcolor{mydarkblue}{55.59} \\
        4 && 73.81 & 73.83 & 77.56 & 78.04 & 76.27 & 69.96 & 55.41 
          && \textcolor{mydarkblue}{76.54} & \textcolor{mydarkblue}{77.00} & \textcolor{mydarkblue}{80.67} & \textcolor{mydarkblue}{80.61} & \textcolor{mydarkblue}{78.77} & \textcolor{mydarkblue}{72.81} & \textcolor{mydarkblue}{56.85} \\
        5 && 63.08 & 60.48 & 62.81 & 64.05 & 64.84 & 68.05 & 54.97 
          && \textcolor{mydarkblue}{65.93} & \textcolor{mydarkblue}{63.66} & \textcolor{mydarkblue}{66.30} & \textcolor{mydarkblue}{67.14} & \textcolor{mydarkblue}{67.14} & \textcolor{mydarkblue}{70.75} & \textcolor{mydarkblue}{56.55} \\
        6 && 36.76 & 33.48 & 32.19 & 33.74 & 35.21 & 45.31 & 52.01 
          && \textcolor{mydarkblue}{39.08} & \textcolor{mydarkblue}{35.12} & \textcolor{mydarkblue}{35.28} & \textcolor{mydarkblue}{36.52} & \textcolor{mydarkblue}{37.29} & \textcolor{mydarkblue}{47.65} & \textcolor{mydarkblue}{53.76}

        \\

        \midrule[0.5mm]
{BL}  && \multicolumn{7}{c}{$\mathcal{L}_{cls}$}
        && \multicolumn{7}{c}{$\mathcal{L}_{cls} + \mathcal{L}_{be}$}
        \\  \cline{3-9} \cline{11-17}
         \diagbox[innerwidth = 0.5cm, height=0.5cm]{D}{Q} &&All 
         &1  &2 &3  &4 &5 &6 &&All&  1 &  2&  3&  4&  5 & 6\\
         \midrule

        1 && 74.00 & 88.92 & 86.00 & 79.61 & 68.22 & 55.62 & 40.36
          && 78.37 & 92.04 & 89.31 & 83.26 & 73.49 & 61.37 & 47.52
        \\
        2 && 75.61 & 85.25 & 87.75 & 82.73 & 73.36 & 59.14 & 39.45
          && 81.12 & 89.09 & 91.45 & 87.15 & 79.51 & 67.28 & 50.00
        \\
        3 && 75.65 & 80.81 & 84.28 & 83.55 & 76.88 & 63.16 & 41.84
          && 81.33 & 85.50 & 88.74 & 88.02 & 82.66 & 70.95 & 51.92
        \\
        4 && 72.48 & 72.60 & 77.40 & 79.04 & 76.52 & 66.41 & 45.51
          && 78.82 & 78.66 & 83.25 & 84.41 & 82.54 & 73.87 & 54.86
        \\
        5 && 58.59 & 53.54 & 57.33 & 60.53 & 62.00 & 65.47 & 49.45
          && 65.89 & 60.81 & 64.65 & 67.29 & 69.11 & 73.28 & 57.67
        \\
        6 && 30.00 & 24.94 & 23.56 & 26.59 & 29.15 & 40.88 & 49.35
          && 36.83 & 31.73 & 30.71 & 33.38 & 35.56 & 47.76 & 56.63
        \\

        \addlinespace[2pt]
        \hdashline
         \addlinespace[2pt]
        {BL}  && \multicolumn{7}{c}{$\mathcal{L}_{cls} + \mathcal{L}_{loc}$}
        && \multicolumn{7}{c}{$\mathcal{L}_{cls} + \mathcal{L}_{be} + \mathcal{L}_{loc}$}
        \\  \cline{3-9} \cline{11-17}
         \diagbox[innerwidth = 0.5cm, height=0.5cm]{D}{Q} &&All 
         &1  &2 &3  &4 &5 &6 &&All&  1 &  2&  3&  4&  5 & 6\\
         \midrule
        
        1 && 83.27 & 90.92 & 89.68 & 86.60 & 81.12 & 73.93 & 62.18 
          && \textcolor{mydarkgreen}{86.29} & \textcolor{mydarkgreen}{92.72} & \textcolor{mydarkgreen}{91.90} & \textcolor{mydarkgreen}{89.14} & \textcolor{mydarkgreen}{84.65} & \textcolor{mydarkgreen}{78.21} & \textcolor{mydarkgreen}{67.90} \\
        2 && 84.22 & 88.96 & 91.14 & 88.25 & 83.83 & 75.99 & 61.38 
          && \textcolor{mydarkgreen}{86.97} & \textcolor{mydarkgreen}{90.79} & \textcolor{mydarkgreen}{93.02} & \textcolor{mydarkgreen}{90.55} & \textcolor{mydarkgreen}{86.49} & \textcolor{mydarkgreen}{80.00} & \textcolor{mydarkgreen}{\textbf{67.66}} \\
        3 && 84.23 & 86.89 & 89.20 & 88.69 & 85.43 & 77.95 & 63.02 
          && \textcolor{mydarkgreen}{86.82} & \textcolor{mydarkgreen}{88.83} & \textcolor{mydarkgreen}{91.27} & \textcolor{mydarkgreen}{90.74} & \textcolor{mydarkgreen}{87.61} & \textcolor{mydarkgreen}{81.46} & \textcolor{mydarkgreen}{\textbf{69.12}} \\
        4 && 82.53 & 82.75 & 85.61 & 86.28 & 85.12 & 79.18 & 65.19 
          && \textcolor{mydarkgreen}{85.22} & \textcolor{mydarkgreen}{85.12} & \textcolor{mydarkgreen}{88.19} & \textcolor{mydarkgreen}{88.59} & \textcolor{mydarkgreen}{87.23} & \textcolor{mydarkgreen}{82.20} & \textcolor{mydarkgreen}{\textbf{70.76}} \\
        5 && 71.03 & 67.31 & 69.44 & 71.63 & 72.86 & 77.83 & 66.50 
          && \textcolor{mydarkgreen}{74.47} & \textcolor{mydarkgreen}{71.02} & \textcolor{mydarkgreen}{73.24} & \textcolor{mydarkgreen}{74.65} & \textcolor{mydarkgreen}{75.88} & \textcolor{mydarkgreen}{80.79} & \textcolor{mydarkgreen}{71.23} \\
        6 && 41.87 & 36.45 & 35.29 & 38.18 & 40.20 & 53.16 & 64.45 
          && \textcolor{mydarkgreen}{45.96} & \textcolor{mydarkgreen}{40.82} & \textcolor{mydarkgreen}{39.76} & \textcolor{mydarkgreen}{42.05} & \textcolor{mydarkgreen}{43.68} & \textcolor{mydarkgreen}{57.04} & \textcolor{mydarkgreen}{69.29} 
        \\
        \midrule[0.5mm]
        {BL}  && \multicolumn{7}{c}{$\mathcal{L}_{con} + \mathcal{L}_{cls}$ } 
        && \multicolumn{7}{c}{$\mathcal{L}_{con} + \mathcal{L}_{cls} + \mathcal{L}_{be}$ }
         \\  \cline{3-9} \cline{11-17}
         \diagbox[innerwidth = 0.5cm, height=0.5cm]{D}{Q} &&All 
         &1  &2 &3  &4 &5 &6 &&All&  1 &  2&  3&  4&  5 & 6\\
         \midrule
        1 && 80.67 & 91.14 & 89.81 & 85.39 & 77.65 & 67.19 & 51.88
          && 82.75 & 92.28 & 91.15 & 87.19 & 80.07 & 71.03 & 54.90
        \\
        2 && 81.74 & 88.18 & 90.82 & 87.53 & 80.91 & 70.34 & 51.92
          && 83.83 & 89.37 & 92.35 & 89.03 & 83.39 & 73.94 & 55.40
        \\
        3  && 81.54 & 84.96 & 88.48 & 87.63 & 82.96 & 72.66 & 53.74
           && 83.72 & 86.75 & 90.16 & 89.32 & 85.17 & 75.85 & 57.48
        \\
        4 && 79.34 & 79.78 & 84.01 & 84.46 & 82.48 & 74.42 & 55.72
          && 81.65 & 81.70 & 86.18 & 86.47 & 84.80 & 77.28 & 59.19
        \\
        5 && 66.77 & 63.31 & 66.52 & 68.17 & 68.89 & 72.88 & 56.95
          && 69.68 & 65.72 & 69.22 & 71.13 & 72.12 & 75.96 & 60.41
        \\
        6 && 37.87 & 34.05 & 33.07 & 34.68 & 35.91 & 47.25 & 55.04
          && 39.22 & 34.70 & 33.70 & 35.84 & 37.53 & 49.43 & 58.18
        \\
        \addlinespace[2pt]
        \hdashline
        \addlinespace[2pt]
        {BL}  && \multicolumn{7}{c}{$\mathcal{L}_{con}+\mathcal{L}_{cls}+\mathcal{L}_{loc}$}
        && \multicolumn{7}{c}{$\mathcal{L}_{con}+\mathcal{L}_{cls}+\mathcal{L}_{be}+\mathcal{L}_{loc}$ }
        \\  \cline{3-9} \cline{11-17}
         \diagbox[innerwidth = 0.5cm, height=0.5cm]{D}{Q} &&All 
         &1  &2 &3  &4 &5 &6 &&All&  1 &  2&  3&  4&  5 & 6\\
         \midrule
        1 && 86.28 & 93.44 & 92.71 & 89.56 & 85.10 & 77.67 & 62.86
&& \textcolor{mydarkred}{\textbf{88.57}} & \textcolor{mydarkred}{\textbf{95.09}} & \textcolor{mydarkred}{\textbf{94.24}} & \textcolor{mydarkred}{\textbf{91.54}} & \textcolor{mydarkred}{\textbf{87.19}} & \textcolor{mydarkred}{\textbf{80.91}} & \textcolor{mydarkred}{\textbf{68.15}}
\\
2 && 86.77 & 91.42 & 93.52 & 90.56 & 86.99 & 79.13 & 62.62
&& \textcolor{mydarkred}{\textbf{88.75}} & \textcolor{mydarkred}{\textbf{93.07}} & \textcolor{mydarkred}{\textbf{94.65}} & \textcolor{mydarkred}{\textbf{92.18}} & \textcolor{mydarkred}{\textbf{88.77}} & \textcolor{mydarkred}{\textbf{82.03}} & \textcolor{mydarkred}{{67.18}}
\\
3 && 86.84 & 89.86 & 91.91 & 90.81 & 88.20 & 80.81 & 64.80
&& \textcolor{mydarkred}{\textbf{88.76}} & \textcolor{mydarkred}{\textbf{91.70}} & \textcolor{mydarkred}{\textbf{93.18}} & \textcolor{mydarkred}{\textbf{92.32}} & \textcolor{mydarkred}{\textbf{89.83}} & \textcolor{mydarkred}{\textbf{83.41}} & \textcolor{mydarkred}{{68.94}}
\\
4 && 85.28 & 86.03 & 88.71 & 88.63 & 87.72 & 82.04 & 66.90
&& \textcolor{mydarkred}{\textbf{87.16}} & \textcolor{mydarkred}{\textbf{88.12}} & \textcolor{mydarkred}{\textbf{90.11}} & \textcolor{mydarkred}{\textbf{90.09}} & \textcolor{mydarkred}{\textbf{89.31}} & \textcolor{mydarkred}{\textbf{84.37}} & \textcolor{mydarkred}{{70.35}}
\\
5 && 74.31 & 71.45 & 73.26 & 74.69 & 75.89 & 80.66 & 68.19
&& \textcolor{mydarkred}{\textbf{76.28}} & \textcolor{mydarkred}{\textbf{73.87}} & \textcolor{mydarkred}{\textbf{74.90}} & \textcolor{mydarkred}{\textbf{76.04}} & \textcolor{mydarkred}{\textbf{77.39}} & \textcolor{mydarkred}{\textbf{83.06}} & \textcolor{mydarkred}{\textbf{71.67}}
\\
6 && 45.21 & 40.98 & 39.22 & 41.32 & 43.03 & 55.74 & 66.50
&& \textcolor{mydarkred}{\textbf{47.31}} & \textcolor{mydarkred}{\textbf{43.43}} & \textcolor{mydarkred}{\textbf{41.41}} & \textcolor{mydarkred}{\textbf{42.89}} & \textcolor{mydarkred}{\textbf{44.55}} & \textcolor{mydarkred}{\textbf{58.13}} & \textcolor{mydarkred}{\textbf{69.59}}

        \\
        \bottomrule[0.5mm]
        
    \end{tabularx}
    \end{adjustbox}
    \vspace{0.8em} 
    \caption{\textbf{Ablation study} on the impact of different loss components on the retrieval performance of images with different blur levels (BLs), conducted on our synthetic dataset (with 1M distractors). The best scores under the same BL settings are marked in \textcolor{mydarkblue}{blue} (comparing models w/o $\mathcal{L}_{cls}$), \textcolor{mydarkgreen}{green} (comparing models w/o $\mathcal{L}_{con}$), \textcolor{mydarkred}{red} (comparing models with $\mathcal{L}_{cls} + \mathcal{L}_{con}$), or \textbf{bold} (comparing all models). 
    }
    \label{tab:mAP_db_pure_contra}
\end{table*}

\begin{figure*}
    \centering
    \begin{subfigure}{0.48\textwidth}
        \centering
        \includegraphics[width=\linewidth]{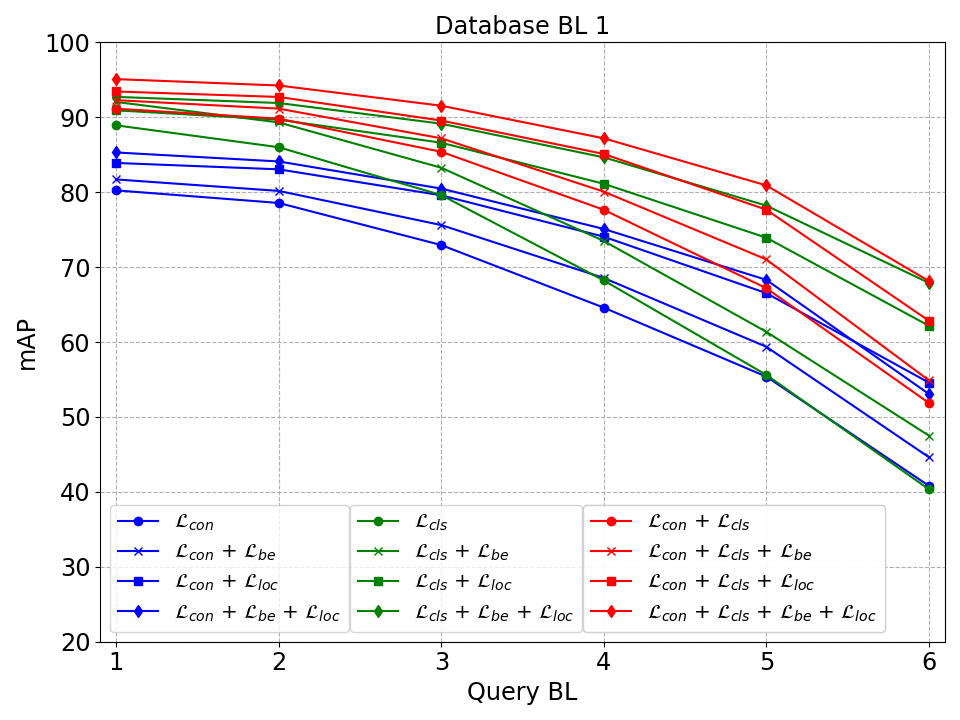}
        \label{fig:a}
    \end{subfigure}\hspace*{\fill}
    \begin{subfigure}{0.48\textwidth}
        \centering
        \includegraphics[width=\linewidth]{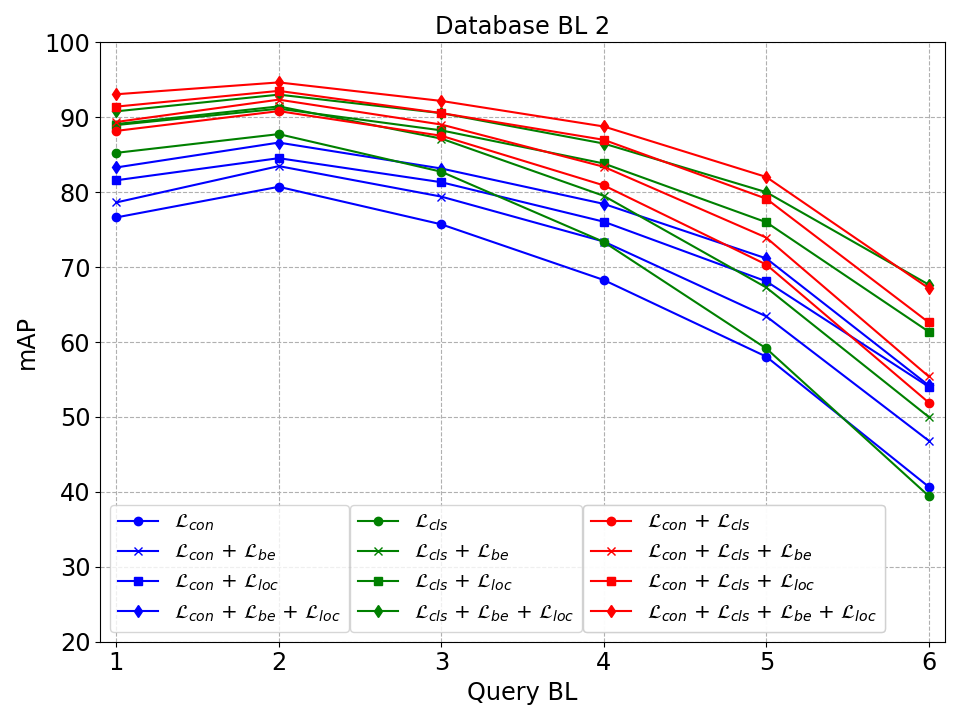}
        \label{fig:b}
    \end{subfigure}
    
    \medskip
    \begin{subfigure}{0.48\textwidth}
        \centering
        \includegraphics[width=\linewidth]{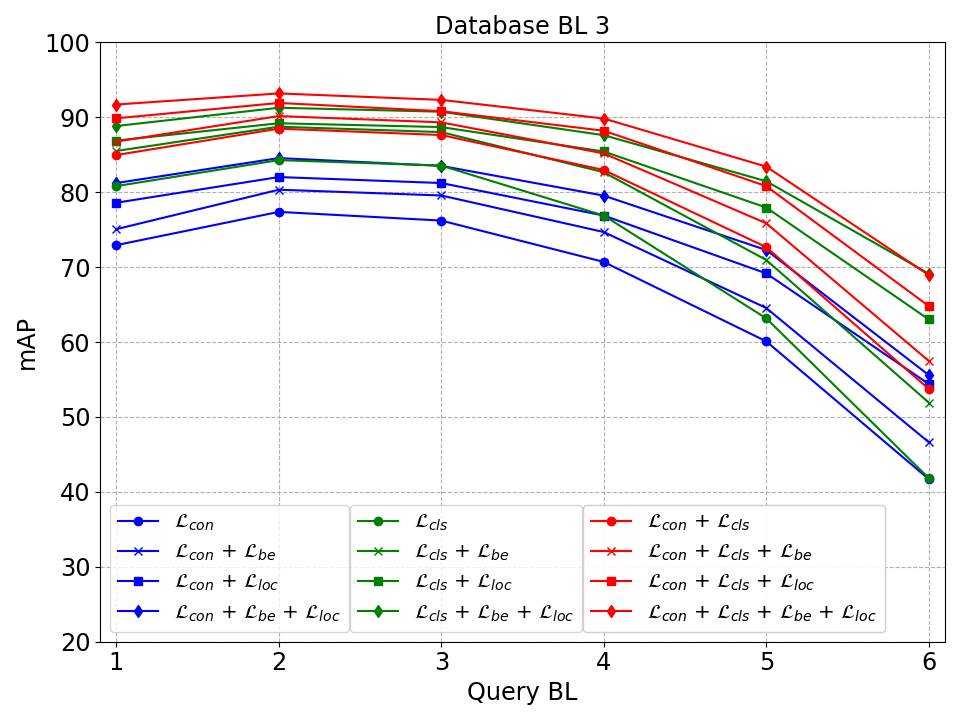}
        \label{fig:c}
    \end{subfigure}\hspace*{\fill}
    \begin{subfigure}{0.48\textwidth}
        \centering
        \includegraphics[width=\linewidth]{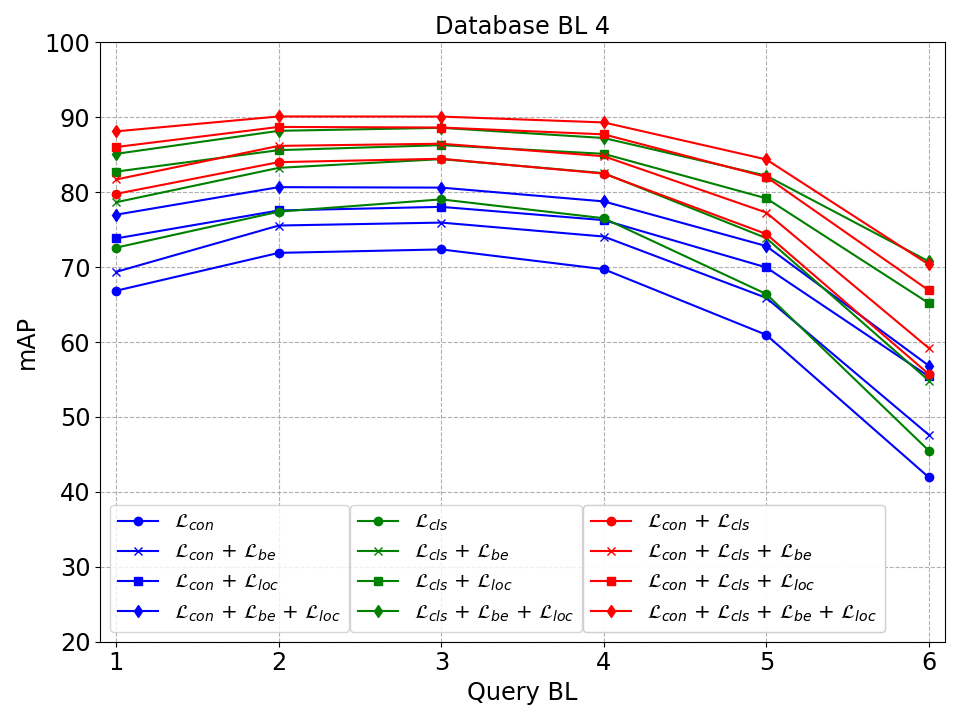}
        \label{fig:d}
    \end{subfigure}
    
    \medskip
    \begin{subfigure}{0.48\textwidth}
        \centering
        \includegraphics[width=\linewidth]{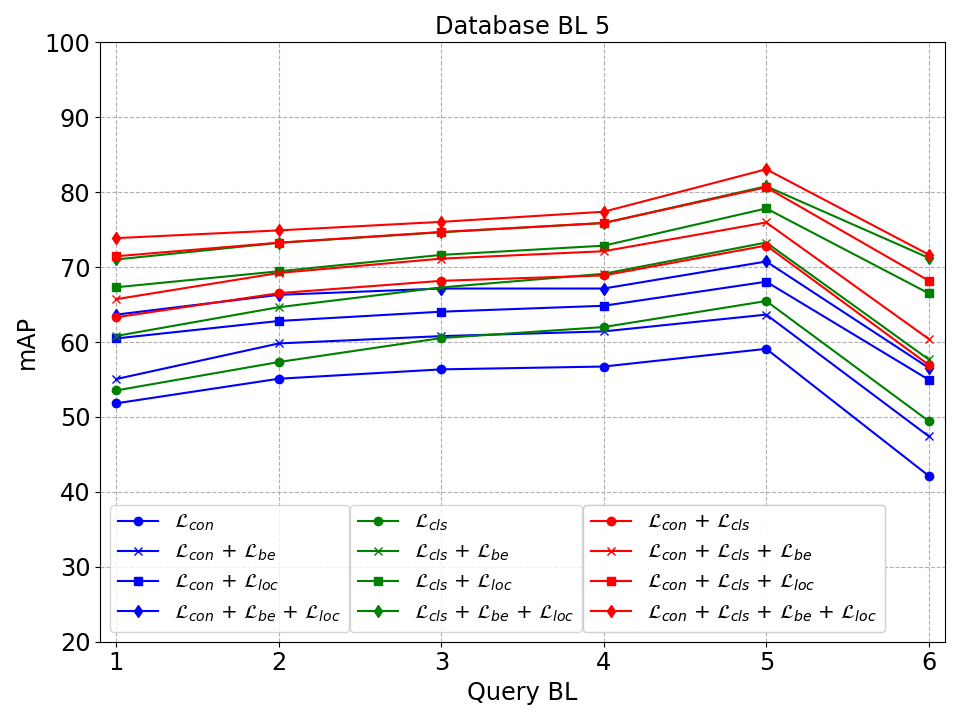}
        \label{fig:e}
    \end{subfigure}\hspace*{\fill}
    \begin{subfigure}{0.48\textwidth}
        \centering
        \includegraphics[width=\linewidth]{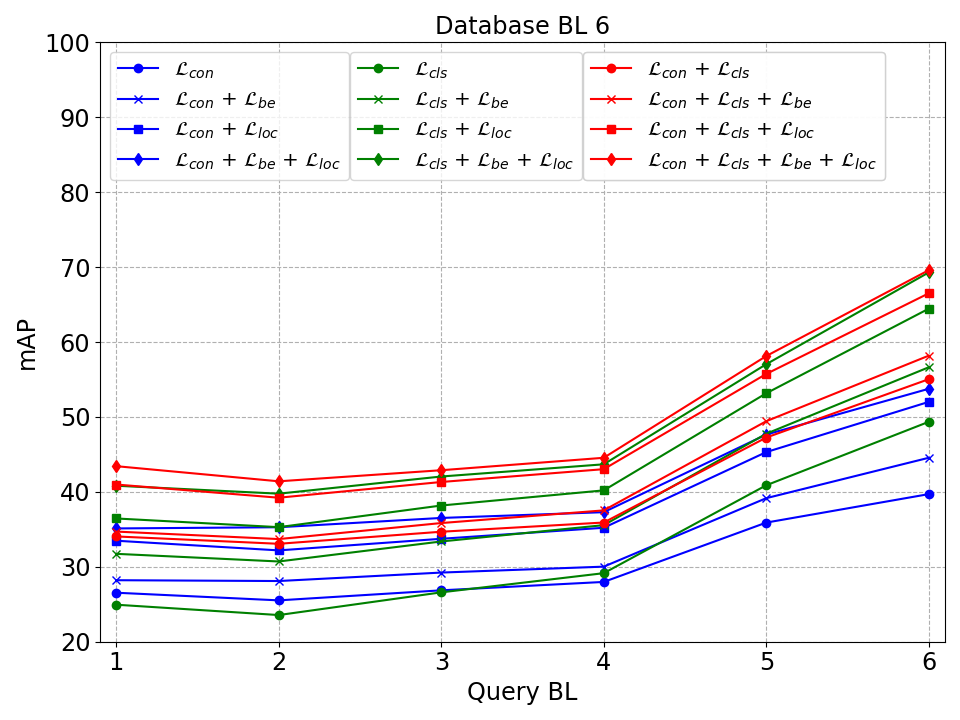}
        \label{fig:f}
    \end{subfigure}

\caption{\textbf{Ablation study} investigating the influence of distinct loss components on the retrieval performance of images with different blur levels (BLs), carried out on our synthetic dataset (with 1M distractors). The results for each database BL are categorized into three groups: \textcolor{mydarkblue}{models without $\mathcal{L}_{cls}$, marked in blue}, \textcolor{mydarkgreen}{models without $\mathcal{L}_{con}$, marked in green}, and \textcolor{mydarkred}{models with both $\mathcal{L}_{cls}$ and $\mathcal{L}_{con}$, marked in red}.}

\label{fig:loss_ablation_perBL}
\end{figure*}

 \global\long\def\figWidth{0.077\linewidth} 
\begin{figure*}
	\centering
 \begin{subfigure}{\linewidth}
    \setlength{\tabcolsep}{2pt}
    \setlength{\fboxrule}{1pt} 
    \setlength{\fboxsep}{0pt} 
	\begin{tabular}{
	 P{\figWidth}
	P{0.2cm}
	P{\figWidth}
	P{\figWidth}
    P{\figWidth}
    P{\figWidth}
    P{\figWidth}
    P{\figWidth}
	P{\figWidth}
    P{\figWidth}
    P{\figWidth}
    P{\figWidth}
    }
	\multicolumn{2}{c}{Query}	&   \multicolumn{10}{c}{Top 20 retrieval results}
		\\
  \addlinespace[5pt]
  
\raisebox{0em}{\multirow{2}{*}{ 
\addimgTtext{\includegraphics[width=\linewidth]{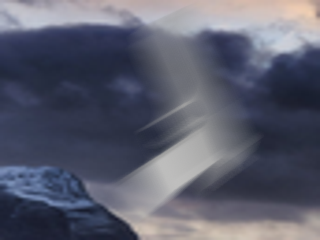}}{6}}}
&\raisebox{1.1em}{\multirow{2}{*}{\rotatebox[origin=c]{90}{DELG~\cite{delg}}}}
    &\addimgTtext{\includegraphics[width=\linewidth]{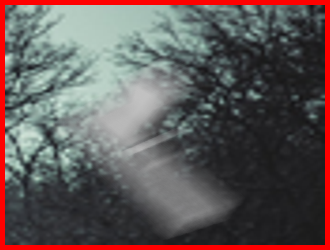}}{6}
      &\addimgTtext{\includegraphics[width=\linewidth]{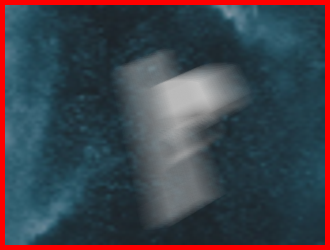}}{6}
      &\addimgTtext{\includegraphics[width=\linewidth]{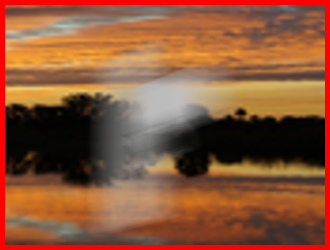}}{6}
      &\addimgTtext{\includegraphics[width=\linewidth]{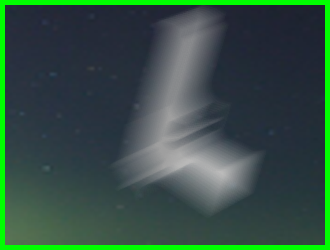}}{6}
      &\addimgTtext{\includegraphics[width=\linewidth]{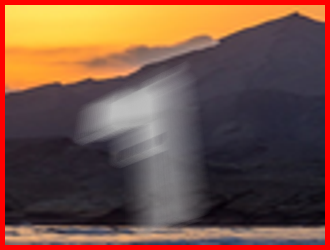}}{6}
      &\addimgTtext{\includegraphics[width=\linewidth]{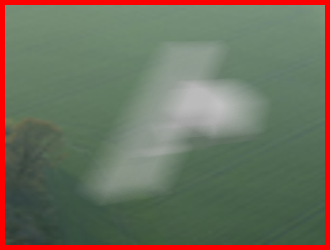}}{6}
      &\addimgTtext{\includegraphics[width=\linewidth]{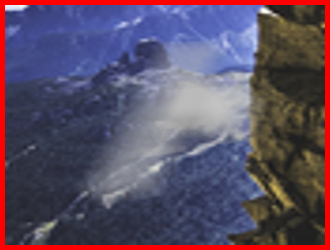}}{6}
      &\addimgTtext{\includegraphics[width=\linewidth]{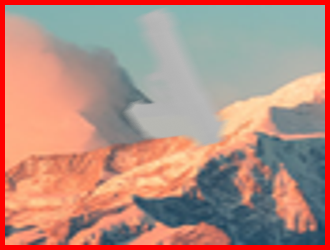}}{3}
      &\addimgTtext{\includegraphics[width=\linewidth]{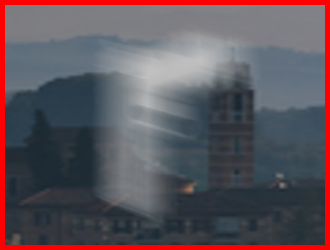}}{6}
      &\addimgTtext{\includegraphics[width=\linewidth]{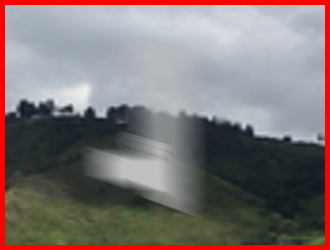}}{6}
		\\
    &&\addimgTtext{\includegraphics[width=\linewidth]{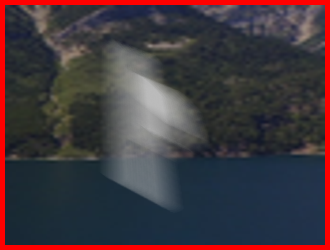}}{6}
      &\addimgTtext{\includegraphics[width=\linewidth]{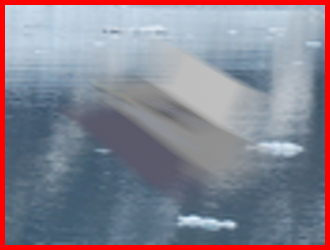}}{3}
      &\addimgTtext{\includegraphics[width=\linewidth]{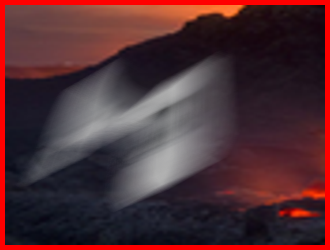}}{5}
      &\addimgTtext{\includegraphics[width=\linewidth]{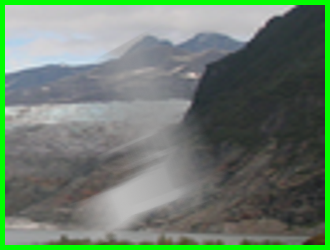}}{6}
      &\addimgTtext{\includegraphics[width=\linewidth]{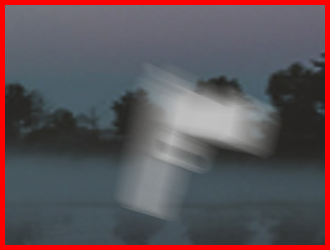}}{5}
      &\addimgTtext{\includegraphics[width=\linewidth]{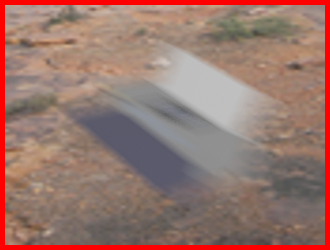}}{3}
      &\addimgTtext{\includegraphics[width=\linewidth]{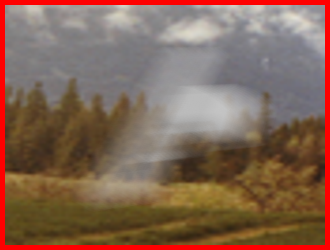}}{6}
      &\addimgTtext{\includegraphics[width=\linewidth]{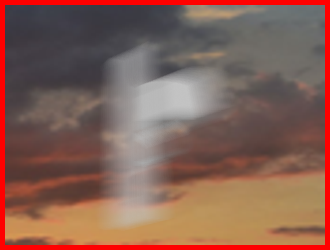}}{5}
      &\addimgTtext{\includegraphics[width=\linewidth]{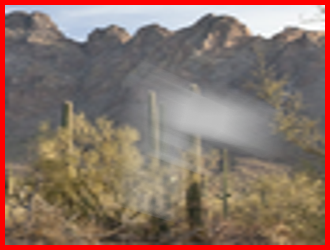}}{6}
      &\addimgTtext{\includegraphics[width=\linewidth]{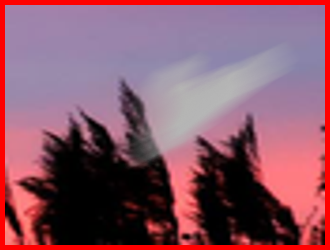}}{5} \\  \addlinespace[5pt]
\raisebox{0em}{\multirow{2}{*}{ 
\addimgTtext{\includegraphics[width=\linewidth]{fig/suppmat/qualitativeBL6_suppmat_gun/query_image_7071.png}}{6}}}
&\raisebox{1.10em}{\multirow{2}{*}{\rotatebox[origin=c]{90}{DOLG~\cite{dolg}}}}
    &\addimgTtext{\includegraphics[width=\linewidth]{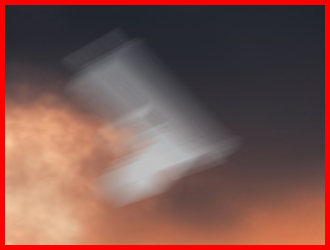}}{6}
      &\addimgTtext{\includegraphics[width=\linewidth]{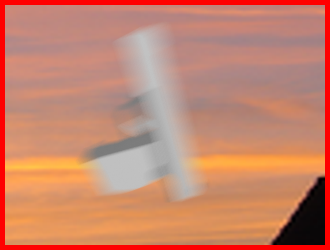}}{4}
      &\addimgTtext{\includegraphics[width=\linewidth]{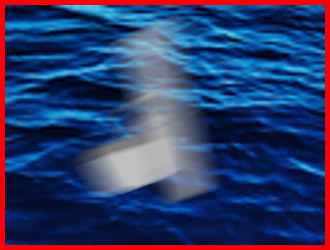}}{5}
      &\addimgTtext{\includegraphics[width=\linewidth]{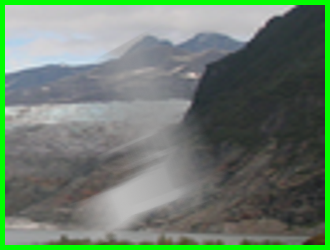}}{6}
      &\addimgTtext{\includegraphics[width=\linewidth]{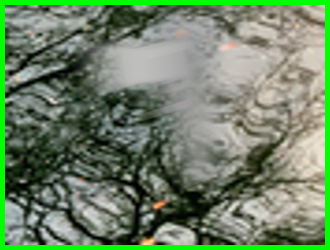}}{5}
      &\addimgTtext{\includegraphics[width=\linewidth]{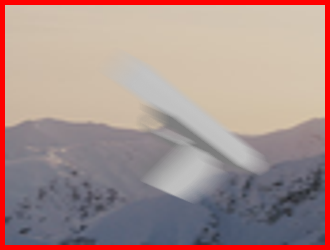}}{3}
      &\addimgTtext{\includegraphics[width=\linewidth]{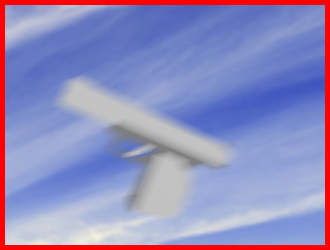}}{3}
      &\addimgTtext{\includegraphics[width=\linewidth]{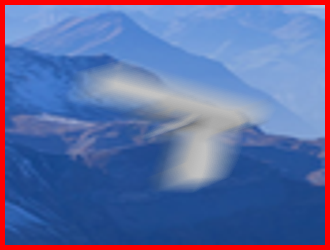}}{5}
      &\addimgTtext{\includegraphics[width=\linewidth]{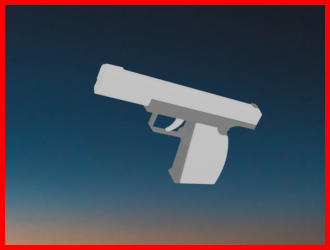}}{1}
      &\addimgTtext{\includegraphics[width=\linewidth]{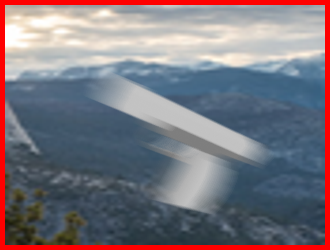}}{4}
		\\
    &&\addimgTtext{\includegraphics[width=\linewidth]{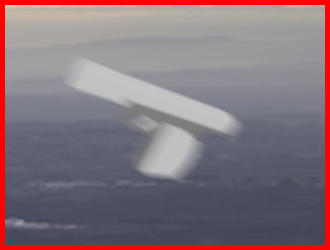}}{2}
      &\addimgTtext{\includegraphics[width=\linewidth]{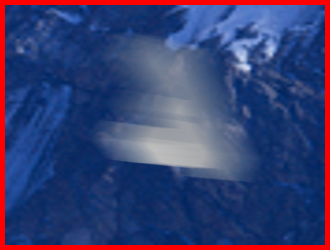}}{6}
      &\addimgTtext{\includegraphics[width=\linewidth]{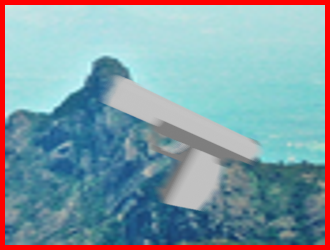}}{2}
      &\addimgTtext{\includegraphics[width=\linewidth]{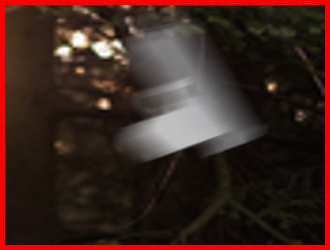}}{6}
      &\addimgTtext{\includegraphics[width=\linewidth]{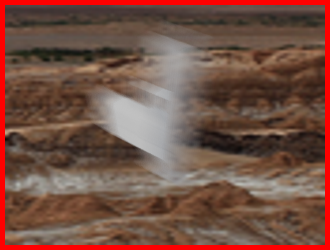}}{5}
      &\addimgTtext{\includegraphics[width=\linewidth]{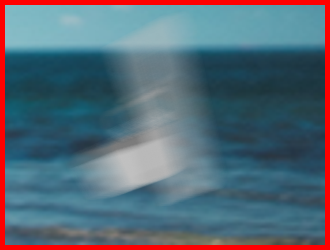}}{6}
      &\addimgTtext{\includegraphics[width=\linewidth]{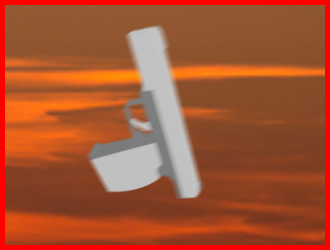}}{2}
      &\addimgTtext{\includegraphics[width=\linewidth]{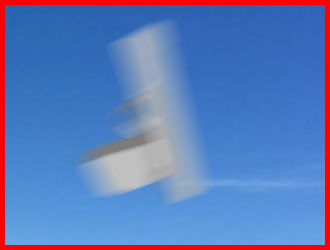}}{5}
      &\addimgTtext{\includegraphics[width=\linewidth]{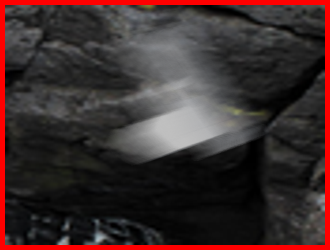}}{6}
      &\addimgTtext{\includegraphics[width=\linewidth]{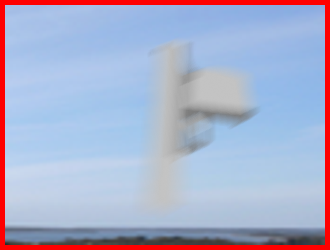}}{4} \\
  \addlinespace[5pt]
\raisebox{0.1em}{\multirow{2}{*}{
\addimgTtext{\includegraphics[width=\linewidth]{fig/suppmat/qualitativeBL6_suppmat_gun/query_image_7071.png}}{6}}}
&\raisebox{1.1em}{\multirow{2}{*}{\rotatebox[origin=c]{90}{Token~\cite{token}}}}
    &\addimgTtext{\includegraphics[width=\linewidth]{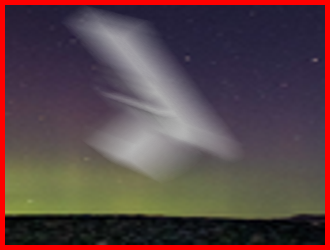}}{6}
      &\addimgTtext{\includegraphics[width=\linewidth]{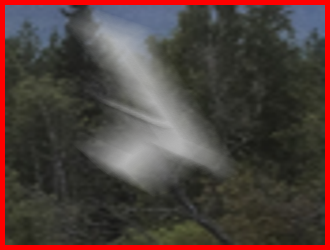}}{6}
      &\addimgTtext{\includegraphics[width=\linewidth]{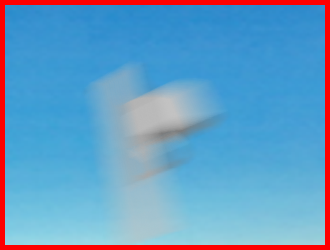}}{6}
      &\addimgTtext{\includegraphics[width=\linewidth]{fig/suppmat/qualitativeBL6_suppmat_gun/Token/rank03.png}}{6}
      &\addimgTtext{\includegraphics[width=\linewidth]{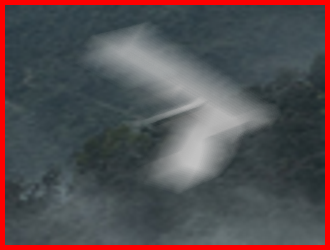}}{5}
      &\addimgTtext{\includegraphics[width=\linewidth]{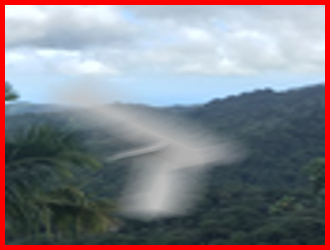}}{5}
      &\addimgTtext{\includegraphics[width=\linewidth]{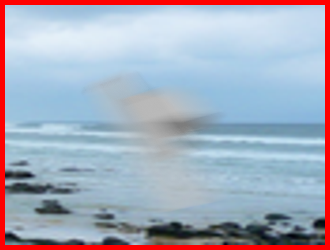}}{6}
      &\addimgTtext{\includegraphics[width=\linewidth]{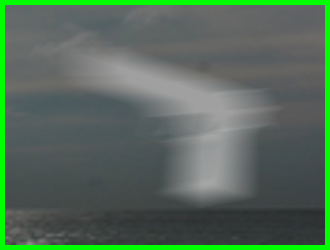}}{6}
      &\addimgTtext{\includegraphics[width=\linewidth]{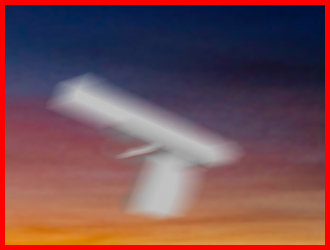}}{4}
      &\addimgTtext{\includegraphics[width=\linewidth]{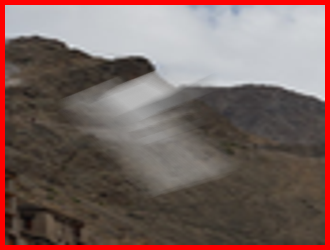}}{6}
		\\
    &&\addimgTtext{\includegraphics[width=\linewidth]{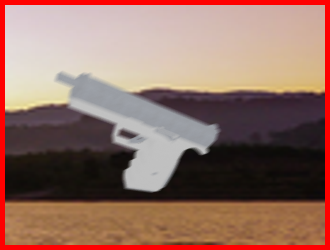}}{1}
      &\addimgTtext{\includegraphics[width=\linewidth]{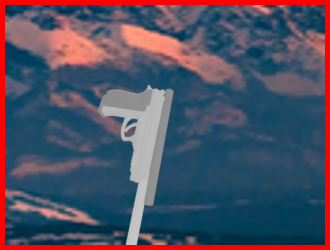}}{1}
      &\addimgTtext{\includegraphics[width=\linewidth]{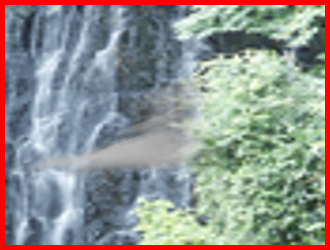}}{5}
      &\addimgTtext{\includegraphics[width=\linewidth]{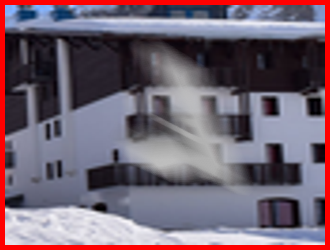}}{5}
      &\addimgTtext{\includegraphics[width=\linewidth]{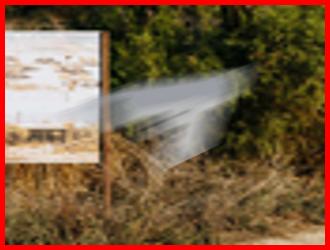}}{5}
      &\addimgTtext{\includegraphics[width=\linewidth]{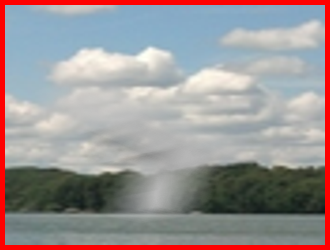}}{6}
      &\addimgTtext{\includegraphics[width=\linewidth]{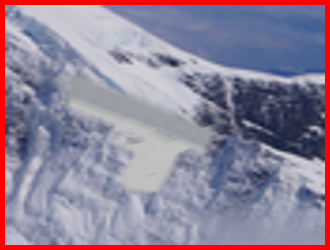}}{2}
      &\addimgTtext{\includegraphics[width=\linewidth]{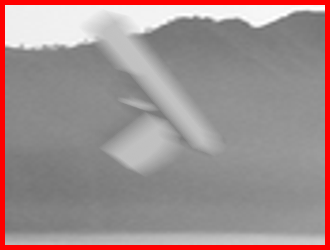}}{4}
      &\addimgTtext{\includegraphics[width=\linewidth]{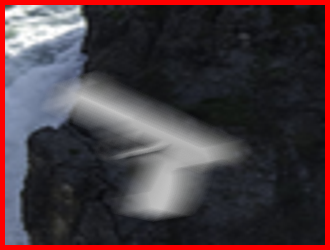}}{5}
      &\addimgTtext{\includegraphics[width=\linewidth]{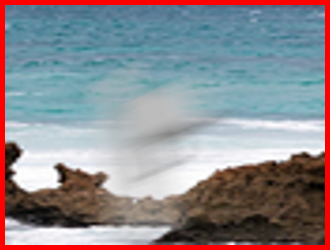}}{6} \\  \addlinespace[5pt]
\raisebox{0em}{\multirow{2}{*}{ 
\addimgTtext{\includegraphics[width=\linewidth]{fig/suppmat/qualitativeBL6_suppmat_gun/query_image_7071.png}}{6}}}
&\raisebox{0.1em}{\multirow{2}{*}{\rotatebox[origin=c]{90}{Ours}}}
    &\addimgTtext{\includegraphics[width=\linewidth]{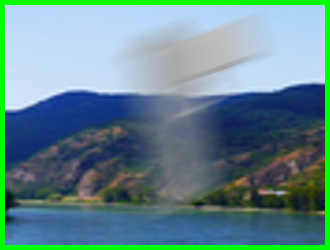}}{6}
      &\addimgTtext{\includegraphics[width=\linewidth]{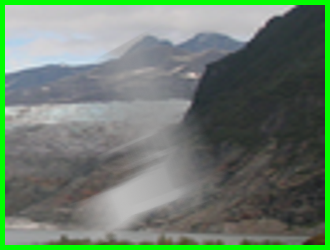}}{6}
      &\addimgTtext{\includegraphics[width=\linewidth]{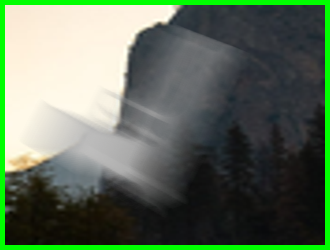}}{6}
      &\addimgTtext{\includegraphics[width=\linewidth]{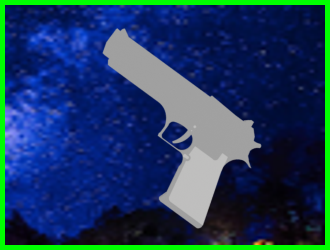}}{1}
      &\addimgTtext{\includegraphics[width=\linewidth]{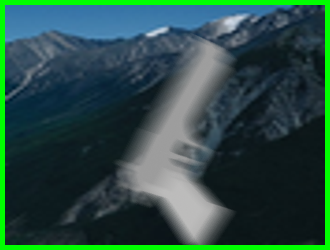}}{4}
      &\addimgTtext{\includegraphics[width=\linewidth]{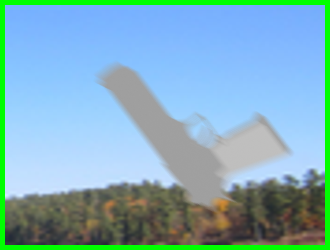}}{2}
      &\addimgTtext{\includegraphics[width=\linewidth]{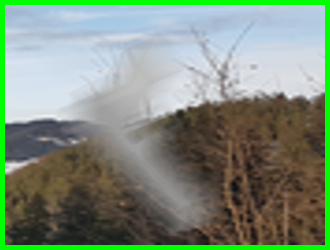}}{5}
      &\addimgTtext{\includegraphics[width=\linewidth]{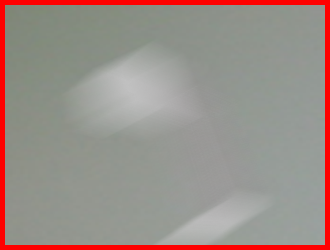}}{6}
      &\addimgTtextblack{\includegraphics[width=\linewidth]{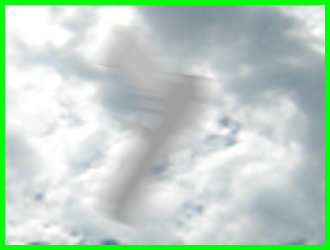}}{5}
      &\addimgTtext{\includegraphics[width=\linewidth]{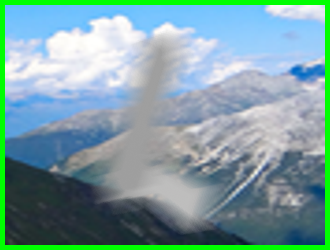}}{5}
		\\
    &&\addimgTtext{\includegraphics[width=\linewidth]{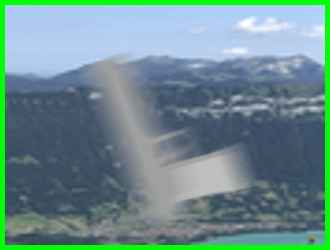}}{4}
      &\addimgTtext{\includegraphics[width=\linewidth]{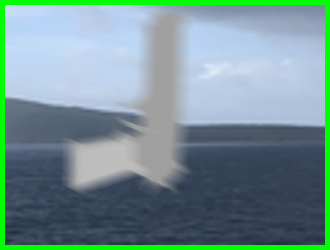}}{4}
      &\addimgTtext{\includegraphics[width=\linewidth]{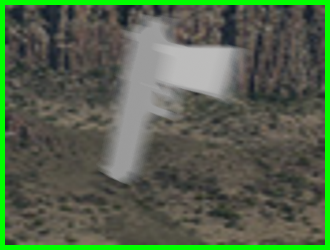}}{3}
      &\addimgTtext{\includegraphics[width=\linewidth]{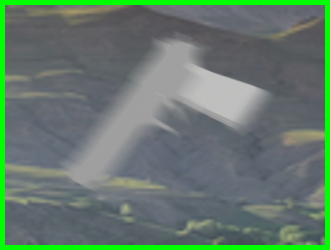}}{3}
      &\addimgTtext{\includegraphics[width=\linewidth]{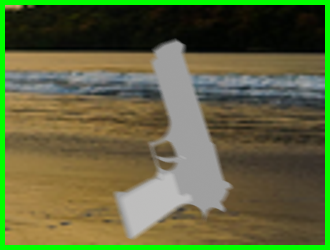}}{2}
      &\addimgTtext{\includegraphics[width=\linewidth]{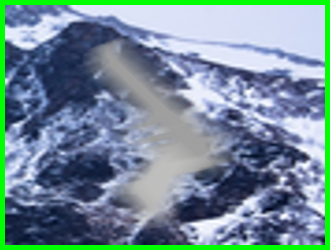}}{4}
      &\addimgTtext{\includegraphics[width=\linewidth]{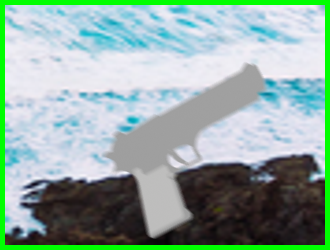}}{1}
      &\addimgTtext{\includegraphics[width=\linewidth]{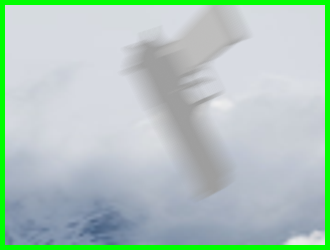}}{4}
      &\addimgTtext{\includegraphics[width=\linewidth]{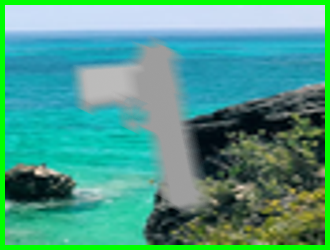}}{3}
      &\addimgTtext{\includegraphics[width=\linewidth]{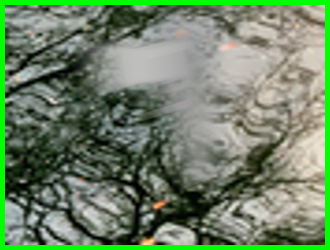}}{5} \\
	\end{tabular}
	\label{fig:qualitative_suppmat_syn2}
 
\end{subfigure}

 \begin{subfigure}{\linewidth}
    \setlength{\tabcolsep}{2pt}
    \setlength{\fboxrule}{1pt} 
    \setlength{\fboxsep}{0pt}  
	\begin{tabular}{
	 P{\figWidth}
	P{0.2cm}
	P{\figWidth}
	P{\figWidth}
    P{\figWidth}
    P{\figWidth}
    P{\figWidth}
    P{\figWidth}
	P{\figWidth}
    P{\figWidth}
    P{\figWidth}
    P{\figWidth}
    }
    \addlinespace[5pt]
	\multicolumn{2}{c}{Query}	&   \multicolumn{10}{c}{Top 20 retrieval results}
		\\  \addlinespace[5pt]
\raisebox{0em}{\multirow{2}{*}{ 
\addimgTtext{\includegraphics[width=\linewidth]{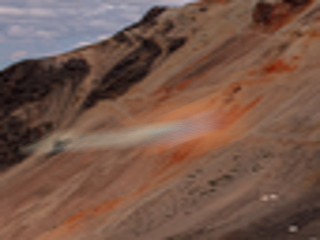}}{6}}}
&\raisebox{1.1em}{\multirow{2}{*}{\rotatebox[origin=c]{90}{DELG~\cite{delg}}}}
    &\addimgTtext{\includegraphics[width=\linewidth]{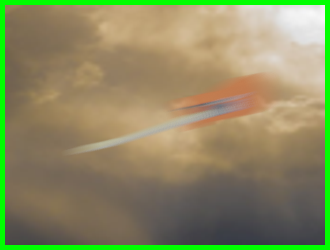}}{4}
      &\addimgTtext{\includegraphics[width=\linewidth]{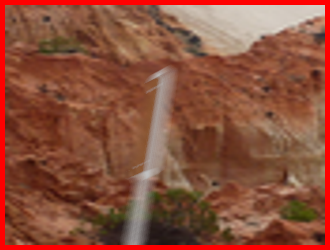}}{5}
      &\addimgTtext{\includegraphics[width=\linewidth]{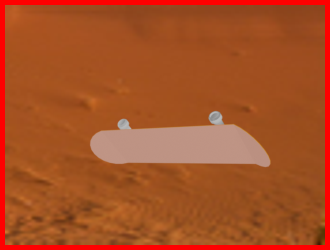}}{1}
      &\addimgTtext{\includegraphics[width=\linewidth]{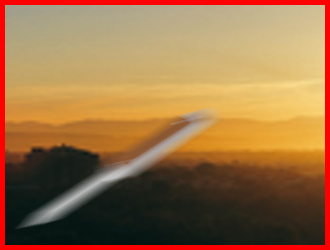}}{5}
      &\addimgTtext{\includegraphics[width=\linewidth]{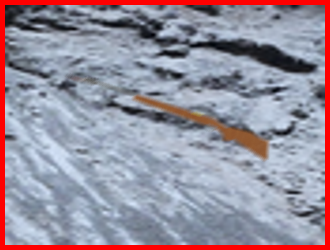}}{4}
      &\addimgTtext{\includegraphics[width=\linewidth]{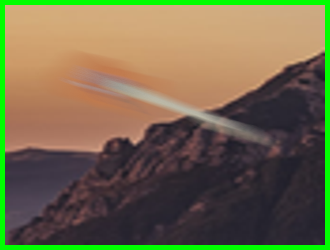}}{5}
      &\addimgTtext{\includegraphics[width=\linewidth]{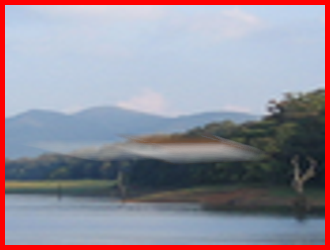}}{6}
      &\addimgTtext{\includegraphics[width=\linewidth]{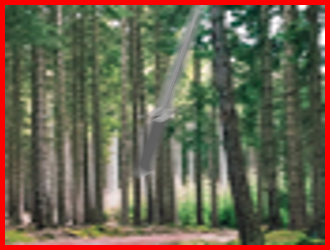}}{4}
      &\addimgTtext{\includegraphics[width=\linewidth]{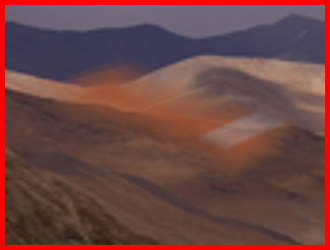}}{6}
      &\addimgTtext{\includegraphics[width=\linewidth]{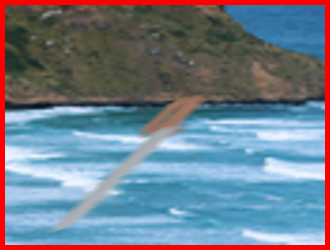}}{4}
		\\
    &&\addimgTtext{\includegraphics[width=\linewidth]{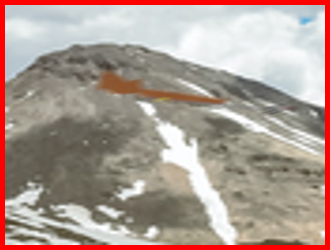}}{4}
      &\addimgTtext{\includegraphics[width=\linewidth]{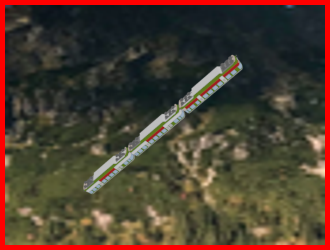}}{1}
      &\addimgTtext{\includegraphics[width=\linewidth]{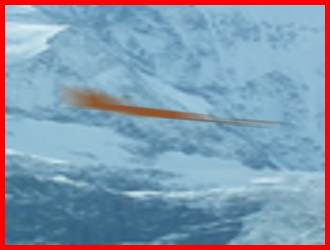}}{5}
      &\addimgTtext{\includegraphics[width=\linewidth]{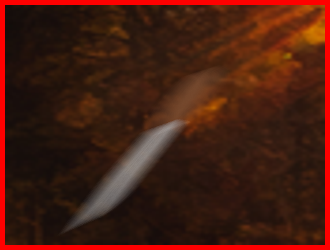}}{6}
      &\addimgTtext{\includegraphics[width=\linewidth]{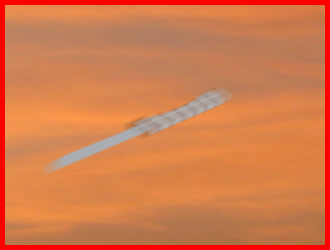}}{2}
      &\addimgTtext{\includegraphics[width=\linewidth]{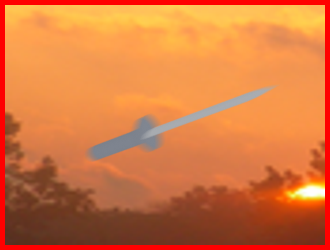}}{2}
      &\addimgTtext{\includegraphics[width=\linewidth]{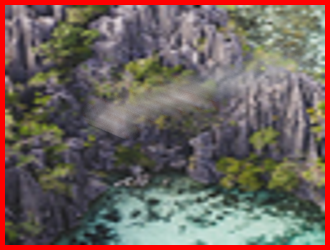}}{6}
      &\addimgTtext{\includegraphics[width=\linewidth]{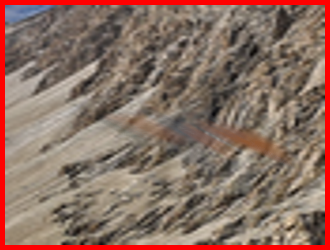}}{6}
      &\addimgTtext{\includegraphics[width=\linewidth]{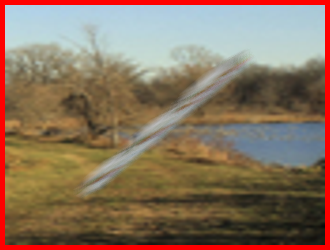}}{4}
      &\addimgTtext{\includegraphics[width=\linewidth]{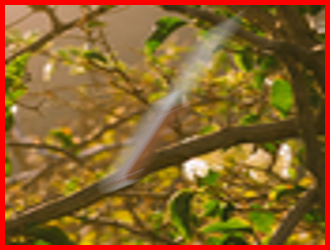}}{5} \\  \addlinespace[5pt]
\raisebox{0em}{\multirow{2}{*}{
\addimgTtext{\includegraphics[width=\linewidth]{fig/suppmat/qualitativeBL6_suppmat_guitar/query_image_12529.png}}{6}}}
&\raisebox{1.10em}{\multirow{2}{*}{\rotatebox[origin=c]{90}{DOLG~\cite{dolg}}}}
    &\addimgTtext{\includegraphics[width=\linewidth]{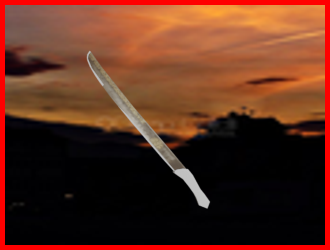}}{1}
      &\addimgTtext{\includegraphics[width=\linewidth]{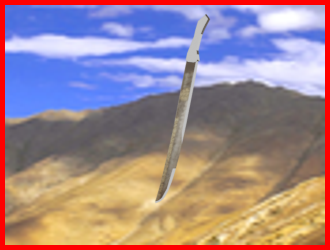}}{1}
      &\addimgTtext{\includegraphics[width=\linewidth]{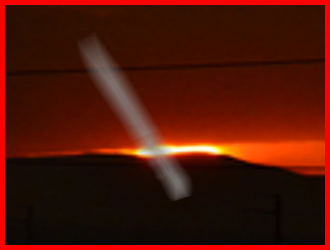}}{6}
      &\addimgTtext{\includegraphics[width=\linewidth]{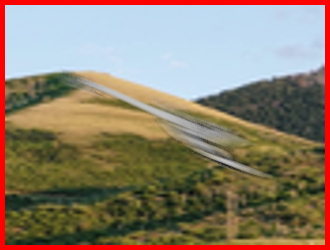}}{5}
      &\addimgTtext{\includegraphics[width=\linewidth]{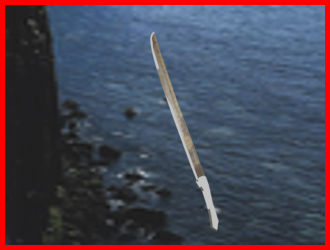}}{1}
      &\addimgTtext{\includegraphics[width=\linewidth]{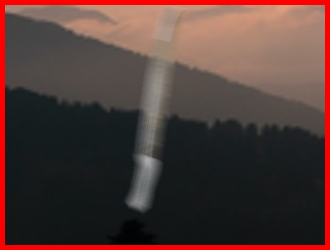}}{6}
      &\addimgTtext{\includegraphics[width=\linewidth]{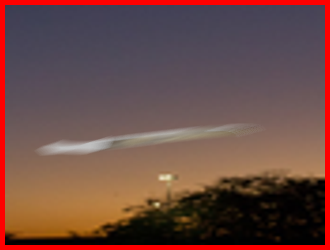}}{4}
      &\addimgTtextblack{\includegraphics[width=\linewidth]{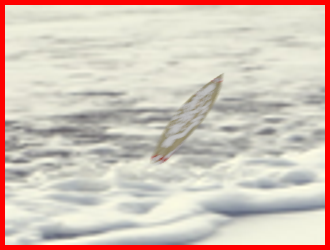}}{1}
      &\addimgTtext{\includegraphics[width=\linewidth]{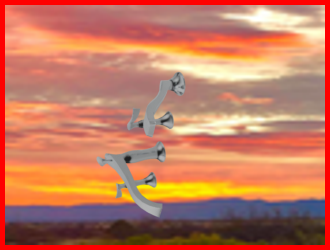}}{1}
      &\addimgTtext{\includegraphics[width=\linewidth]{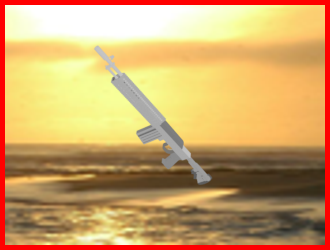}}{1}
		\\
    &&\addimgTtext{\includegraphics[width=\linewidth]{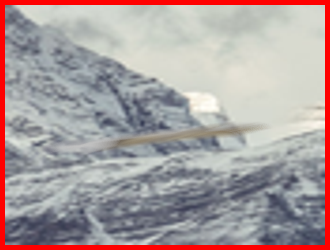}}{4}
      &\addimgTtext{\includegraphics[width=\linewidth]{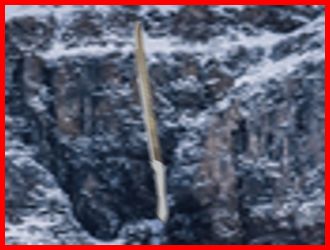}}{4}
      &\addimgTtext{\includegraphics[width=\linewidth]{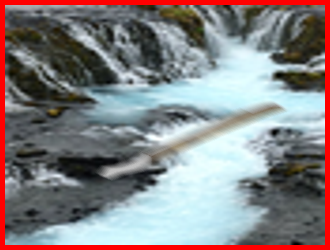}}{4}
      &\addimgTtext{\includegraphics[width=\linewidth]{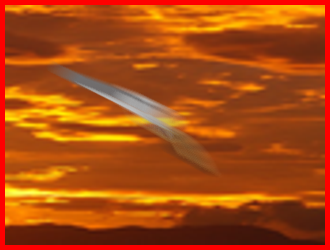}}{5}
      &\addimgTtext{\includegraphics[width=\linewidth]{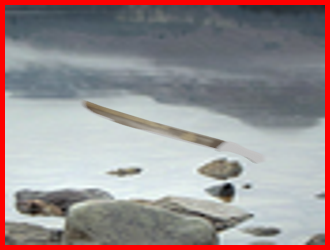}}{1}
      &\addimgTtext{\includegraphics[width=\linewidth]{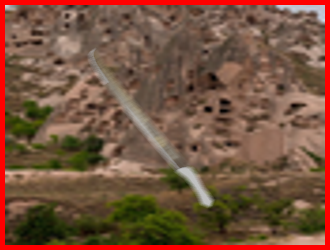}}{3}
      &\addimgTtext{\includegraphics[width=\linewidth]{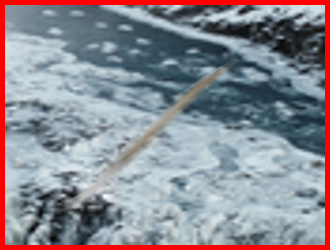}}{5}
      &\addimgTtext{\includegraphics[width=\linewidth]{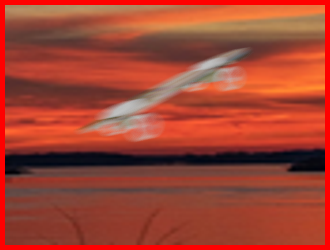}}{4}
      &\addimgTtext{\includegraphics[width=\linewidth]{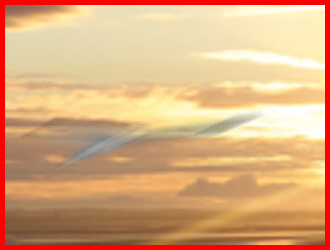}}{6}
      &\addimgTtext{\includegraphics[width=\linewidth]{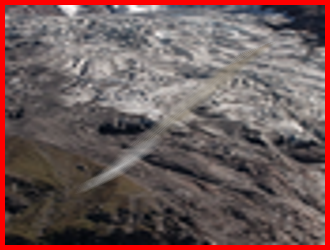}}{6} \\  \addlinespace[5pt]
\raisebox{0.1em}{\multirow{2}{*}{
\addimgTtext{\includegraphics[width=\linewidth]{fig/suppmat/qualitativeBL6_suppmat_guitar/query_image_12529.png}}{6}}}
&\raisebox{1.1em}{\multirow{2}{*}{\rotatebox[origin=c]{90}{Token~\cite{token}}}}
    &\addimgTtext{\includegraphics[width=\linewidth]{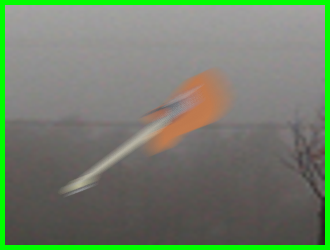}}{4}
      &\addimgTtext{\includegraphics[width=\linewidth]{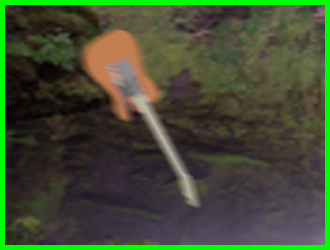}}{3}
      &\addimgTtext{\includegraphics[width=\linewidth]{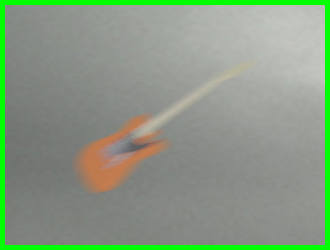}}{3}
      &\addimgTtext{\includegraphics[width=\linewidth]{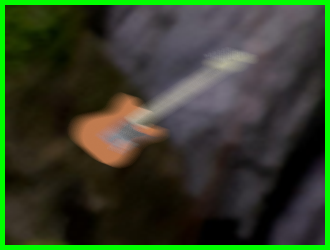}}{4}
      &\addimgTtext{\includegraphics[width=\linewidth]{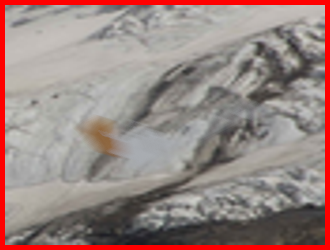}}{5}
      &\addimgTtext{\includegraphics[width=\linewidth]{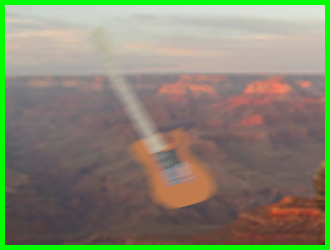}}{4}
      &\addimgTtext{\includegraphics[width=\linewidth]{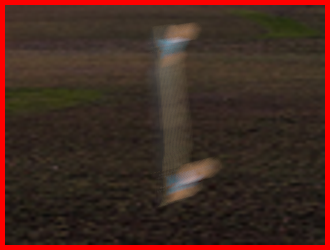}}{6}
      &\addimgTtext{\includegraphics[width=\linewidth]{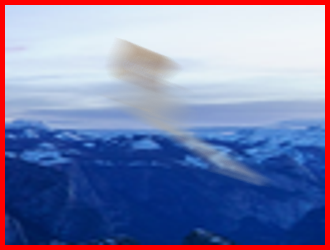}}{6}
      &\addimgTtext{\includegraphics[width=\linewidth]{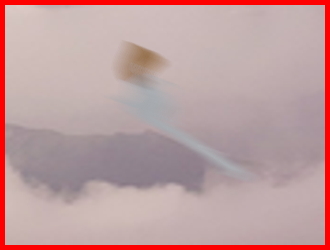}}{5}
      &\addimgTtext{\includegraphics[width=\linewidth]{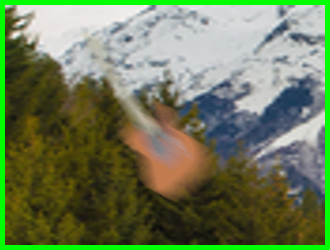}}{4}
		\\
    &&\addimgTtext{\includegraphics[width=\linewidth]{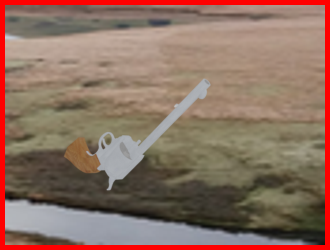}}{1}
      &\addimgTtext{\includegraphics[width=\linewidth]{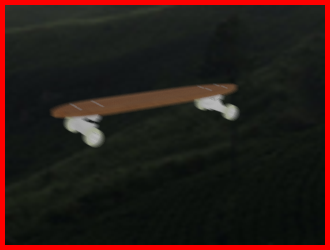}}{2}
      &\addimgTtext{\includegraphics[width=\linewidth]{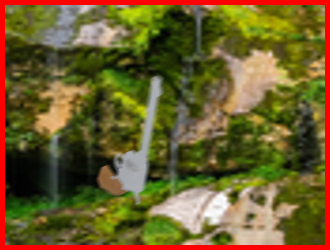}}{3}
      &\addimgTtext{\includegraphics[width=\linewidth]{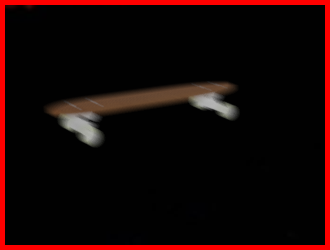}}{3}
      &\addimgTtext{\includegraphics[width=\linewidth]{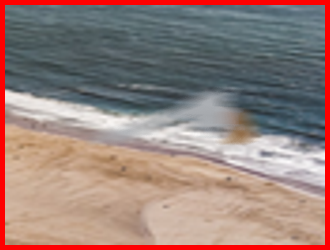}}{5}
      &\addimgTtext{\includegraphics[width=\linewidth]{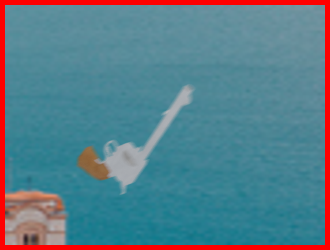}}{2}
      &\addimgTtext{\includegraphics[width=\linewidth]{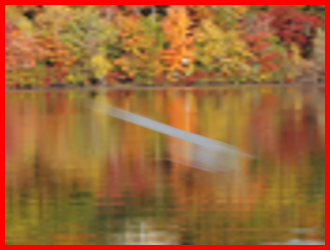}}{5}
      &\addimgTtext{\includegraphics[width=\linewidth]{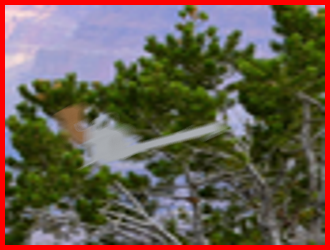}}{4}
      &\addimgTtext{\includegraphics[width=\linewidth]{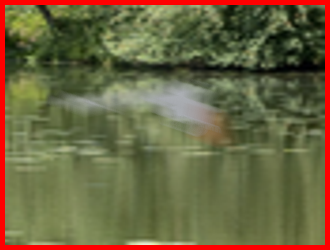}}{6}
      &\addimgTtext{\includegraphics[width=\linewidth]{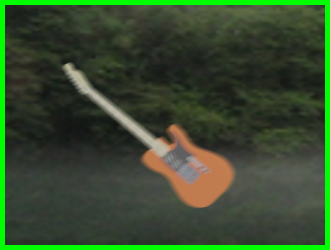}}{2} \\
        \addlinespace[5pt]
\raisebox{0em}{\multirow{2}{*}{ 
\addimgTtext{\includegraphics[width=\linewidth]{fig/suppmat/qualitativeBL6_suppmat_guitar/query_image_12529.png}}{6}}}
&\raisebox{0.1em}{\multirow{2}{*}{\rotatebox[origin=c]{90}{Ours}}}
    &\addimgTtext{\includegraphics[width=\linewidth]{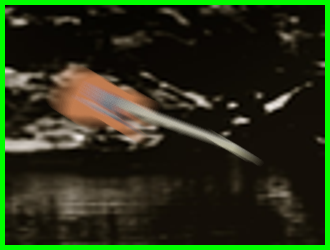}}{3}
      &\addimgTtext{\includegraphics[width=\linewidth]{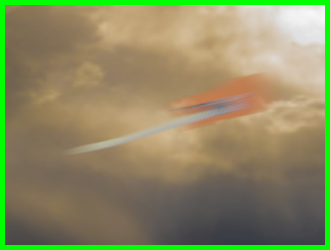}}{4}
      &\addimgTtext{\includegraphics[width=\linewidth]{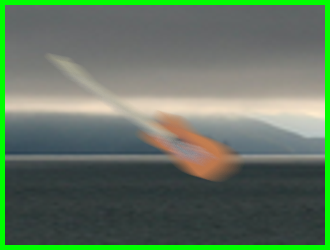}}{4}
      &\addimgTtext{\includegraphics[width=\linewidth]{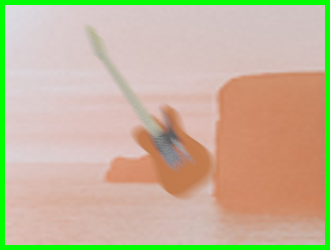}}{2}
      &\addimgTtext{\includegraphics[width=\linewidth]{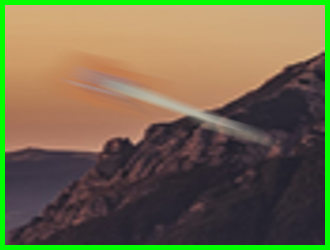}}{5}
      &\addimgTtext{\includegraphics[width=\linewidth]{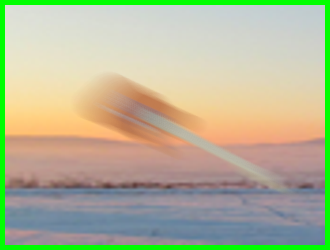}}{5}
      &\addimgTtext{\includegraphics[width=\linewidth]{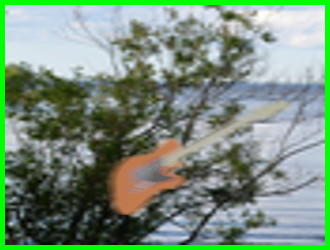}}{3}
      &\addimgTtext{\includegraphics[width=\linewidth]{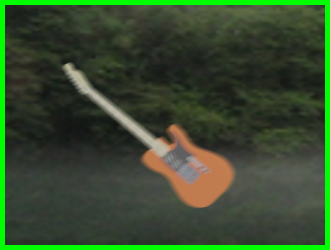}}{2}
      &\addimgTtext{\includegraphics[width=\linewidth]{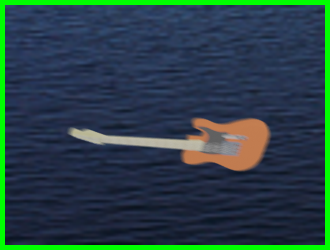}}{1}
      &\addimgTtext{\includegraphics[width=\linewidth]{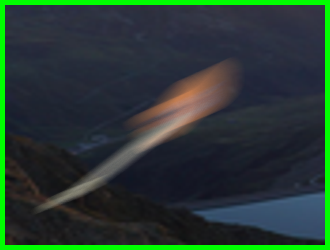}}{6}
		\\
    &&\addimgTtext{\includegraphics[width=\linewidth]{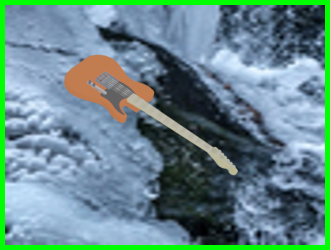}}{1}
      &\addimgTtext{\includegraphics[width=\linewidth]{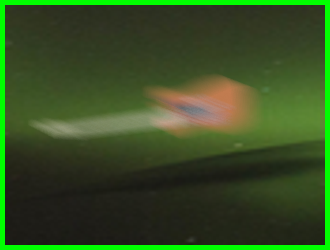}}{5}
      &\addimgTtext{\includegraphics[width=\linewidth]{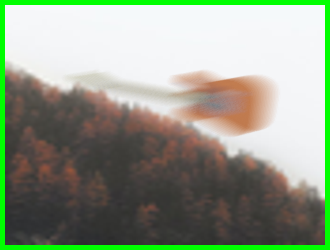}}{4}
      &\addimgTtext{\includegraphics[width=\linewidth]{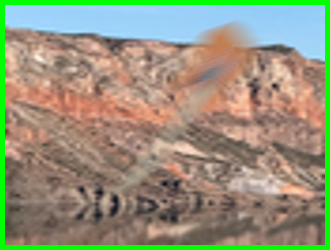}}{5}
      &\addimgTtextblack{\includegraphics[width=\linewidth]{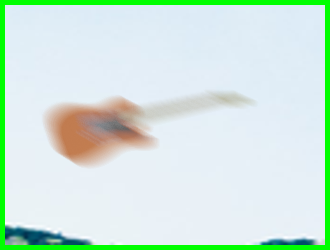}}{4}
      &\addimgTtext{\includegraphics[width=\linewidth]{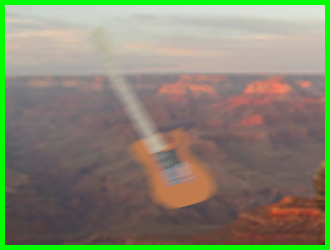}}{4}
      &\addimgTtext{\includegraphics[width=\linewidth]{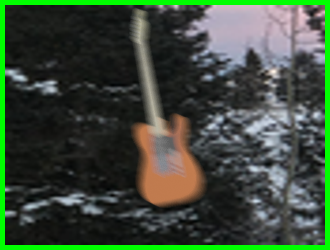}}{3}
      &\addimgTtext{\includegraphics[width=\linewidth]{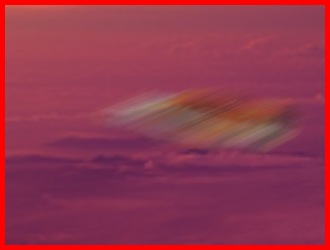}}{5}
      &\addimgTtext{\includegraphics[width=\linewidth]{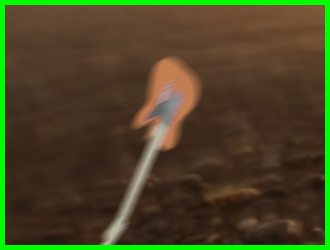}}{3}
      &\addimgTtext{\includegraphics[width=\linewidth]{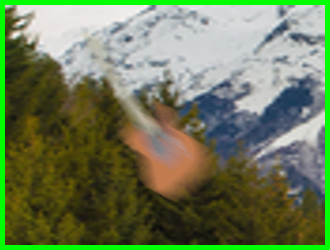}}{4} \\

	\end{tabular}
	\label{fig:qualitative_suppmat_syn2}
\end{subfigure}
\caption{\textbf{Comparison of retrieval results} between our approach and state-of-the-art retrieval methods on our synthetic dataset (with 1M distractors). The retrieved images are sorted from left to right, top to bottom, with the ranking from the 1\textsuperscript{st} to the 20\textsuperscript{th}. The blur level of each image is shown in the bottom left corner.
Correct results are indicated with green boxes, while incorrect ones are marked with red boxes.}
\label{fig:qualitative_suppmat_syn}

\end{figure*}

 \global\long\def\figWidth{0.077\linewidth}
\begin{figure*}
    \centering

 \begin{subfigure}{\linewidth}
    \setlength{\tabcolsep}{2pt}
    \setlength{\fboxrule}{1pt} 
    \setlength{\fboxsep}{0pt} 
    \begin{tabular}{
     P{\figWidth}
    P{0.2cm}
    P{\figWidth}
    P{\figWidth}
    P{\figWidth}
    P{\figWidth}
    P{\figWidth}
    P{\figWidth}
    P{\figWidth}
    P{\figWidth}
    P{\figWidth}
    P{\figWidth}
    }
        \addlinespace[5pt]
    \multicolumn{2}{c}{Query}	&   \multicolumn{10}{c}{Top 20 retrieval results}
        \\
    \addlinespace[5pt]

\raisebox{0em}{\multirow{2}{*}{ 
\addimgTtext{\includegraphics[width=\linewidth]{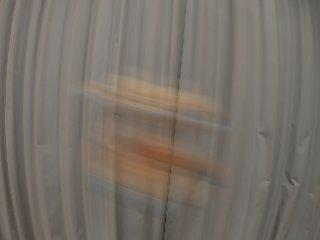}}{6}}}
&\raisebox{1.1em}{\multirow{2}{*}{\rotatebox[origin=c]{90}{DELG~\cite{delg}}}}
    &\addimgTtext{\includegraphics[width=\linewidth]{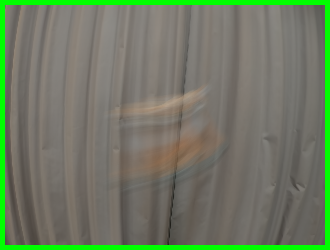}}{6}
      &\addimgTtext{\includegraphics[width=\linewidth]{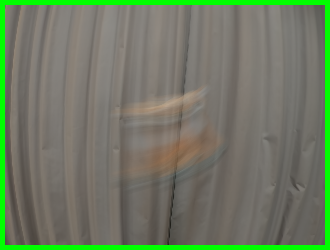}}{6}
      &\addimgTtext{\includegraphics[width=\linewidth]{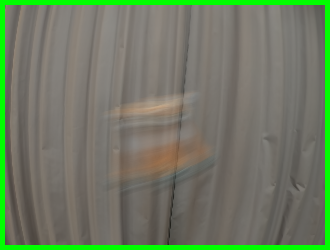}}{5}
      &\addimgTtext{\includegraphics[width=\linewidth]{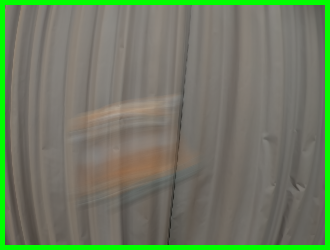}}{6}
      &\addimgTtext{\includegraphics[width=\linewidth]{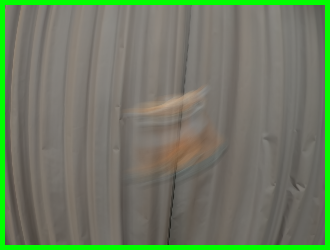}}{5}
      &\addimgTtext{\includegraphics[width=\linewidth]{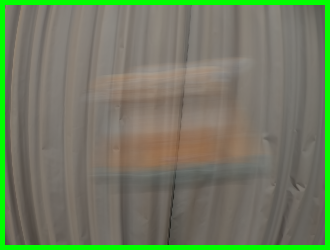}}{6}
      &\addimgTtext{\includegraphics[width=\linewidth]{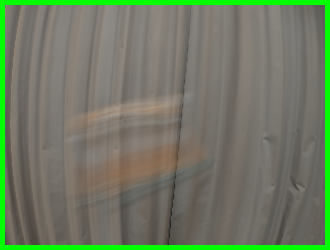}}{6}
      &\addimgTtext{\includegraphics[width=\linewidth]{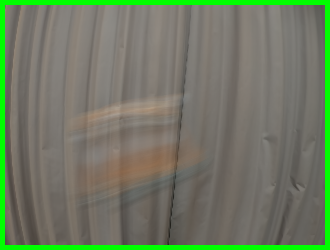}}{6}
      &\addimgTtext{\includegraphics[width=\linewidth]{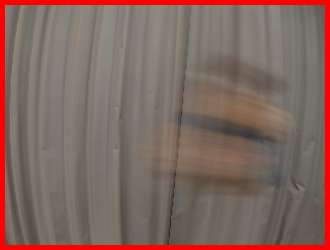}}{5}
      &\addimgTtext{\includegraphics[width=\linewidth]{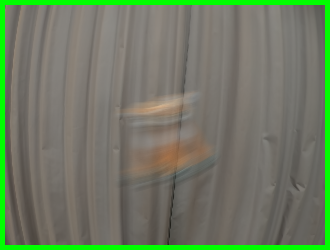}}{5}
        \\
    &&\addimgTtext{\includegraphics[width=\linewidth]{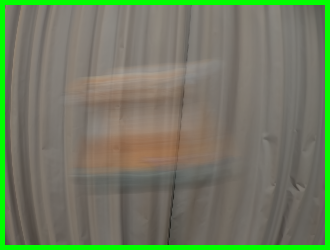}}{6}
      &\addimgTtext{\includegraphics[width=\linewidth]{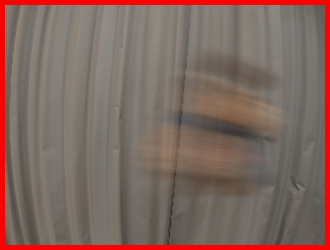}}{5}
      &\addimgTtext{\includegraphics[width=\linewidth]{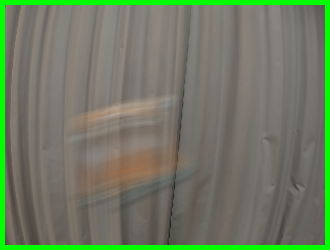}}{6}
      &\addimgTtext{\includegraphics[width=\linewidth]{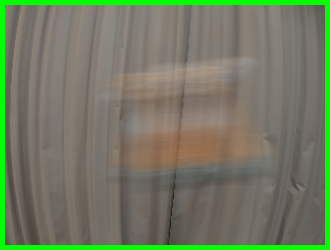}}{6}
      &\addimgTtext{\includegraphics[width=\linewidth]{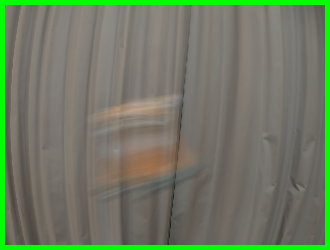}}{5}
      &\addimgTtext{\includegraphics[width=\linewidth]{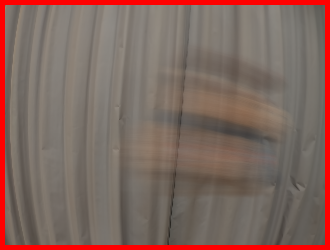}}{6}
      &\addimgTtext{\includegraphics[width=\linewidth]{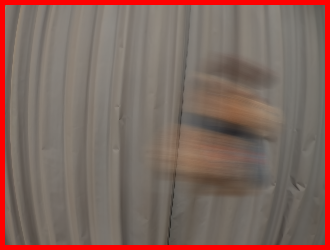}}{4}
      &\addimgTtext{\includegraphics[width=\linewidth]{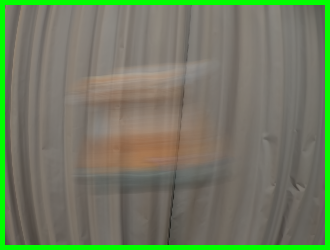}}{5}
      &\addimgTtext{\includegraphics[width=\linewidth]{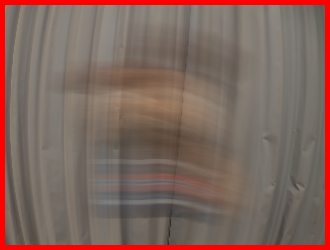}}{6}
      &\addimgTtext{\includegraphics[width=\linewidth]{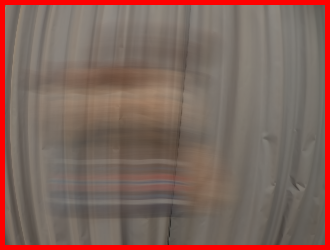}}{6} \\ 
      \addlinespace[5pt]
\raisebox{0em}{\multirow{2}{*}{ 
\addimgTtext{\includegraphics[width=\linewidth]{fig/suppmat/qualitativeBL6_suppmat_Deer/query_image_1200.png}}{6}}}
&\raisebox{1.10em}{\multirow{2}{*}{\rotatebox[origin=c]{90}{DOLG~\cite{dolg}}}}
    &\addimgTtext{\includegraphics[width=\linewidth]{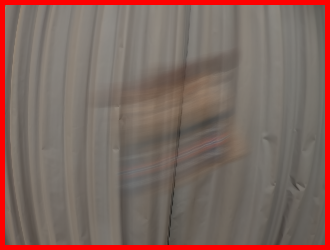}}{6}
      &\addimgTtext{\includegraphics[width=\linewidth]{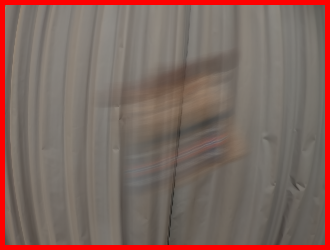}}{6}
      &\addimgTtext{\includegraphics[width=\linewidth]{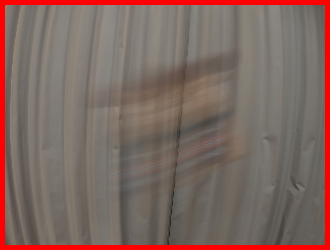}}{6}
      &\addimgTtext{\includegraphics[width=\linewidth]{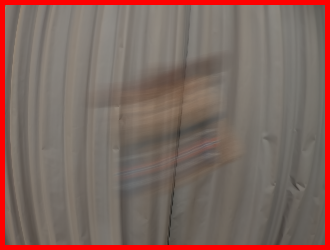}}{6}
      &\addimgTtext{\includegraphics[width=\linewidth]{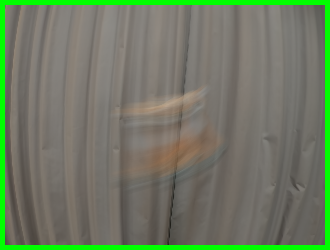}}{6}
      &\addimgTtext{\includegraphics[width=\linewidth]{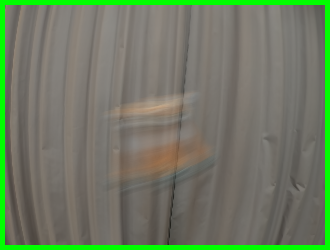}}{5}
      &\addimgTtext{\includegraphics[width=\linewidth]{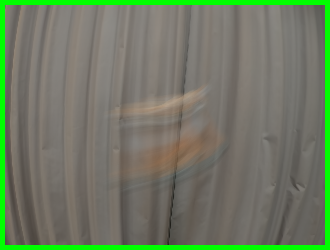}}{6}
      &\addimgTtext{\includegraphics[width=\linewidth]{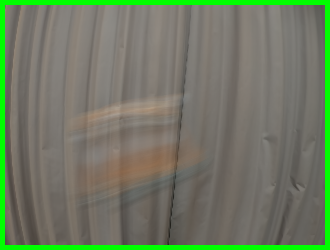}}{6}
      &\addimgTtext{\includegraphics[width=\linewidth]{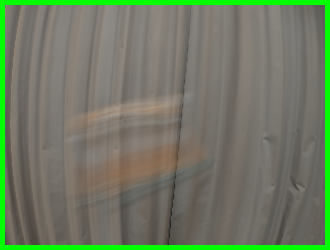}}{6}
      &\addimgTtext{\includegraphics[width=\linewidth]{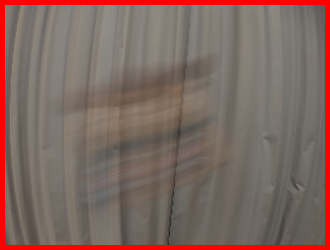}}{6}
        \\
    &&\addimgTtext{\includegraphics[width=\linewidth]{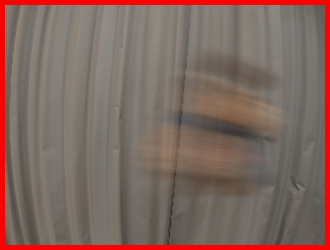}}{5}
      &\addimgTtext{\includegraphics[width=\linewidth]{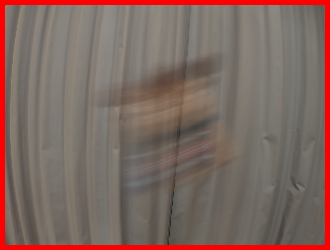}}{5}
      &\addimgTtext{\includegraphics[width=\linewidth]{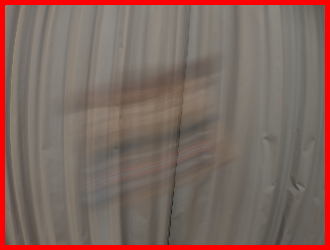}}{6}
      &\addimgTtext{\includegraphics[width=\linewidth]{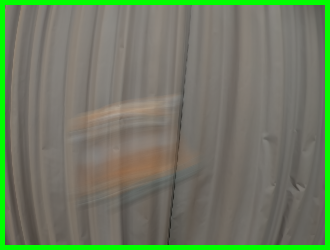}}{6}
      &\addimgTtext{\includegraphics[width=\linewidth]{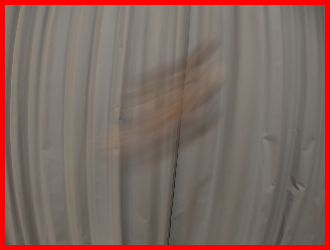}}{6}
      &\addimgTtext{\includegraphics[width=\linewidth]{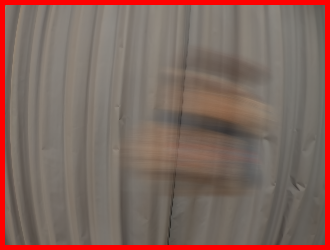}}{5}
      &\addimgTtext{\includegraphics[width=\linewidth]{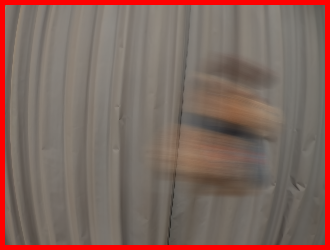}}{4}
      &\addimgTtext{\includegraphics[width=\linewidth]{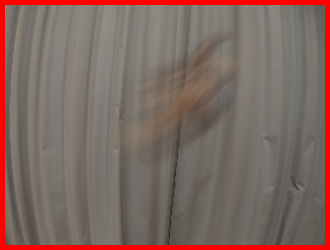}}{6}
      &\addimgTtext{\includegraphics[width=\linewidth]{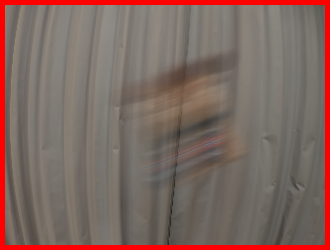}}{5}
      &\addimgTtext{\includegraphics[width=\linewidth]{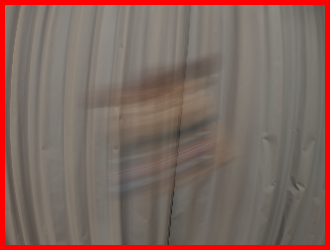}}{6} \\
      \addlinespace[5pt]
\raisebox{0.1em}{\multirow{2}{*}{
\addimgTtext{\includegraphics[width=\linewidth]{fig/suppmat/qualitativeBL6_suppmat_Deer/query_image_1200.png}}{6}}}
&\raisebox{1.1em}{\multirow{2}{*}{\rotatebox[origin=c]{90}{Token~\cite{token}}}}
    &\addimgTtext{\includegraphics[width=\linewidth]{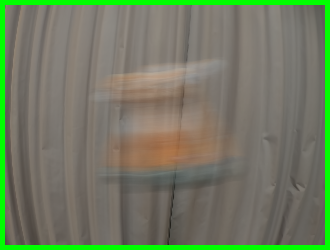}}{5}
      &\addimgTtext{\includegraphics[width=\linewidth]{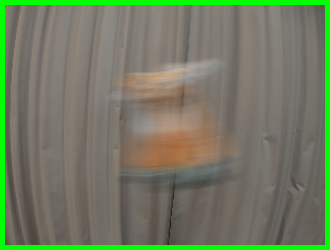}}{4}
      &\addimgTtext{\includegraphics[width=\linewidth]{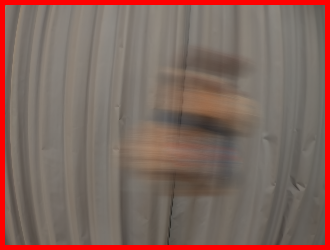}}{4}
      &\addimgTtext{\includegraphics[width=\linewidth]{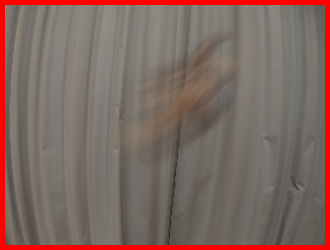}}{6}
      &\addimgTtext{\includegraphics[width=\linewidth]{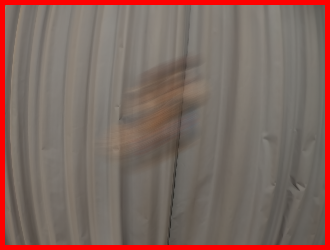}}{4}
      &\addimgTtext{\includegraphics[width=\linewidth]{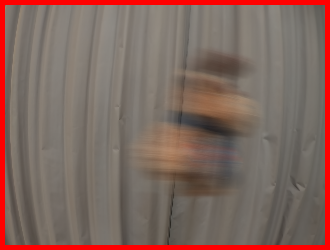}}{4}
      &\addimgTtext{\includegraphics[width=\linewidth]{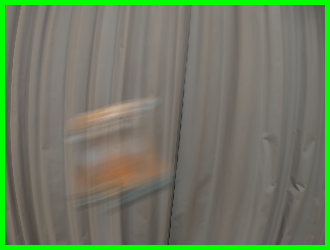}}{4}
      &\addimgTtext{\includegraphics[width=\linewidth]{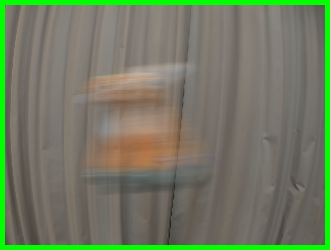}}{5}
      &\addimgTtext{\includegraphics[width=\linewidth]{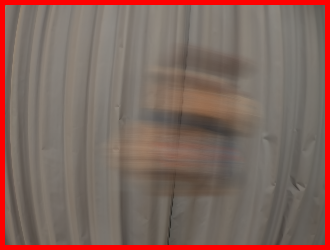}}{4}
      &\addimgTtext{\includegraphics[width=\linewidth]{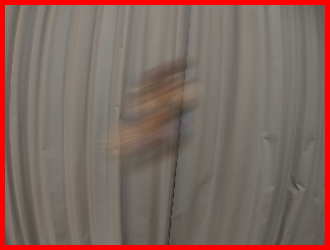}}{4}
        \\
    &&\addimgTtext{\includegraphics[width=\linewidth]{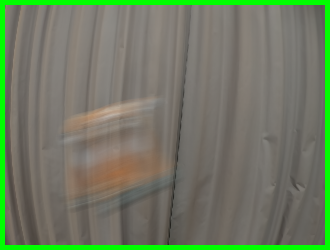}}{5}
      &\addimgTtext{\includegraphics[width=\linewidth]{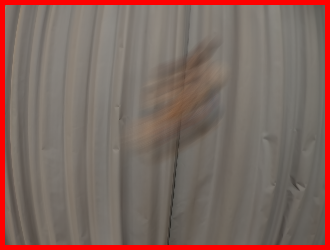}}{5}
      &\addimgTtext{\includegraphics[width=\linewidth]{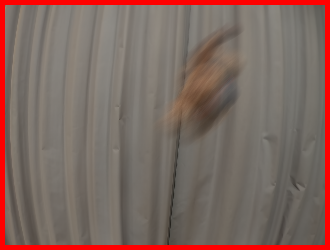}}{4}
      &\addimgTtext{\includegraphics[width=\linewidth]{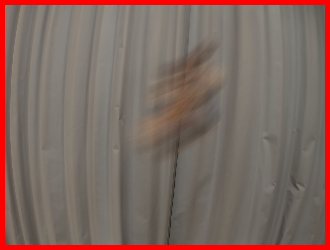}}{5}
      &\addimgTtext{\includegraphics[width=\linewidth]{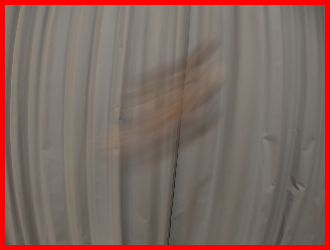}}{6}
      &\addimgTtext{\includegraphics[width=\linewidth]{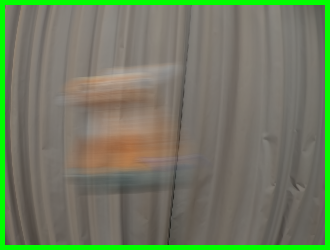}}{5}
      &\addimgTtext{\includegraphics[width=\linewidth]{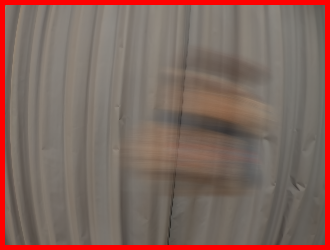}}{5}
      &\addimgTtext{\includegraphics[width=\linewidth]{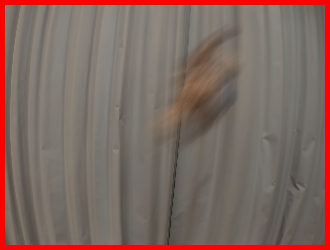}}{4}
      &\addimgTtext{\includegraphics[width=\linewidth]{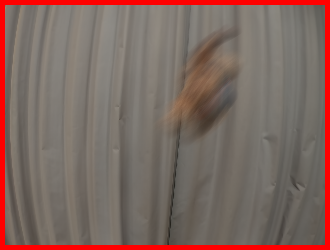}}{4}
      &\addimgTtext{\includegraphics[width=\linewidth]{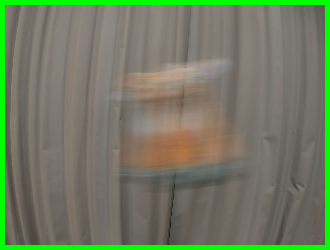}}{5} \\
      \addlinespace[5pt]
\raisebox{0em}{\multirow{2}{*}{ 
\addimgTtext{\includegraphics[width=\linewidth]{fig/suppmat/qualitativeBL6_suppmat_Deer/query_image_1200.png}}{6}}}
&\raisebox{0.1em}{\multirow{2}{*}{\rotatebox[origin=c]{90}{Ours}}}
    &\addimgTtext{\includegraphics[width=\linewidth]{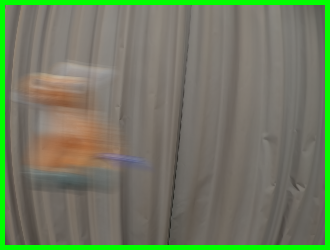}}{4}
      &\addimgTtext{\includegraphics[width=\linewidth]{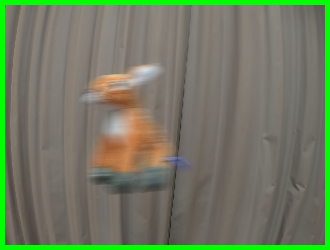}}{2}
      &\addimgTtext{\includegraphics[width=\linewidth]{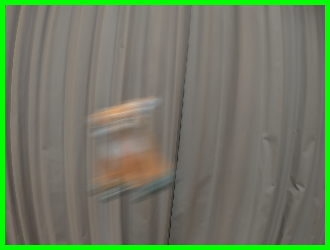}}{3}
      &\addimgTtext{\includegraphics[width=\linewidth]{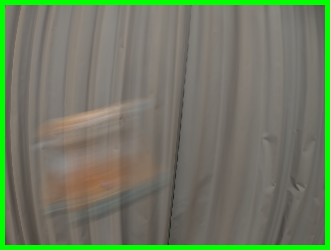}}{5}
      &\addimgTtext{\includegraphics[width=\linewidth]{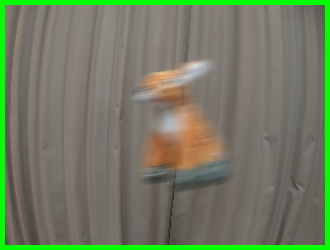}}{3}
      &\addimgTtext{\includegraphics[width=\linewidth]{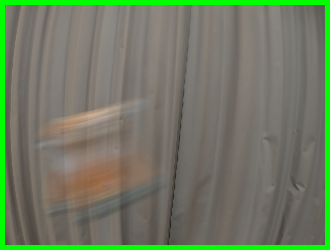}}{5}
      &\addimgTtext{\includegraphics[width=\linewidth]{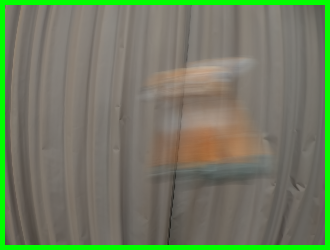}}{4}
      &\addimgTtext{\includegraphics[width=\linewidth]{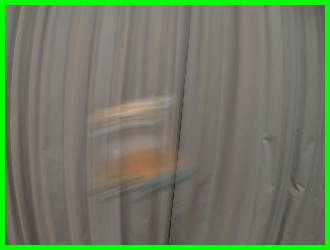}}{4}
      &\addimgTtext{\includegraphics[width=\linewidth]{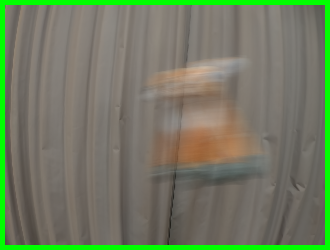}}{4}
      &\addimgTtext{\includegraphics[width=\linewidth]{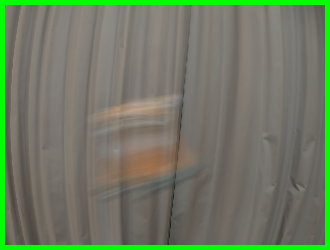}}{5}
        \\
    &&\addimgTtext{\includegraphics[width=\linewidth]{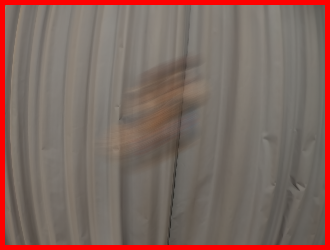}}{4}
      &\addimgTtext{\includegraphics[width=\linewidth]{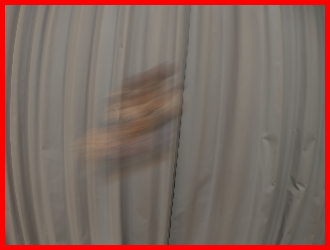}}{5}
      &\addimgTtext{\includegraphics[width=\linewidth]{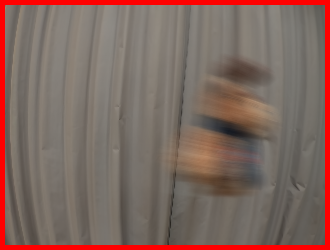}}{4}
      &\addimgTtext{\includegraphics[width=\linewidth]{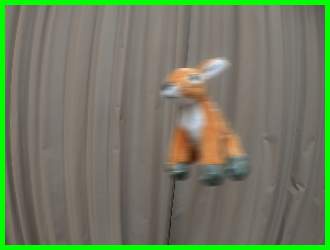}}{2}
      &\addimgTtext{\includegraphics[width=\linewidth]{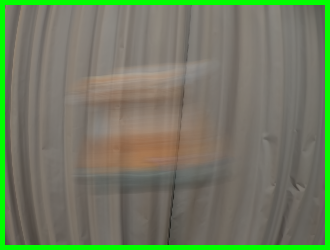}}{5}
      &\addimgTtext{\includegraphics[width=\linewidth]{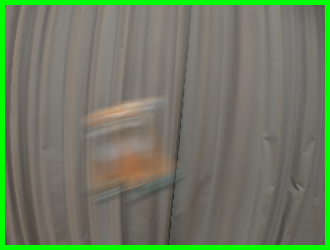}}{4}
      &\addimgTtext{\includegraphics[width=\linewidth]{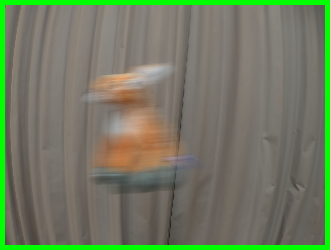}}{3}
      &\addimgTtext{\includegraphics[width=\linewidth]{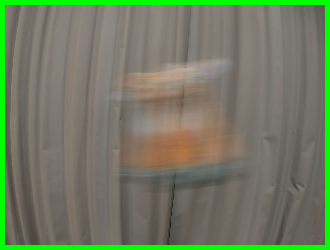}}{5}
      &\addimgTtext{\includegraphics[width=\linewidth]{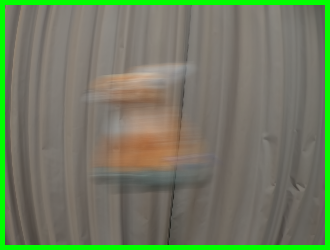}}{4}
      &\addimgTtext{\includegraphics[width=\linewidth]{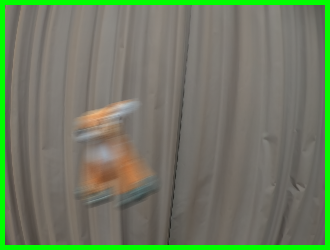}}{3} \\
    \end{tabular}
    \label{fig:real_qualitative_2}
\end{subfigure}

 \begin{subfigure}{\linewidth}
    \setlength{\tabcolsep}{2pt}
    \setlength{\fboxrule}{1pt} 
    \setlength{\fboxsep}{0pt}  
    \begin{tabular}{
     P{\figWidth}
    P{0.2cm}
    P{\figWidth}
    P{\figWidth}
    P{\figWidth}
    P{\figWidth}
    P{\figWidth}
    P{\figWidth}
    P{\figWidth}
    P{\figWidth}
    P{\figWidth}
    P{\figWidth}
    }
    \multicolumn{2}{c}{Query}	&   \multicolumn{10}{c}{Top 20 retrieval results}
        \\
      \addlinespace[5pt]
\raisebox{0em}{\multirow{2}{*}{
\addimgTtext{\includegraphics[width=\linewidth]{fig/suppmat/qualitativeBL5_suppmat_football/query_image_1425.png}}{5}}}
&\raisebox{1.1em}{\multirow{2}{*}{\rotatebox[origin=c]{90}{DELG~\cite{delg}}}}
      &\addimgTtext{\includegraphics[width=\linewidth]{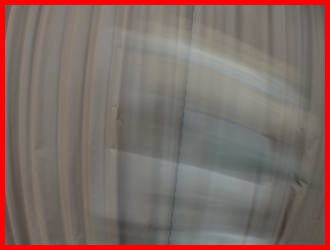}}{6}
      &\addimgTtext{\includegraphics[width=\linewidth]{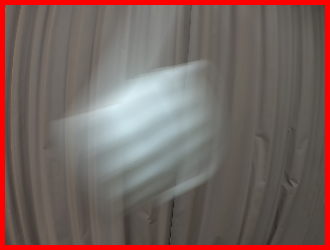}}{4}
      &\addimgTtext{\includegraphics[width=\linewidth]{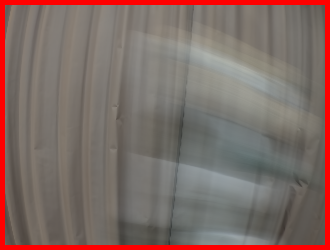}}{5}
      &\addimgTtext{\includegraphics[width=\linewidth]{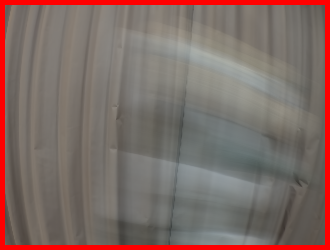}}{6}
      &\addimgTtext{\includegraphics[width=\linewidth]{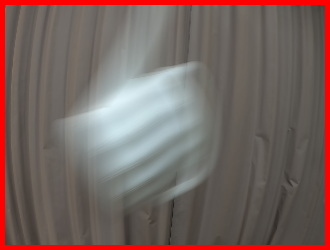}}{4}
      &\addimgTtext{\includegraphics[width=\linewidth]{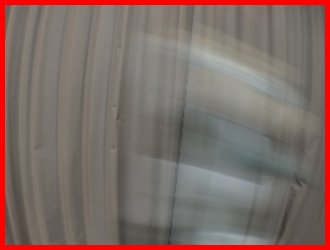}}{5}
      &\addimgTtext{\includegraphics[width=\linewidth]{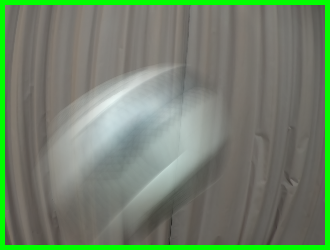}}{4}
      &\addimgTtext{\includegraphics[width=\linewidth]{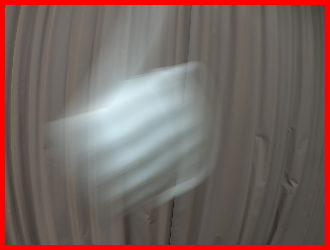}}{4}
      &\addimgTtext{\includegraphics[width=\linewidth]{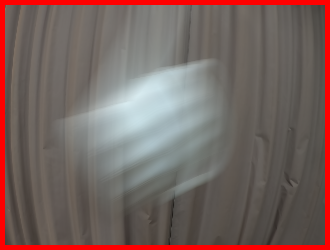}}{4}
      &\addimgTtext{\includegraphics[width=\linewidth]{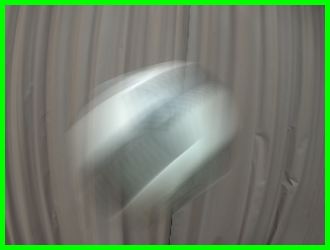}}{4}
        \\
    &&\addimgTtext{\includegraphics[width=\linewidth]{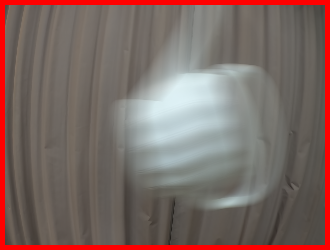}}{3}
      &\addimgTtext{\includegraphics[width=\linewidth]{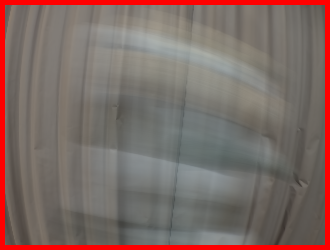}}{6}
      &\addimgTtext{\includegraphics[width=\linewidth]{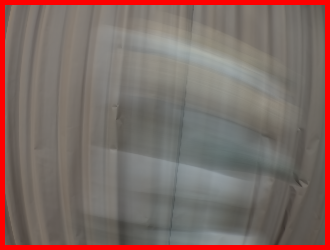}}{5}
      &\addimgTtext{\includegraphics[width=\linewidth]{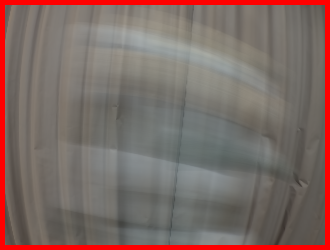}}{6}
      &\addimgTtext{\includegraphics[width=\linewidth]{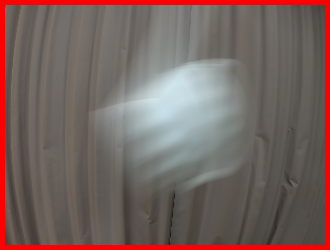}}{4}
      &\addimgTtext{\includegraphics[width=\linewidth]{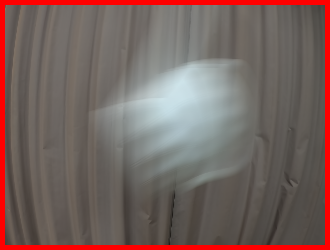}}{4}
      &\addimgTtext{\includegraphics[width=\linewidth]{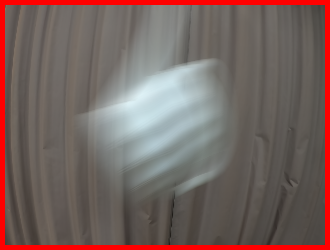}}{3}
      &\addimgTtext{\includegraphics[width=\linewidth]{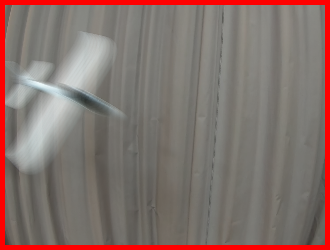}}{3}
      &\addimgTtext{\includegraphics[width=\linewidth]{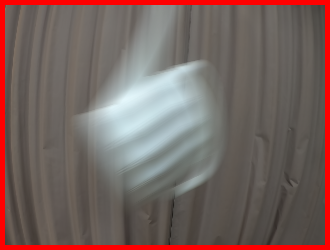}}{3}
      &\addimgTtext{\includegraphics[width=\linewidth]{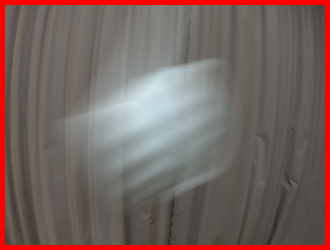}}{5} \\      \addlinespace[5pt]
\raisebox{0em}{\multirow{2}{*}{ 
\addimgTtext{\includegraphics[width=\linewidth]{fig/suppmat/qualitativeBL5_suppmat_football/query_image_1425.png}}{5}}}
&\raisebox{1.10em}{\multirow{2}{*}{\rotatebox[origin=c]{90}{DOLG~\cite{dolg}}}}
    &\addimgTtext{\includegraphics[width=\linewidth]{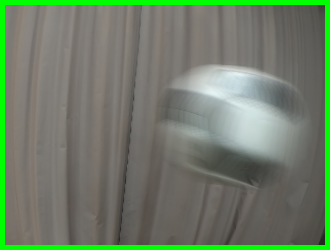}}{3}
      &\addimgTtext{\includegraphics[width=\linewidth]{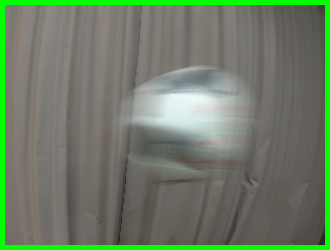}}{3}
      &\addimgTtext{\includegraphics[width=\linewidth]{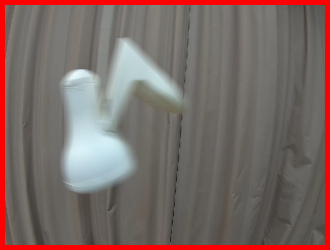}}{2}
      &\addimgTtext{\includegraphics[width=\linewidth]{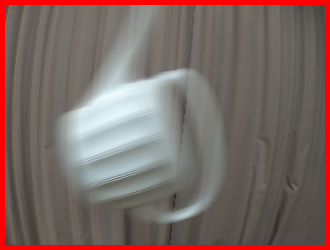}}{2}
      &\addimgTtext{\includegraphics[width=\linewidth]{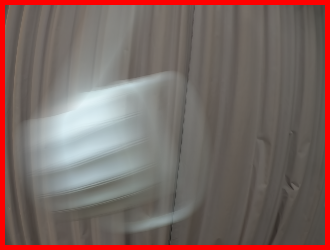}}{3}
      &\addimgTtext{\includegraphics[width=\linewidth]{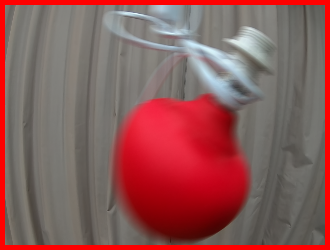}}{1}
      &\addimgTtext{\includegraphics[width=\linewidth]{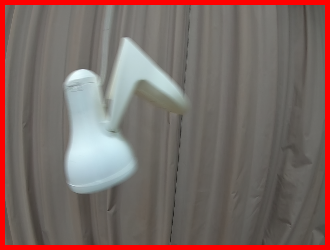}}{1}
      &\addimgTtext{\includegraphics[width=\linewidth]{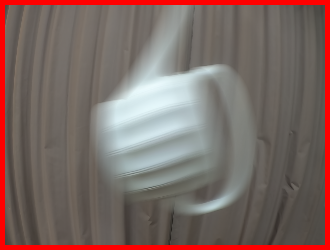}}{2}
      &\addimgTtext{\includegraphics[width=\linewidth]{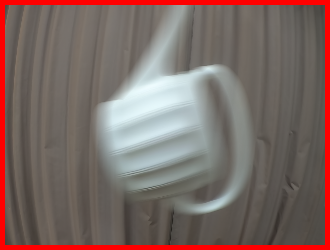}}{2}
      &\addimgTtext{\includegraphics[width=\linewidth]{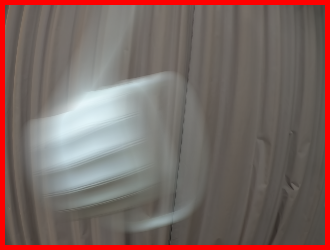}}{3}
        \\
    &&\addimgTtext{\includegraphics[width=\linewidth]{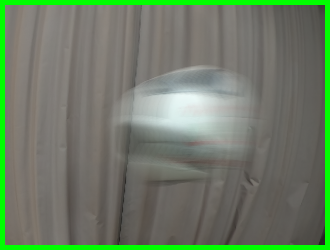}}{3}
      &\addimgTtext{\includegraphics[width=\linewidth]{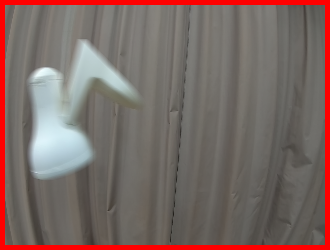}}{1}
      &\addimgTtext{\includegraphics[width=\linewidth]{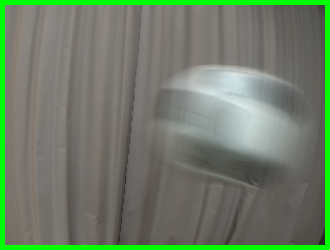}}{3}
      &\addimgTtext{\includegraphics[width=\linewidth]{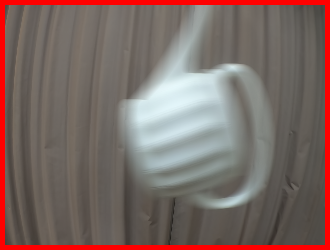}}{2}
      &\addimgTtext{\includegraphics[width=\linewidth]{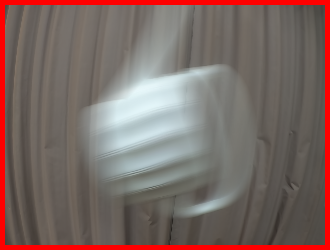}}{2}
      &\addimgTtext{\includegraphics[width=\linewidth]{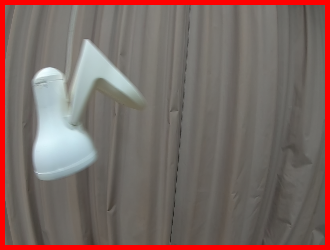}}{1}
      &\addimgTtext{\includegraphics[width=\linewidth]{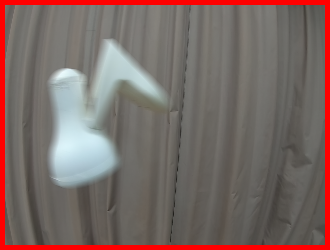}}{1}
      &\addimgTtext{\includegraphics[width=\linewidth]{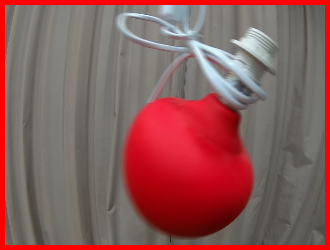}}{1}
      &\addimgTtext{\includegraphics[width=\linewidth]{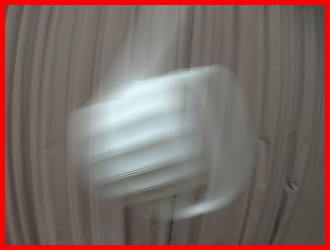}}{3}
      &\addimgTtext{\includegraphics[width=\linewidth]{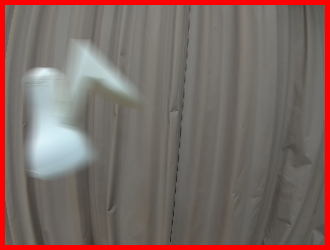}}{3} \\      \addlinespace[5pt]
\raisebox{0.1em}{\multirow{2}{*}{
\addimgTtext{\includegraphics[width=\linewidth]{fig/suppmat/qualitativeBL5_suppmat_football/query_image_1425.png}}{5}}}
&\raisebox{1.1em}{\multirow{2}{*}{\rotatebox[origin=c]{90}{Token~\cite{token}}}}
    &\addimgTtext{\includegraphics[width=\linewidth]{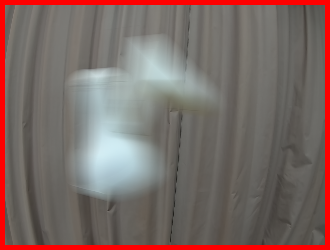}}{3}
      &\addimgTtext{\includegraphics[width=\linewidth]{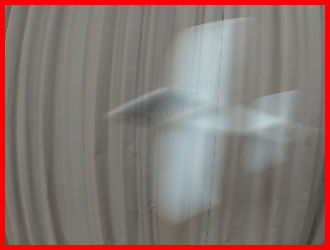}}{4}
      &\addimgTtext{\includegraphics[width=\linewidth]{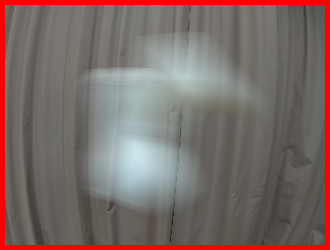}}{4}
      &\addimgTtext{\includegraphics[width=\linewidth]{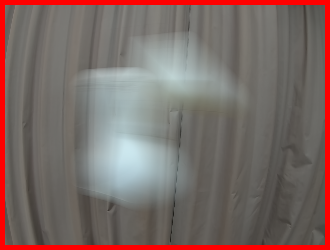}}{4}
      &\addimgTtext{\includegraphics[width=\linewidth]{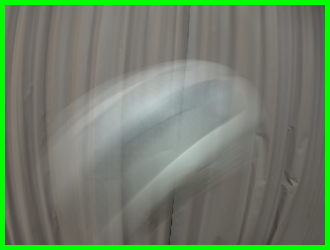}}{6}
      &\addimgTtext{\includegraphics[width=\linewidth]{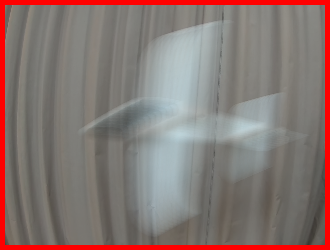}}{4}
      &\addimgTtext{\includegraphics[width=\linewidth]{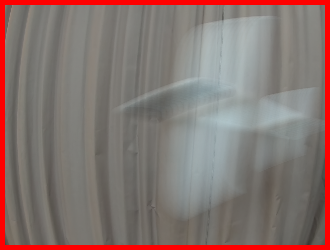}}{4}
      &\addimgTtext{\includegraphics[width=\linewidth]{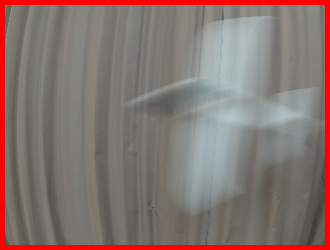}}{4}
      &\addimgTtext{\includegraphics[width=\linewidth]{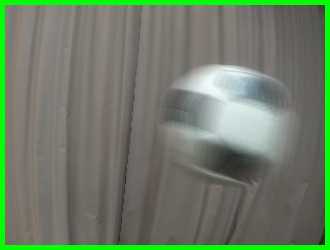}}{3}
      &\addimgTtext{\includegraphics[width=\linewidth]{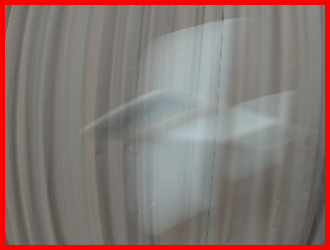}}{4}
        \\
    &&\addimgTtext{\includegraphics[width=\linewidth]{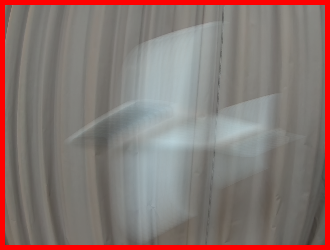}}{4}
      &\addimgTtext{\includegraphics[width=\linewidth]{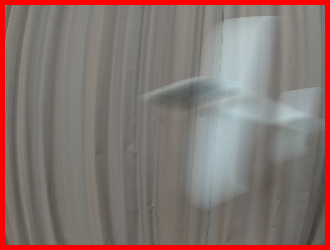}}{4}
      &\addimgTtext{\includegraphics[width=\linewidth]{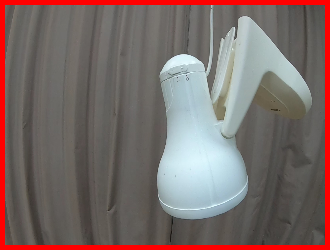}}{1}
      &\addimgTtext{\includegraphics[width=\linewidth]{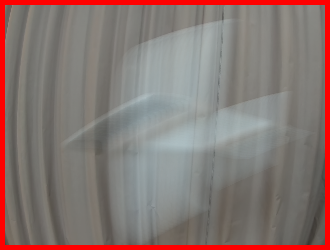}}{4}
      &\addimgTtext{\includegraphics[width=\linewidth]{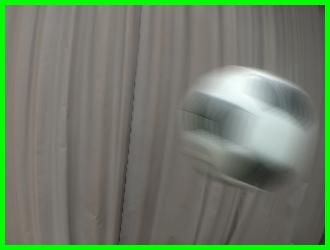}}{3}
      &\addimgTtext{\includegraphics[width=\linewidth]{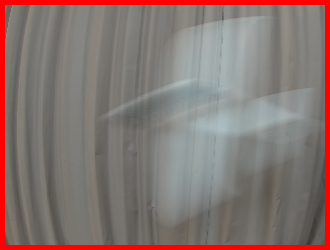}}{4}
      &\addimgTtext{\includegraphics[width=\linewidth]{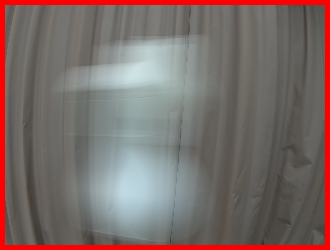}}{5}
      &\addimgTtext{\includegraphics[width=\linewidth]{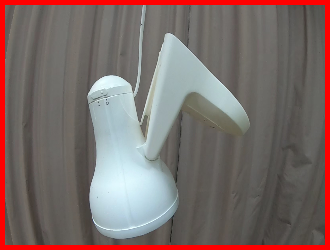}}{1}
      &\addimgTtext{\includegraphics[width=\linewidth]{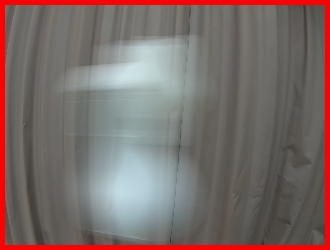}}{4}
      &\addimgTtext{\includegraphics[width=\linewidth]{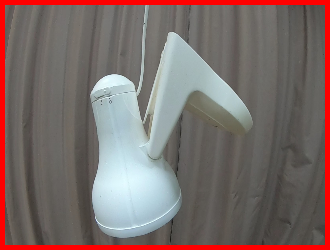}}{1} \\      \addlinespace[5pt]
\raisebox{0em}{\multirow{2}{*}{ 
\addimgTtext{\includegraphics[width=\linewidth]{fig/suppmat/qualitativeBL5_suppmat_football/query_image_1425.png}}{5}}}
&\raisebox{0.1em}{\multirow{2}{*}{\rotatebox[origin=c]{90}{Ours}}}
    &\addimgTtext{\includegraphics[width=\linewidth]{fig/suppmat/qualitativeBL5_suppmat_football/Ours/rank00_True.png}}{3}
      &\addimgTtext{\includegraphics[width=\linewidth]{fig/suppmat/qualitativeBL5_suppmat_football/Ours/rank01_True.png}}{3}
      &\addimgTtext{\includegraphics[width=\linewidth]{fig/suppmat/qualitativeBL5_suppmat_football/Ours/rank02_True.png}}{3}
      &\addimgTtext{\includegraphics[width=\linewidth]{fig/suppmat/qualitativeBL5_suppmat_football/Ours/rank03_True.png}}{3}
      &\addimgTtext{\includegraphics[width=\linewidth]{fig/suppmat/qualitativeBL5_suppmat_football/Ours/rank04_True.png}}{3}
      &\addimgTtext{\includegraphics[width=\linewidth]{fig/suppmat/qualitativeBL5_suppmat_football/Ours/rank05_True.png}}{2}
      &\addimgTtext{\includegraphics[width=\linewidth]{fig/suppmat/qualitativeBL5_suppmat_football/Ours/rank06_True.png}}{3}
      &\addimgTtext{\includegraphics[width=\linewidth]{fig/suppmat/qualitativeBL5_suppmat_football/Ours/rank07_True.png}}{3}
      &\addimgTtext{\includegraphics[width=\linewidth]{fig/suppmat/qualitativeBL5_suppmat_football/Ours/rank08_True.png}}{4}
      &\addimgTtext{\includegraphics[width=\linewidth]{fig/suppmat/qualitativeBL5_suppmat_football/Ours/rank09_True.png}}{2}
        \\
    &&\addimgTtext{\includegraphics[width=\linewidth]{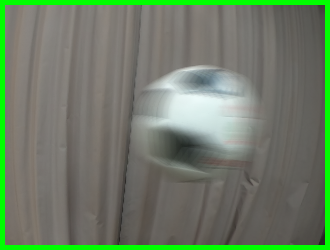}}{3}
      &\addimgTtext{\includegraphics[width=\linewidth]{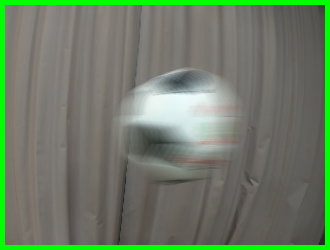}}{3}
      &\addimgTtext{\includegraphics[width=\linewidth]{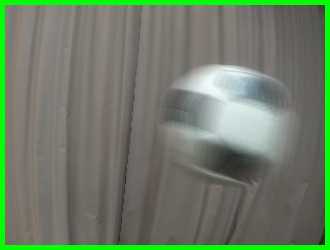}}{3}
      &\addimgTtext{\includegraphics[width=\linewidth]{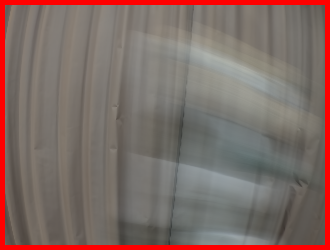}}{5}
      &\addimgTtext{\includegraphics[width=\linewidth]{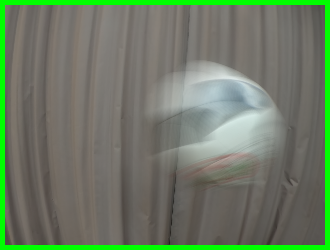}}{3}
      &\addimgTtext{\includegraphics[width=\linewidth]{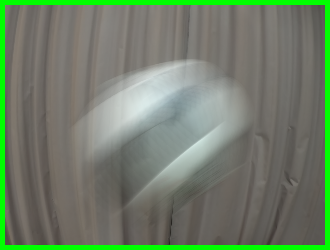}}{4}
      &\addimgTtext{\includegraphics[width=\linewidth]{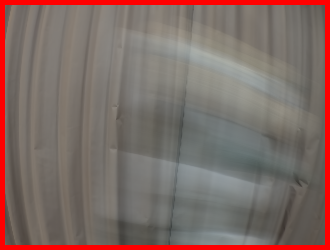}}{6}
      &\addimgTtext{\includegraphics[width=\linewidth]{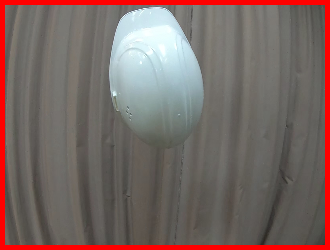}}{1}
      &\addimgTtext{\includegraphics[width=\linewidth]{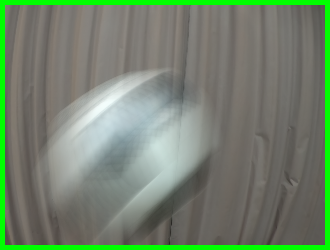}}{4}
      &\addimgTtext{\includegraphics[width=\linewidth]{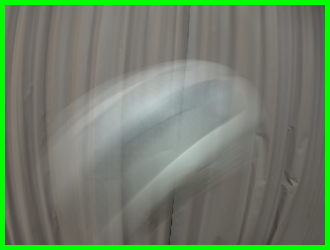}}{6} \\
    \end{tabular}
    \label{fig:real_qualitative_1}
\end{subfigure}
\caption{\textbf{Comparison of retrieval results} between our approach and state-of-the-art retrieval methods on our real-world dataset. The retrieved images are sorted from left to right, top to bottom, with the ranking from the 1\textsuperscript{st} to the 20\textsuperscript{th}. The blur level of each image is shown in the bottom left corner.
Correct results are indicated with green boxes, while incorrect ones are marked with red boxes.}
\label{fig:qualitative_suppmat_real}

\end{figure*}

\clearpage

%
%
\bibliographystyle{splncs04}
\bibliography{main}
\end{document}